\documentclass[acmsmall]{acmart}  
\AtBeginDocument{%
  \providecommand\BibTeX{{%
    \normalfont B\kern-0.5em{\scshape i\kern-0.25em b}\kern-0.8em\TeX}}}

\setcopyright{acmcopyright}
\copyrightyear{2021}
\acmYear{2021}
\acmDOI{10.1145/1122445.1122456}

\acmJournal{JACM}
\acmVolume{37}
\acmNumber{4}
\acmArticle{111}
\acmMonth{8}

\usepackage{float}
\usepackage{enumerate}
\usepackage{subfig}
\usepackage{graphicx}
\usepackage{xcolor}
\usepackage{enumitem}
\usepackage{pgfplotstable}
\usepackage{url}
\pgfplotsset{compat=1.9}
\RequirePackage[normalem]{ulem} 
\RequirePackage{color}\definecolor{RED}{rgb}{1,0,0}\definecolor{BLUE}{rgb}{0,0,1} 

\providecommand{\DIFadd}[1]{{{#1}}} 
\providecommand{\DIFdel}[1]{}

\providecommand{\DIFaddbegin}{} 
\providecommand{\DIFaddend}{} 
\providecommand{\DIFdelbegin}{} 
\providecommand{\DIFdelend}{} 
\providecommand{\DIFdelFL}[1]{\DIFdel{#1}} 
\providecommand{\DIFaddbeginFL}{} 
\providecommand{\DIFaddendFL}{} 
\providecommand{\DIFdelbeginFL}{} 
\providecommand{\DIFdelendFL}{} 
\newcommand{\DIFscaledelfig}{0.5}
\RequirePackage{settobox} 
\RequirePackage{letltxmacro} 
\newsavebox{\DIFdelgraphicsbox} 
\newlength{\DIFdelgraphicswidth} 
\newlength{\DIFdelgraphicsheight} 
\LetLtxMacro{\DIFOincludegraphics}{\includegraphics} 
\newcommand{\DIFaddincludegraphics}[2][]{{\color{blue}\fbox{\DIFOincludegraphics[#1]{#2}}}} 
\newcommand{\DIFdelincludegraphics}[2][]{
\sbox{\DIFdelgraphicsbox}{\DIFOincludegraphics[#1]{#2}}
\settoboxwidth{\DIFdelgraphicswidth}{\DIFdelgraphicsbox} 
\settoboxtotalheight{\DIFdelgraphicsheight}{\DIFdelgraphicsbox} 
\scalebox{\DIFscaledelfig}{
\parbox[b]{\DIFdelgraphicswidth}{\usebox{\DIFdelgraphicsbox}\\[-\baselineskip] \rule{\DIFdelgraphicswidth}{0em}}\llap{\resizebox{\DIFdelgraphicswidth}{\DIFdelgraphicsheight}{
\setlength{\unitlength}{\DIFdelgraphicswidth}
\begin{picture}(1,1)
\thicklines\linethickness{2pt} 
{\color[rgb]{1,0,0}\put(0,0){\framebox(1,1){}}}
{\color[rgb]{1,0,0}\put(0,0){\line( 1,1){1}}}
{\color[rgb]{1,0,0}\put(0,1){\line(1,-1){1}}}
\end{picture}
}\hspace*{3pt}}} 
} 
\LetLtxMacro{\DIFOaddbegin}{\DIFaddbegin} 
\LetLtxMacro{\DIFOaddend}{\DIFaddend} 
\LetLtxMacro{\DIFOdelbegin}{\DIFdelbegin} 
\LetLtxMacro{\DIFOdelend}{\DIFdelend} 
\DeclareRobustCommand{\DIFaddbegin}{\DIFOaddbegin \let\includegraphics\DIFaddincludegraphics} 
\DeclareRobustCommand{\DIFaddend}{\DIFOaddend \let\includegraphics\DIFOincludegraphics} 
\DeclareRobustCommand{\DIFdelbegin}{\DIFOdelbegin \let\includegraphics\DIFdelincludegraphics} 
\DeclareRobustCommand{\DIFdelend}{\DIFOaddend \let\includegraphics\DIFOincludegraphics} 
\LetLtxMacro{\DIFOaddbeginFL}{\DIFaddbeginFL} 
\LetLtxMacro{\DIFOaddendFL}{\DIFaddendFL} 
\LetLtxMacro{\DIFOdelbeginFL}{\DIFdelbeginFL} 
\LetLtxMacro{\DIFOdelendFL}{\DIFdelendFL} 
\DeclareRobustCommand{\DIFaddbeginFL}{\DIFOaddbeginFL \let\includegraphics\DIFaddincludegraphics} 
\DeclareRobustCommand{\DIFaddendFL}{\DIFOaddendFL \let\includegraphics\DIFOincludegraphics} 
\DeclareRobustCommand{\DIFdelbeginFL}{\DIFOdelbeginFL \let\includegraphics\DIFdelincludegraphics} 
\DeclareRobustCommand{\DIFdelendFL}{\DIFOaddendFL \let\includegraphics\DIFOincludegraphics} 

\begin{document}

\title{Robotic Vision for Human-Robot Interaction and Collaboration: A Survey and Systematic Review}
\author{Nicole Robinson}
\email{nicole.robinson@monash.edu}
\orcid{0000-0002-7144-3082}
\affiliation{%
  \institution{Australian Research Council Centre of Excellence for Robotic Vision, School of Electrical Engineering \& Robotics, QUT Centre for Robotics, Queensland University of Technology. Faculty of Engineering, Turner Institute for Brain and Mental Health, Monash University}
  \streetaddress{18 Alliance Lane}
  \city{Clayton}
  \state{Victoria}
  \country{Australia}
  \postcode{3800}
}

\author{Brendan Tidd}
\email{brendan.tidd@hdr.qut.edu.au}
\orcid{0000-0002-7721-7799}
\affiliation{%
  \institution{Australian Research Council Centre of Excellence for Robotic Vision, School of Electrical Engineering \& Robotics, QUT Centre for Robotics, Queensland University of Technology}
  \streetaddress{2 George Street}
  \city{Brisbane}
  \state{Queensland}
  \country{Australia}
  \postcode{4000}
}

\author{Dylan Campbell}
\email{dylan@robots.ox.ac.uk}
\orcid{0000-0002-4717-6850}
\affiliation{%
  \institution{Visual Geometry Group, Department of Engineering Science, University of Oxford}
  \streetaddress{17 Parks Road}
  \city{Oxford}
  \state{Oxfordshire}
  \country{United Kingdom}
  \postcode{OX1 3PJ}
}

\author{Dana Kuli{\'c}}
\email{dana.kulic@monash.edu}
\orcid{0000-0002-4169-2141}
\affiliation{%
  \institution{Australian Research Council Centre of Excellence for Robotic Vision, Faculty of Engineering, Monash University}
  \streetaddress{18 Alliance Lane}
  \city{Clayton}
  \state{Victoria}
  \country{Australia}
  \postcode{3800}
}

\author{Peter Corke}
\email{peter.corke@qut.edu.au}
\orcid{0000-0001-6650-367X}
\affiliation{%
  \institution{Australian Research Council Centre of Excellence for Robotic Vision, School of Electrical Engineering \& Robotics, QUT Centre for Robotics, Queensland University of Technology}
  \streetaddress{2 George Street}
  \city{Brisbane}
  \state{Queensland}
  \country{Australia}
  \postcode{4000}
}

\renewcommand{\shortauthors}{Robinson et al.}

\begin{abstract}
Robotic vision for human-robot interaction and collaboration is a critical process for robots to collect and interpret detailed information related to human actions, goals, and preferences, enabling robots to provide more useful services to people. This survey and systematic review presents a comprehensive analysis on robotic vision in human-robot interaction and collaboration over the last 10 years. From a detailed search of 3850 articles, systematic extraction and evaluation was used to identify and explore 310 papers in depth. These papers described robots with some level of autonomy using robotic vision for locomotion, manipulation and/or visual communication to collaborate or interact with people. This paper provides an in-depth analysis of current trends, common domains, methods and procedures, technical processes, data sets and models, experimental testing, sample populations, performance metrics and future challenges. This manuscript found that robotic vision was often used in action and gesture recognition, robot movement in human spaces, object handover and collaborative actions, social communication and learning from demonstration. Few high-impact and novel techniques from the computer vision field had been translated into human-robot interaction and collaboration. Overall, notable advancements have been made on how to develop and deploy robots to assist people.  \end{abstract}

\ccsdesc[500]{General and reference~Surveys and overviews}
\ccsdesc[300]{Human-centered computing~HCI theory, concepts and models}
\ccsdesc[500]{Computing methodologies~Vision for robotics}

\keywords{robotic vision, computer vision, human-robot interaction, gesture recognition, robot movement in human spaces, object handover, collaborative actions, learning from demonstration, social communication}

\maketitle
\section{Introduction}\label{intro}
This paper presents a comprehensive survey and review of robotic vision \DIFaddbegin \DIFadd{methods }\DIFaddend for human-robot interaction and collaboration (HRI/C) \DIFdelbegin \DIFdel{. Robotic vision (RV) uses computer vision (CV) }\DIFdelend \DIFaddbegin \DIFadd{based on a  review of 3850 articles to create a collection of 310 eligible articles for in-depth analysis. The selected 310 published papers examine  how  robotic vision is used to facilitate human-robot interaction tasks such as robot navigation in human spaces, social interaction with people to exchange information, and human-robot handovers of everyday objects. 
} Such a \DIFdelbegin \DIFdel{review }\DIFdelend \DIFaddbegin \DIFadd{combination of a systematic review to calculate trends and prevalence alongside a comprehensive survey for each subsection }\DIFaddend will help to explore emerging patterns, \DIFdelbegin \DIFdel{trends, common use cases }\DIFdelend \DIFaddbegin \DIFadd{statistical trends }\DIFaddend and recommendations on how robotic vision can help to improve human-robot interaction and collaboration. 

\subsection{\DIFadd{Purpose}}\label{purpose}
\DIFadd{The purpose of this systematic review and survey was to provide detailed insight into underlying emergent research themes pursued by the community, and to explore the trajectory and impact that robotic vision will have on enabling robots to better interact and collaborate with humans.  For the purpose of this paper, robotic vision will be defined as computer vision that is used }\DIFaddend to inform or direct a robot on what actions to perform \DIFdelbegin \DIFdel{. Robotic }\DIFdelend \DIFaddbegin \DIFadd{that will contribute to achieving the chosen goal. In practice, robotic }\DIFaddend vision can enable robots to sense, perceive, and respond to people \DIFdelbegin \DIFdel{by providing }\DIFdelend \DIFaddbegin \DIFadd{through capturing and responding to a }\DIFaddend rich continuous information \DIFdelbegin \DIFdel{about human states, actions, intentions, and communication . Robotic vision can be used in human-machine interaction to facilitate information exchange, create }\DIFdelend \DIFaddbegin \DIFadd{stream. Visual information provided by humans can help robots to better understand the scenario and to plan their actions, such as interpreting hand gesture movements as communication signals and human body movements as an indication of future intent to perform an action. Robots with robotic vision can therefore help to create and facilitate an important information exchange between the human and the robot, opening up }\DIFaddend new communication channels \DIFaddbegin \DIFadd{using a method that is natural and intuitive to people}\DIFaddend , improving the effectiveness of collaborative tasks. 

\subsection{\DIFadd{Scope}}\label{scope}
\DIFaddend This survey and review \DIFdelbegin \DIFdel{covered }\DIFdelend \DIFaddbegin \DIFadd{explored }\DIFaddend published papers from the last 10 years using a systematic search, screen and evaluation protocol \DIFdelbegin \DIFdel{. This review provides }\DIFdelend \DIFaddbegin \DIFadd{to extract }\DIFaddend a general overview of current research trends, common applications and domains, methods and procedures, technical processes, relevant data sets and models, experimental testing setups, sample populations, vision algorithm metrics, and performance evaluations. \DIFaddbegin \DIFadd{To create the systematic search strategy, several key parameters needed to be defined before commencing the extraction and evaluation of papers. Firstly, given the extensive scope of reviewing all relevant papers in the broad field of robotic vision for human-robot interaction and collaboration, this review focused on the last 10 years (i.e. 2010-2020). This time frame helped to showcase the more contemporary use of robotic vision based on newly emergent techniques, and was chosen to coincide with the introduction of critical camera hardware that boosted the applied use of robotic vision to enable robots to be more reactive and suitable for human interaction, such as the release date of the Kinect camera~\mbox{
\cite{han2013enhanced}}\hspace{0pt}}.
\DIFaddend
\subsection{\DIFadd{Related Surveys and Systematic Reviews}}\label{Surveys}
\DIFadd{No systematic review or comprehensive survey on the development and use of robotic vision }\DIFaddend for human-robot interaction and collaboration \DIFaddbegin \DIFadd{had been conducted}\DIFaddend . Current surveys or reviews have \DIFdelbegin \DIFdel{instead addressed specific application domains such as }\DIFdelend \DIFaddbegin \DIFadd{focused on other areas, such as specific domains including robotics in }\DIFaddend industry~\cite{VILLANI2018248, hentout2019human}, agriculture~\cite{VASCONEZ201935}, public \DIFdelbegin \DIFdel{settings}\DIFdelend \DIFaddbegin \DIFadd{areas}\DIFaddend ~\cite{10.1007/978-3-319-47437-3_74}, healthcare~\cite{RobinsonSys} and education~\cite{belpaeme2018social}. \DIFdelbegin \DIFdel{Others surveys or reviews have focused on work related to other components}\DIFdelend \DIFaddbegin \DIFadd{All of these works described different robots or methods relevant to the domain of interest, including describing robot types and use cases that did not use robotic vision. Other reviews or surveys focused on components related to the process of human-robot interaction and collaboration}\DIFaddend , such as \DIFdelbegin \DIFdel{touch }\DIFdelend \DIFaddbegin \DIFadd{the use of physical touch and tactile sensing techniques~}\DIFaddend \cite{argall2010survey}, safety \DIFaddbegin \DIFadd{bounds for vision-based safety systems~}\DIFaddend \cite{ZACHARAKI2020104667, halme2018review}, trust \DIFaddbegin \DIFadd{modelling and trust-related factors~}\DIFaddend \cite{khavas2020modeling, doi:10.1177/0018720811417254}, distance \DIFdelbegin \DIFdel{\mbox{
\cite{leichtmann2020much} }\hspace{0pt}
and }\DIFdelend \DIFaddbegin \DIFadd{between humans and robots~\mbox{
\cite{leichtmann2020much} }\hspace{0pt}
and the use of }\DIFaddend non-verbal communication \cite{saunderson2019robots}. There have also been other published works on specific methods \DIFaddend used in human-robot interaction \DIFaddbegin \DIFadd{that did not have a direct focus on vision}\DIFaddend , such as \DIFaddbegin \DIFadd{tests with }\DIFaddend psycho-physiological measures~\cite{4415182}\DIFdelbegin \DIFdel{or robot perception in general~\mbox{
\cite{yan2014survey}}\hspace{0pt}
. Others instead focused on }\DIFdelend \DIFaddbegin \DIFadd{, exploring general robot perception methods~\mbox{
\cite{yan2014survey}}\hspace{0pt}
, investigating }\DIFaddend a single robot platform~\cite{tezza2019state} or \DIFaddbegin \DIFadd{a specific }\DIFaddend form of robot behaviour~\cite{mi2013human}.  
\DIFdelbegin \DIFdel{In the computer vision field, published reviews }\DIFdelend \DIFaddbegin 

\DIFadd{The computer vision field has contributed to providing detailed surveys and reviews that show the technical process for computer vision }\DIFaddend related to humans\DIFdelbegin \DIFdel{include more general }\DIFdelend \DIFaddbegin \DIFadd{, such as gesture-based }\DIFaddend human-machine interaction~\cite{suma2019computer} \DIFdelbegin \DIFdel{, }\DIFdelend \DIFaddbegin \DIFadd{and }\DIFaddend multi-modal machine collaboration \DIFdelbegin \DIFdel{~\mbox{
\cite{jaimes2007multimodal} }\hspace{0pt}
}\DIFdelend \DIFaddbegin \DIFadd{with a focus on body, gesture, gaze, and affective interaction~\mbox{
\cite{jaimes2007multimodal}}\hspace{0pt}
. Others have explored more detailed }\DIFaddend and specific use cases such as action recognition~\cite{zhang2019comprehensive}, hand gesture recognition~\cite{rautaray2015vision, liu2018gesture,xia_vision-based_2019} and human motion capture~\cite{moeslund2001survey}. \DIFdelbegin \DIFdel{However, no }\DIFdelend \DIFaddbegin There have also been detailed surveys and reviews that explored vision-based techniques in robots. For instance, reviews or surveys that included recent developments in robotic vision techniques~\mbox{
\cite{chen2011kalman}}\hspace{0pt}
, learning for robotic vision~\mbox{
\cite{10.1007/978-3-030-28619-4_14} }\hspace{0pt}
and the use of computer vision for a specific type of robot, such as aerial robots~\mbox{
\cite{5508131}}\hspace{0pt}
. Other surveys and reviews instead had more general overviews of vision for robots such as object recognition and modeling, site reconstruction and inspection, robotic manipulation, localization, path following, map construction, autonomous navigation and exploration~\mbox{
\cite{chen2011active, bonin2008visual}}\hspace{0pt}
. Others also included a brief mention of applications to people, but did not provide a detailed analysis on how this could better facilitate human-robot interaction and collaboration across different technique types. In collection, the identified surveys and reviews provided an excellent commentary on their respective fields and target areas, but there were limited works that presented a detailed investigation into robotic vision techniques, hardware integration, and evaluation \DIFaddend of its use \DIFaddbegin \DIFadd{in real-world scenarios for human-robot interaction and collaboration}\DIFaddend. 

\subsection{Contribution}\label{contribution}
\DIFadd{The contribution of this paper is the systematic extraction, discovery and detailed synthesis of literature to showcase the current use of robotic vision for robots that can interact and collaborate with people. This survey and systematic review contributes new knowledge on how robots can be improved by integrating and refining functionality related to robotic vision, showcases real-world use of robots with vision capabilities to improve collaborative outcomes, and provides a critical discussion to help push the field forward.
}

\section{Background}\label{background}
\DIFdelbegin \DIFdel{Computer vision }\DIFdelend \DIFaddbegin \subsection{\DIFadd{A Brief History of the Field of Computer Vision}}\label{CVintro}
\DIFadd{Computer vision is important to help machines to better understand and interact with the real world, making relevant actions and decisions based on visual information \mbox{
\cite{corke2011robotics}}\hspace{0pt}
.  }\DIFaddend  Common sensors in computer vision include RGB cameras \DIFdelbegin \DIFdel{, which provide color information }\DIFdelend \DIFaddbegin \DIFadd{which provide detailed information by capturing light in red, green and blue wavelengths (RGB) to create a color representation of the world. The use of visual information to understand the world can help emulate how humans perceive the world, creating a common language and understanding between humans and robots when sharing details, objects and task-related information}\DIFaddend . \DIFdelbegin \DIFdel{Visual information can then be used to help detect objects~\mbox{
\cite{girshick_fast_rcnn, liu_ssd_2016}}\hspace{0pt}
, understand scenes~\mbox{
\cite{ronneberger2015unet, he2017mask} }\hspace{0pt}
and classify human actions \mbox{
\cite{han2013enhanced, carreira2017quo, feichtenhofer2019slowfast}}\hspace{0pt}
. }\DIFdelend Computer vision involves techniques such as object detection to localise where an object is in the scene, image classification to determine what it is in the image, and pixel level classification to classify what part of the image belongs to an area of interest~\DIFdelbegin \DIFdel{\mbox{
\cite{ForsythDavid2012Cv:a}}\hspace{0pt}
. Computer vision also addresses }\DIFdelend \DIFaddbegin \DIFadd{\mbox{
\cite{ForsythDavid2012Cv:a, corke2011robotics}}\hspace{0pt}
. In relation to computer vision for humans, computer vision can address }\DIFaddend the detection and analysis of humans in visual scenes, including methods such as face detection~\cite{viola_robust_2004}, pose estimation~\cite{Cao_2017_CVPR}, and human motion tracking~\cite{henriques_high-speed_2015}. \DIFdelbegin \DIFdel{Computer vision is important to help machines to better understand and interact with the real world, making relevant actions and decisions based on visual information}\DIFdelend This type of visual information can then further assist in creating shared knowledge and understanding between humans and machines. \DIFdelbegin \DIFdel{This includes being able to identify the same objects, people, places, and events within a scene, working together to perform an action or reach a goal. }\DIFdelend 

\DIFdelbegin \DIFdel{Robotic vision can enable }\DIFdelend 
\DIFadd{The field of computer vision has evolved rapidly in the last decade from 2010 to 2020. Deep learning has played a dominant role since its success at the 2012 ImageNet competition \mbox{
\cite{krizhevsky2012imagenet}}\hspace{0pt}
 . Learning complex parametrised functions from data has also served to make computer vision algorithms more robust and effective in real-world situations, making it ideal for the field of human-robot interaction and collaboration. However, this comes at the expense of increased hardware requirements and longer development time, such as the need for data collection, labelling, and network training. Another significant change at the start of this period was the advent of more readily available RGB-D sensors with the Microsoft Kinect camera released in 2010~\mbox{
\cite{han2013enhanced}}\hspace{0pt}
. This allowed researchers to reason about colour and 2.5D geometry jointly, facilitating new breakthroughs such as real-time 3D reconstruction \mbox{
\cite{izadi2011kinectfusion}}\hspace{0pt}
. 

In the decades before this period, computer vision had several major successes relevant to human-robot interaction and collaboration. The first was the codification of the principles of multiple-view geometry \mbox{
\cite{hartley2003multiple} }\hspace{0pt}
and their successful application in large-scale reconstruction tasks \mbox{
\cite{agarwal2009building} }\hspace{0pt}
using the techniques of structure-from-motion. The period was also marked by increasingly sophisticated handcrafted features such as SIFT \mbox{
\cite{lowe1999object} }\hspace{0pt}
and histogram of oriented gradients (HOG) \mbox{
\cite{dalal2005histograms} }\hspace{0pt}
features and the use of increasingly sophisticated learning algorithms, such as kernel support vector machines (SVM) \mbox{
\cite{boser1992training} }\hspace{0pt}
and AdaBoost \mbox{
\cite{freund1997decision}}\hspace{0pt}
, the latter used to great effect in the Viola--Jones face detector \mbox{
\cite{viola2001rapid}}\hspace{0pt}
. The topics of image classification, object detection, image segmentation, and optical flow received significant research attention, among many others. Some highlights include deformable part models \mbox{
\cite{felzenszwalb2005pictorial, felzenszwalb2009object} }\hspace{0pt}
that demonstrated unprecedented performance on object detection benchmarks before deep learning, conditional random fields for image segmentation \mbox{
\cite{triggs2007scene, gould2008multi, krahenbuhl2011efficient}}\hspace{0pt}
, graph cuts for tasks such as stereo depth estimation \mbox{
\cite{boykov2001fast}}\hspace{0pt}
, and variational methods for optical flow estimation \mbox{
\cite{lucas1981iterative, horn1981determining}}\hspace{0pt}
. These approaches continue to be used in robotics and embodied vision settings due to their efficiency and low hardware requirements.
}

\subsection{\DIFadd{A Brief History of the Field of Robotic Vision}}\label{historyrv}
\DIFadd{Robotic vision, by contrast, exists at the intersection of robotics and computer vision, enabling }\DIFaddend robots to sense, perceive, and respond to people by providing rich, continuous information about human states, actions, intentions, and communication. \DIFdelbegin \DIFdel{Computer vision that is }\DIFdelend \DIFaddbegin \DIFadd{Robotic vision involves a vision sensor (RGB, RGB-D) and supporting algorithms that translate raw images to a control signal for a robot. That is, any computer vision techniques }\DIFaddend used to guide a robot on what action to perform \DIFdelbegin \DIFdel{is otherwise known as }\DIFdelend \DIFaddbegin \DIFadd{can be considered }\DIFaddend robotic vision. Robotic vision \DIFdelbegin \DIFdel{has started to become a key perception channel for the robot to interact with and provide assistance to people. Many human-based computer vision topics are related to robotic vision for HRI/C, and some of those techniques could have benefit or value if used within a robot. Robotic vision has }\DIFdelend benefited from the advancements in the computer vision research community, such as large datasets, computing power, complex algorithms and scientific methods.  \DIFaddbegin \DIFadd{Robotic vision has started to become a key perception channel for the robot to interact with and provide assistance to people. Robotic vision has important advantages for enabling robots to smartly interact with the environment, such as better camera control, physical movement around the space, and the capacity to adapt its viewpoint to gather further information\mbox{
~\cite{sunderhauf2018limits, corke2011robotics}}\hspace{0pt}
. There have also been notable advances to effectively handle multi-modal data in robotic sensing, including visual processing for intelligent robot decisions and actions~\mbox{
\cite{doi:10.1080/01691864.2017.1365009,sunderhauf2018limits}}. \hspace{0pt}
Robotic vision can also create new opportunities for humans to interact with \DIFdelbegin \DIFdel{machines }\DIFdelend \DIFaddbegin \DIFadd{robots }\DIFaddend in a way that does not inhibit natural actions\DIFdelbegin \DIFdel{from people}\DIFdelend , such as removing the need to use a computer \DIFaddbegin \DIFadd{terminal }\DIFaddend or to wear \DIFdelbegin \DIFdel{physical apparatus. Therefore, vision }\DIFdelend \DIFaddbegin \DIFadd{a physical apparatus}. Improvements to robots through visual perception can therefore help to contribute to creating more general-purpose robots, extending the potential for a wide range of tasks that a robot can complete for a person\mbox{
~\cite{10.1007/978-3-030-28619-4_14}}\hspace{0pt}
. 
}\DIFaddend 

\DIFaddbegin \subsection{\DIFadd{Human-Robot Interaction and Collaboration}}\label{HRI}
\DIFadd{ Human-robot interaction (HRI) focuses on the interactivity between humans and robots, and often involves creating a robotic system that can identify and respond to the complexities of human behaviour. For the robot to behave in socially acceptable ways, the robot should be able to sense, perceive and respond to human states, actions, intentions and emotions. HRI related topics include  improving robot social acuity using visual perception of the person~\mbox{
\cite{socialerrorsHRI}}\hspace{0pt}
. Human-robot collaboration (HRC) instead focuses on how humans and robots work together to achieve shared goals with a common purpose and directed outcome. In HRC, robots work to complement or add value to the intended goal of the human \mbox{
\cite{bauer2008human}}\hspace{0pt}
. Collaboration with a robot can help to improve task speed and work productivity, reduce the number of errors, and improve human safety to minimise repetition fatigue and injuries~\mbox{
\cite{VASCONEZ201935,HaddadinSami2009RfSR}}\hspace{0pt}
. 
}

\subsubsection{\DIFadd{Robots with Computer Vision to Improve Collaborative Outcomes}}\label{alltogether}
\DIFaddend Robotic vision techniques have been used to create new interaction methods and improve the current process of \DIFdelbegin \DIFdel{HRI/C. For example, vision can }\DIFdelend \DIFaddbegin \DIFadd{human-robot interaction, such as using vision to }\DIFaddend create the ability for people to communicate with the robot, such as to signal information or commands. These contribute to the ability for the robot to provide a more functional service to the human. \DIFaddbegin \DIFadd{Visual information captured by the robot through the camera system can then be used to help enable the robot to make more informed decisions about its next set of actions. Examples could include to detect target objects in its field of view when humans request a specific object, to understand events and scenarios that are occurring in the scene for social group dynamics and to classify and better understand human actions to offer predictive assistance~\mbox{
\cite{ronneberger2015unet, he2017mask, han2013enhanced, carreira2017quo, feichtenhofer2019slowfast, girshick_fast_rcnn, liu_ssd_2016}}\hspace{0pt}
}\DIFaddend . \DIFdelbegin \DIFdel{Visual }\DIFdelend \DIFaddbegin \DIFadd{For instance, visual }\DIFaddend information from people can help robots to make informed decisions on how to interact or assist the person, such as to help the robot to decide how to approach a person \cite{rios2015proxemics}, how to follow a person \cite{doi:10.1177/0278364919881683} or when to offer to hand over an item to a person \cite{ortenzi2021object}. Robotic vision can help to identify, classify or predict human movements through action or activity recognition \cite{beddiar2020vision}. Activity recognition to perceive human movements can give robots the ability to better predict or recognise what a human is doing in the environment so that the robot can better provide useful information, advice or assistance to the person in settings such as in industrial settings \cite{7041588} or in different contexts such as recognition for multiple people in a robot's field of view~\cite{gori_multitype_2016}. Gesture recognition has often been tested as a communication and control method through the translation of human pose into a command signal, an action to trigger a state change, or to signify the start of an information exchange between the human and the robot~\cite{katsuki_high_speed_2015,gori_multitype_2016,tsiami_multi3_2018,jevtic_comparison_2015,shukla_probabilistic_2015}. Gestures can also be used to signal to the robot which object the robot should use~\cite{shukla_probabilistic_2015, shukla_proactive_2017} and where a robot should move~\cite{jevtic_comparison_2015}. Visual information can also support collaborative robot actions with the person such as human-robot object handover through perception and interpretation of humans and objects in the scene, including human reach ability, motion, and collision range \cite{8955665, ortenzi2021object}. There is significant opportunity to draw from principles and concepts of computer vision to improve robot capacity to perceive and act upon visual information to improve human-robot \DIFdelbegin \DIFdel{interaction and human-robot collaboration. Human-robot interaction (HRI) focuses on the interactivity between humans and robots, and often involves creating a robotic system that can identify and respond to the complexities of human behaviour. For the robot to behave in socially acceptable ways, the robot should be able to sense, perceive and respond to human states, actions, intentions and emotions. HRI related topics can include areas such as improving robot social acuity using visual perception of the person~\mbox{
\cite{socialerrorsHRI}}\hspace{0pt}
. Human-robot collaboration (HRC) instead focuses on how humans and robots work together to achieve shared goals with a common purpose and directed outcome. In HRC, robots work to complement or add value to the intended goal of the human \mbox{
\cite{bauer2008human}}\hspace{0pt}
. Collaboration with a robot can help to improve task speed; work productivity; human safety to minimise repetition fatigue, injuries; and reduce the number of errors~\mbox{
\cite{VASCONEZ201935,HaddadinSami2009RfSR}}\hspace{0pt}
. Specific tasks within HRC that can be improved by visual information include human-robot object handover~\mbox{
\cite{8955665,9206048}}\hspace{0pt}
, and human-robot co-ordination in shared spaces~\mbox{
\cite{VillaniValeria2018Sohc, robla2017working}}\hspace{0pt}
}\DIFdelend \DIFaddbegin \DIFadd{collaboration}\DIFaddend . This includes robotic vision with the intention to improve robot functionality, user experience, interface design, control methods, and robot utility for certain actions or tasks. 

\section{Review Protocol}\label{protocol}
 \DIFaddbegin \DIFadd{ This survey and systematic review will provide insight into the underlying emergent research themes pursued by the community, and explore the broader use of robotic vision to enhance human-robot interactivity and collaborative outcomes. The purpose of this systematic review is to inform readers about the current state of robotic vision applied to interpreting and responding to human actions, activities, tasks, states, and emotions. For the purpose of this survey and review, a robot was defined as a system that can perform (semi-)autonomously through an algorithm/s, action through actuator/s in the world in response to perception through sensor/s, with the potential inclusion of an externally provided goal directive. 
}

\DIFadd{This systematic review }\DIFaddend protocol followed the Preferred Reporting Items for Systematic Reviews and Meta-Analyses (PRISMA) \DIFdelbegin \DIFdel{guidelines for }\DIFdelend \DIFaddbegin \DIFadd{methodology for systematic research, which specifies the}\DIFaddend search, screening, and evaluation \DIFaddbegin \DIFadd{steps. This method involves a comprehensive and reproducible search strategy to present and critically appraise the research findings related to the topic of interest}\DIFaddend ~\cite{moher2009preferred}. \DIFdelbegin \DIFdel{The }\DIFdelend PRISMA guidelines~\cite{moher2009preferred} are considered to be a gold-standard reporting method with over 80,000 citations in the last 15 years. \DIFdelbegin \DIFdel{. It involves a standard checklist and reporting methodology for systematic research evaluation, allowing }\DIFdelend \DIFaddbegin \DIFadd{ The PRISMA method also allows for }\DIFaddend clear conclusions to be drawn across a \DIFdelbegin \DIFdel{large }\DIFdelend \DIFaddbegin \DIFadd{expansive }\DIFaddend pool of studies with minimal selection bias \DIFaddbegin \DIFadd{alongside balanced reporting of research findings}\DIFaddend ~\cite{moher2009preferred}. \DIFdelbegin \DIFdel{It is intended that the review will inform readers about the current state of robotic vision applied to interpreting and responding to human actions, activities, tasks, states, and emotions. This review explored }\DIFdelend \DIFaddbegin \DIFadd{All of the studies were assessed for inclusion/exclusion criteria to provide a defined view of the topic, as well as to assess the included studies for quality assurance. The search strategy of the systematic review was designed to capture }\DIFaddend different aspects related to robotic vision \DIFdelbegin \DIFdel{, including }\DIFdelend \DIFaddbegin \DIFadd{for human-robot interaction and collaboration, including robot behaviours, }\DIFaddend collaborative tasks and communicative behaviors. This review \DIFdelbegin \DIFdel{includes papers from the last 10 years that described robots using robotic vision with a mobile, manipulation and/or visual communication capacity on a task that had a direct benefit to the person's work or lifestyle. This review }\DIFdelend was designed to answer these research questions (RQ):

\begin{enumerate}[label={RQ\arabic*.}]
    \item What is the general trend of robotic vision \DIFaddbegin \DIFadd{work }\DIFaddend in human-robot collaboration and interaction in the last 10 years?
    \item What are the most common application areas and domains for robotic vision in human-robot collaboration and interaction?
    \item What is the \DIFdelbegin \DIFdel{general taxonomy for }\DIFdelend \DIFaddbegin \DIFadd{human-robot interaction taxonomy for robots with }\DIFaddend robotic vision in human-robot collaboration and interaction?
    \item What are the vision techniques and tools used in \DIFdelbegin \DIFdel{human-robot }\DIFdelend \DIFaddbegin \DIFadd{human--robot }\DIFaddend collaboration and interaction\DIFdelbegin \DIFdel{and what control techniques are used to link visual perception to robotic action}\DIFdelend ?
    \item What are the data sets and models that have been used for robotic vision in human-robot collaboration and interaction? 
    \item What has been the main participant sample, and how is robotic vision in human-robot collaboration and interaction evaluated?
    \item What is the state-of-the-art in vision algorithm performance for robotic vision in human-robot collaboration and interaction?
    \item What are the upcoming challenges for robotic vision in human-robot collaboration and interaction?
\end{enumerate}

\DIFdelbegin \DIFdel{Databases chosen for search included IEEE Xplore, ACM Library, and Scopus.Initial scans } Preliminary searches \DIFaddend were undertaken in \DIFdelbegin \DIFdel{relevant }\DIFdelend \DIFaddbegin \DIFadd{field-relevant }\DIFaddend journals and conferences to \DIFdelbegin \DIFdel{inform }\DIFdelend \DIFaddbegin \DIFadd{help inform search }\DIFaddend criteria keywords. \DIFdelbegin \DIFdel{Search terms }\DIFdelend \DIFaddbegin \DIFadd{The search terms went through extensive iteration and the final terms }\DIFaddend were chosen to be broad enough to capture \DIFdelbegin \DIFdel{papers }\DIFdelend \DIFaddbegin \DIFadd{works }\DIFaddend across multiple disciplines \DIFaddbegin \DIFadd{and topic keywords}\DIFaddend , but scoped \DIFdelbegin \DIFdel{to the }\DIFdelend \DIFaddbegin \DIFadd{as best as possible to systematically extract papers on the intended }\DIFaddend topic of interest: \DIFdelbegin \DIFdel{robotics}\DIFdelend \DIFaddbegin \DIFadd{robots}\DIFaddend , vision, and humans: (((("Abstract": Robot* OR "Abstract": UAV OR "Abstract": AUV OR "Abstract": UUV OR "Abstract": Drone OR "Abstract":Humanoid OR "Abstract":Manipulator) AND ("Abstract":Vision OR Abstract":Image OR "Abstract":Camera OR "Abstract":RGB* OR "Abstract":Primesense OR "Abstract":Realsense OR "Abstract":Kinect) AND ("Abstract":Human OR "Abstract":\DIFaddbegin \DIFadd{Humans OR "Abstract":}\DIFaddend Person OR "Abstract":\DIFdelbegin \DIFdel{User}\DIFdelend \DIFaddbegin \DIFadd{People OR "Abstract":User OR "Abstract":Users}\DIFaddend ) AND ("Abstract":HRI OR "Abstract":HRC OR "Abstract":Collaborat* OR "Abstract":Interact* OR "Abstract":"Human-in-the-Loop" OR "Abstract":Team* OR "Abstract":"Human-to-Robot"  OR "Abstract":"Robot-to-Human")))). \DIFaddbegin \DIFadd{To create the search method, the following databases were chosen for systematic search and data extraction, representing  multi-disciplinary avenues for published works: IEEE Xplore, ACM Library, and Scopus. }\DIFaddend Inclusion and exclusion criteria were \DIFaddbegin \DIFadd{generated, }\DIFaddend reviewed and approved by subject matter experts \DIFdelbegin \DIFdel{from }\DIFdelend \DIFaddbegin \DIFadd{across }\DIFaddend robotics, human-robot interaction, and behavioural science to confirm keyword relevance to identify suitable papers \DIFdelbegin \DIFdel{. Inclusion criteria }\DIFdelend \DIFaddbegin \DIFadd{for the topic of interest, and to reduce the chance of extracting unrelated works. The final inclusion criteria markers }\DIFaddend were used in a sequential order \DIFaddbegin \DIFadd{when categorising extracted papers to determine its inclusion into this systematic review}\DIFaddend :

\begin{enumerate}[label={(C\arabic*)}]
    \item The research must include at least one physically embodied robot that can perceive through a vision system.
    \item The robot(s) must be capable of at least one closed-loop interaction or information exchange between the human and the robot(s), where the robot(s) vision system is utilised in the exchange, and a human is the focus of the vision system.
    \item The robot(s) must be able to make a decision and/or action based on visual input that is real-time or at least fast enough for an interactive channel to occur between the human and the robot (i.e. ~60 seconds). 
\end{enumerate} 

\DIFdelbegin \DIFdel{Exclusions included }\DIFdelend \DIFaddbegin \DIFadd{The purpose of C1 was to ensure that only physical robot systems with a vision system were analysed with digital avatars and software systems running on computers removed from analysis. The purpose of C2 was to ensure that robots were able to perceive visual information relevant to creating a robot signal, task or action based on the vision system, and that the robot could in fact perceive some or part of the person during the interaction or collaborative exchange. The purpose of C3 was to ensure that robots could perform a decision and/or action based on interpretation of the visual information, and the interaction exchange occurs without extended wait times. Taken together, these chosen criteria would ensure that the robot was acting on the visual information, that the human was classified and/or involved in the process, and the information and/or exchange was occurring in a functional amount of time for an interaction. 
}

\DIFadd{To maintain the proposed review theme of humans, robotic vision and interaction or collaboration, several exclusions were created and used in this review. Papers were excluded if they contained }\DIFaddend non-embodied agents \DIFaddbegin \DIFadd{that did not operate as a robotic system, such as having no actuation system and/or capacity to make or execute decisions }\DIFaddend (i.e. cameras, computers, smartphones, tele-operated devices, avatars)\DIFdelbegin \DIFdel{, robots with an open-loop interaction, robotic vision not being }\DIFdelend \DIFaddbegin \DIFadd{. Papers were also excluded if there was no physical or verbal robot action involved in the process as a result of processing visual information, as well as instances in which the robot could have been substituted with a camera on its own, a computer screen on its own, or another simple input signal such as using the robot as a speaker only. Given the clear focus on robotic vision, this review excluded papers that did not meet the criteria of a vision sensor (RGB, RGB-D) paired with an algorithm/s that could translate raw images to a control signal for a robot. Papers were also excluded if vision was not }\DIFaddend central to the system's operation \DIFdelbegin \DIFdel{, and involved simple devices (i. 
e. children's toys). Research }\DIFdelend \DIFaddbegin \DIFadd{or lacked control, such as vision being a function of the robot, but not used to inform or update the decision-making process or resulting action of the robot. 
}

\DIFadd{Papers }\DIFaddend that did not have \DIFaddbegin \DIFadd{any }\DIFaddend human-relevant information\DIFdelbegin \DIFdel{or experimentation was not included }\DIFdelend \DIFaddbegin \DIFadd{, use case application or research experiment with people were also excluded }\DIFaddend because it did not meet inclusion criteria for \DIFdelbegin \DIFdel{investigating }\DIFdelend \DIFaddbegin \DIFadd{the intention to explore robotic vision in relation to }\DIFaddend human-robot interaction or collaboration. \DIFdelbegin \DIFdel{This included }\DIFdelend \DIFaddbegin \DIFadd{Examples of these papers include }\DIFaddend early-stage design work \DIFdelbegin \DIFdel{and competition papers. Papers were also excluded if there was no robot action involved in the process, and instances in which the robot could have been readily substituted with a camera, computer screen, or other simple input signal. 
Papers }\DIFdelend \DIFaddbegin \DIFadd{on proposed concepts of robot systems that had not yet been built, and robot competition papers where the robot was intended for a human environment, but its proposed performance or relationship with people was not reported at all. For this review, only papers in which there was a clear interaction or collaboration between the human and the robot were included. Therefore, robots that only used an open-loop interaction were excluded from analysis for not meeting the criteria for an interaction, especially if the visual signal input was independent of the robot output and did not influence the action or decision-making of the robot. Simple devices such as children's toys were also excluded given the limited interaction set often involved in these devices, and the intention to focus on robots that could provide benefit to support a persons work or lifestyle. 
}

\DIFadd{All papers }\DIFaddend must have been published \DIFdelbegin \DIFdel{between }\DIFdelend \DIFaddbegin \DIFadd{and available for access from the publication venue between 1st January }\DIFaddend 2010 to \DIFaddbegin \DIFadd{31st December }\DIFaddend 2020 in a peer-reviewed journal or conference\DIFaddbegin \DIFadd{. Papers that were not formally published between these dates were not extracted}\DIFaddend . If authors published multiple versions \DIFaddbegin \DIFadd{of the work}\DIFaddend , the most complete version was \DIFdelbegin \DIFdel{reviewed}\DIFdelend \DIFaddbegin \DIFadd{included}\DIFaddend . E-Print services (e.g. arXiv.org) were not included for three reasons: 1) \DIFaddbegin \DIFadd{the }\DIFaddend abundance of early stage work \DIFaddbegin \DIFadd{that did not yet include humans into the proposed system}\DIFaddend , 2) \DIFdelbegin \DIFdel{no quality control }\DIFdelend \DIFaddbegin \DIFadd{there was limited quality control without a peer-review process to ensure that only high-quality papers were identified in an unbiased way}\DIFaddend , and 3) \DIFdelbegin \DIFdel{reported work that did not meet peer review standards would }\DIFdelend \DIFaddbegin \DIFadd{reporting on early-stage work that has not yet undergone peer review could }\DIFaddend have created a skewed commentary on the \DIFdelbegin \DIFdel{field. 
}\DIFdelend \DIFaddbegin \DIFadd{current prevalence and impact of the field. We do acknowledge the importance of robotic vision use cases that occurred in works that would have fallen into the excluded criteria category for this systematic review. As such, we will present a subsection below to acknowledge and investigate the use cases that did not meet the search strategy criteria, including any key papers that were not captured as part of the systematic search. Examples include tele-operated robots, robots with an open-loop system, simple devices and early-stage work that did not have tests with people. 
}\DIFaddend 

\subsection{Review Information and Categorisation}\label{category}
Each eligible article underwent systematic data extraction informed by robot classification and human-robot interaction taxonomies, e.g. \cite{yanco2004classifying, beer2014toward}. \DIFdelbegin \DIFdel{Categories included }\DIFdelend \DIFaddbegin \DIFadd{For each, manuscript information was extracted into categories such as }\DIFaddend task type, task criticality (low, medium, high), robot morphology (anthropomorphic, zoomorphic, functional), ratio of people to robots (i.e. a non-reduced fraction with number of humans over number of robots), composition of robot teams (formation), level of shared interaction among teams, interaction roles, type of human-robot physical proximity, time/space taxonomy \cite{yanco2004classifying}, level of autonomy \cite{beer2014toward}, task evaluation, sensor fusion (i.e. vision and speech), camera system and type, vision techniques and algorithm, training method and data sets. User study information was extracted if a user study was reported, including participant details and experimental outcomes. A custom metric for overall task evaluation was computed using 3-point scaling (low, medium, high) for task complexity, risk, importance, and robot complexity. 

\DIFaddbegin \DIFadd{Application areas were clustered and labelled using the following criteria. Gestures were defined as a hand, arm, head or body movement intended to indicate, convey a message or send information. Action recognition was defined as the recognition of human actions or activities that were not related to explicit gestures. Robot movement in human spaces was classified if the robot had physical movement in a human environment and robot movement did not require the human to perform a set pose to signal movement commands to the robot, including if the robot was classified as a (semi-)autonomous vehicle. Object handover and collaborative action papers included a robot capable of manipulating objects while the interaction did not require the person to perform a set pose (i.e the person was detected without performing a gesture or action). Categorization for social communication captured papers in which the robot needed to perform a social behaviour, or be capable of socially interacting with a person. Lastly, learning from demonstration must have used some form of demonstration learning. 
}

\DIFaddend \section{Results}\label{results}
\subsection{Selected Articles}\label{resultselected}
\begin{figure}[!ht]
\centering
\includegraphics[width=0.7\columnwidth]{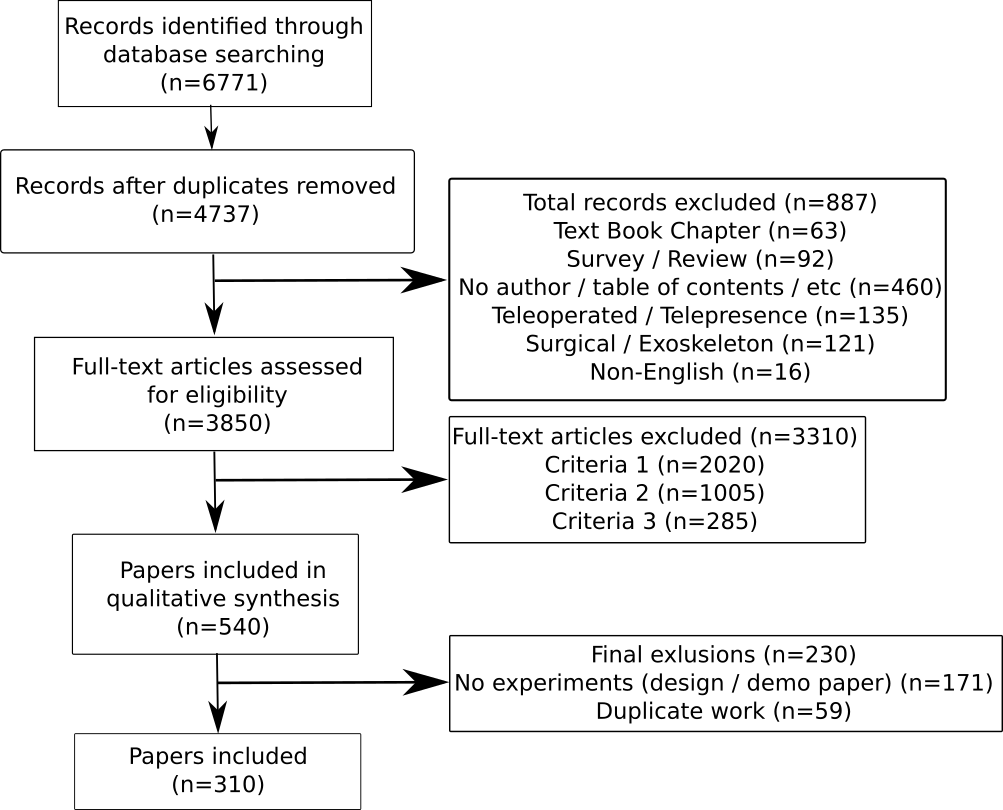}
\hfil
\caption{CONSORT Chart for Systematic Review to Determine Inclusion\DIFdelbeginFL \DIFdelFL{in the Survey}\DIFdelendFL }
\label{fig:consort}
\vspace{-3mm}
\end{figure}

The initial search \DIFdelbegin \DIFdel{found 6607 papers, 1991 }\DIFdelend \DIFaddbegin \DIFadd{across three databases found 6771 papers, 2034 }\DIFaddend of which were \DIFdelbegin \DIFdel{duplicate records}\DIFdelend \DIFaddbegin \DIFadd{identified as duplicate records (See Figure~\ref{fig:consort})}\DIFaddend . The remaining \DIFdelbegin \DIFdel{4616 }\DIFdelend \DIFaddbegin \DIFadd{4737 }\DIFaddend papers were screened for titles and abstracts to assess initial eligibility \DIFdelbegin \DIFdel{with 871 }\DIFdelend \DIFaddbegin \DIFadd{and 887 }\DIFaddend papers excluded based on format: textbook chapters that did not include original \DIFaddbegin \DIFadd{research }\DIFaddend work (n = 63, 7\%), reviews or surveys (n = \DIFdelbegin \DIFdel{89}\DIFdelend \DIFaddbegin \DIFadd{92, 10\%}\DIFaddend ), no English version \DIFaddbegin \DIFadd{of the work }\DIFaddend (n = 16, 2\%), and other non-related works (n = \DIFdelbegin \DIFdel{448}\DIFdelend \DIFaddbegin \DIFadd{460, 52\%}\DIFaddend ) such as front pages, table of contents, plenary talks, copyright notices, and keynotes. A total of 255 \DIFdelbegin \DIFdel{were excluded on title: surgical tools }\DIFdelend \DIFaddbegin \DIFadd{papers were excluded based on title alone: robotic surgical tools only }\DIFaddend (n = \DIFdelbegin \DIFdel{120) and teleoperation }\DIFdelend \DIFaddbegin \DIFadd{121, 14\%) and tele-operation only }\DIFaddend (n = 135, 15\%). The remaining \DIFdelbegin \DIFdel{3745 }\DIFdelend \DIFaddbegin \DIFadd{3850 papers }\DIFaddend were assessed by detailed review of the text and \DIFdelbegin \DIFdel{3217 }\DIFdelend \DIFaddbegin \DIFadd{3310 }\DIFaddend papers omitted for not meeting inclusion criteria 1-3: \DIFdelbegin \DIFdel{1974 }\DIFdelend \DIFaddbegin \DIFadd{2020 }\DIFaddend on C1 (61\%), \DIFdelbegin \DIFdel{965 }\DIFdelend \DIFaddbegin \DIFadd{1005 }\DIFaddend on C2 (30\%) and \DIFdelbegin \DIFdel{278 }\DIFdelend \DIFaddbegin \DIFadd{285 }\DIFaddend on C3 (9\%). \DIFdelbegin \DIFdel{Papers }\DIFdelend \DIFaddbegin \DIFadd{The large volume of papers }\DIFaddend omitted on C1 \DIFdelbegin \DIFdel{often }\DIFdelend \DIFaddbegin \DIFadd{showed that most works }\DIFaddend had no physical robot, e.g.~\cite{ueno2014efficient, zhao2016intuitive, hacinecipoglu2020pose, mcfassel2018prototyping}, \DIFaddbegin \DIFadd{involved }\DIFaddend simulation testing such as \DIFaddbegin \DIFadd{using }\DIFaddend a virtual robot, e.g.~\cite{yamamoto20112d, papadopoulos2019advanced, du2014markerless, ben2016kinect}, \DIFaddbegin \DIFadd{involved }\DIFaddend cameras on their own \DIFaddbegin \DIFadd{that were not connected to robotic systems }\DIFaddend e.g.~\cite{moladande2019implicit}, or \DIFdelbegin \DIFdel{robots that }\DIFdelend \DIFaddbegin \DIFadd{the robot/s }\DIFaddend did not use a vision system \DIFaddbegin \DIFadd{as part of the interaction with the person }\DIFaddend e.g.~\cite{pop2019control,mendes2017human}. Papers omitted on C2 often had a \DIFaddbegin \DIFadd{clear }\DIFaddend focus on other \DIFdelbegin \DIFdel{aspects }\DIFdelend \DIFaddbegin \DIFadd{components }\DIFaddend such as speech, e.g.~\cite{liu2018learning} or visual servoing, e.g.~\cite{jiang2018personalize}. C2 papers often had no humans at all e.g.~\cite{del2011interaction}, or no human vision involved in the vision process of the interaction or collaboration, e.g.~\cite{haddadin2011towards, randelli2013knowledge,cicconet2013human, khatib2017visual, yamakawa2018human}. Papers omitted on C3 often did not have a robot perform an action based on visual input e.g.~\cite{tornow2013multi, mckeague2013hand, zhang2013real, massardi2020parc, gonzalez2020audiovisual}, robots that were \DIFdelbegin \DIFdel{teleoperated }\DIFdelend \DIFaddbegin \DIFadd{tele-operated }\DIFaddend e.g.~\cite{rehm2013negative} or no near real-time information exchange, e.g.~\cite{chen2020human}. A total of 540 papers that met \DIFdelbegin \DIFdel{full inclusion . Records were further }\DIFdelend \DIFaddbegin \DIFadd{the C1-3 inclusion criteria were then subject to another round of investigation. Papers were then }\DIFaddend excluded if it was the same study published across multiple venues (n = 59, 26\%), or there were no clear \DIFdelbegin \DIFdel{experiments or demonstrations }\DIFdelend \DIFaddbegin \DIFadd{experiment or demonstration }\DIFaddend of human-robot interaction or collaboration \DIFaddbegin \DIFadd{despite reporting on a system designed for human-robot interaction and collaboration }\DIFaddend (n = 171, 74\%). A final total of \DIFdelbegin \DIFdel{303 papers (4.59}\DIFdelend \DIFaddbegin \DIFadd{310 papers (8}\DIFaddend \% of full articles assessed for eligibility) met final inclusion criteria, which provides a significant pool of \DIFdelbegin \DIFdel{records }\DIFdelend \DIFaddbegin \DIFadd{research works }\DIFaddend for detailed analysis \DIFaddbegin \DIFadd{on the chosen topic}\DIFaddend . The CONSORT chart of inclusion and exclusion steps can be seen at Figure~\ref{fig:consort}. Two independent raters went through 10\% of \DIFdelbegin \DIFdel{303 }\DIFdelend \DIFaddbegin \DIFadd{310 }\DIFaddend eligible papers and achieved a 100\% consensus on inclusion and exclusion criteria. \DIFaddbegin \DIFadd{Some notable works that would have fallen into the excluded criteria category for this systematic review were reported in a separate subsection as part of presenting a comprehensive survey on the topic, but these papers were not included in the final systematic review. 
}\DIFaddend 

\section{RQ1. What is the general overall trend of robotic vision in human-robot collaboration and interaction in the last 10 years?}\label{sec:rq1}

\begin{figure}[ht!]
\centering
\subfloat[Total Annual Published Papers]{\includegraphics[width=0.45\columnwidth]{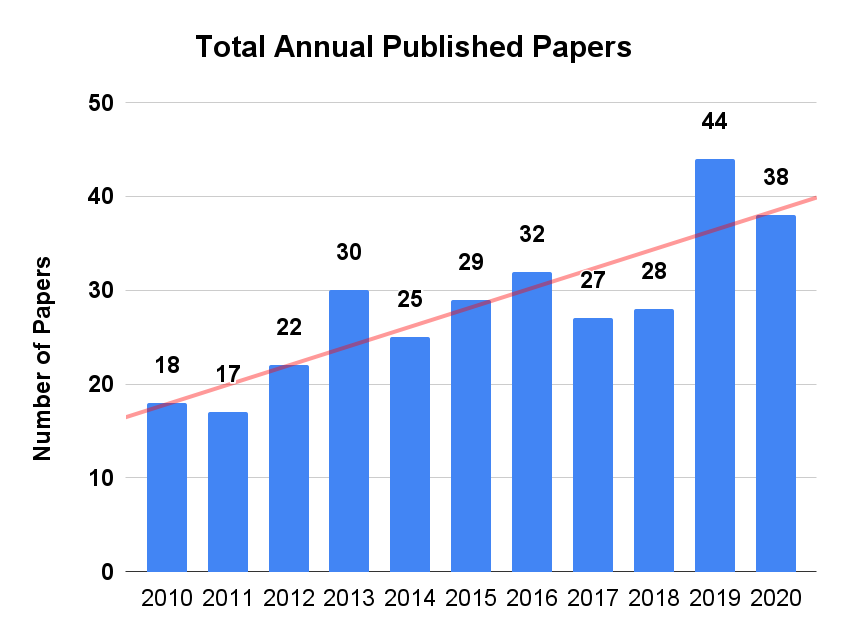} \label{fig:total_annual}}
\hfil
\subfloat[Number of Papers Published Across Themes]{\includegraphics[width=0.45\columnwidth]{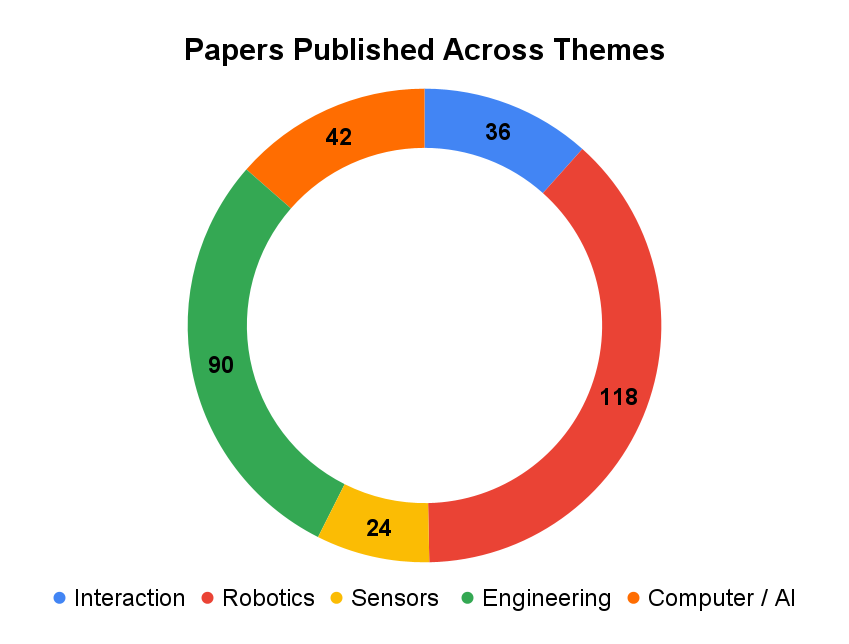}\label{fig:total_theme}}
\hfil
\subfloat[Number of Published Papers in Each Venue]{\includegraphics[width=0.45\columnwidth]{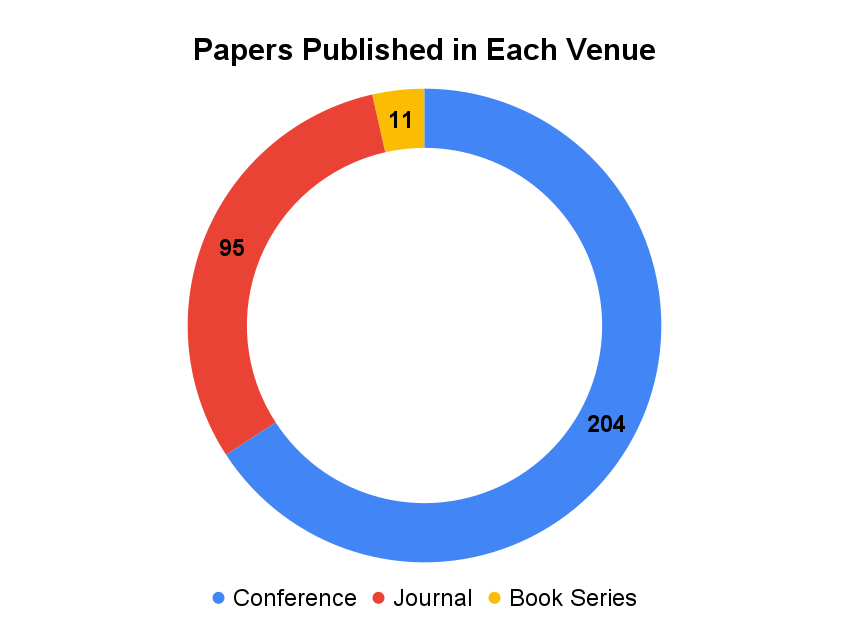}\label{fig:total_venue}}
\subfloat[Application Area Total]{\includegraphics[width=0.45\columnwidth]{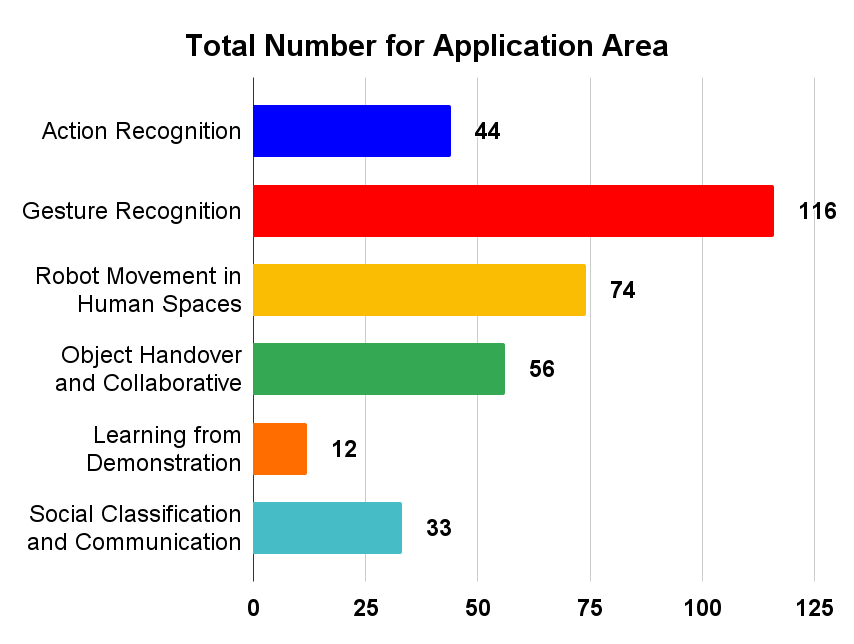}\label{fig:total_application}}
\hfil
\subfloat[Domain Area Total]{\includegraphics[width=0.45\columnwidth]{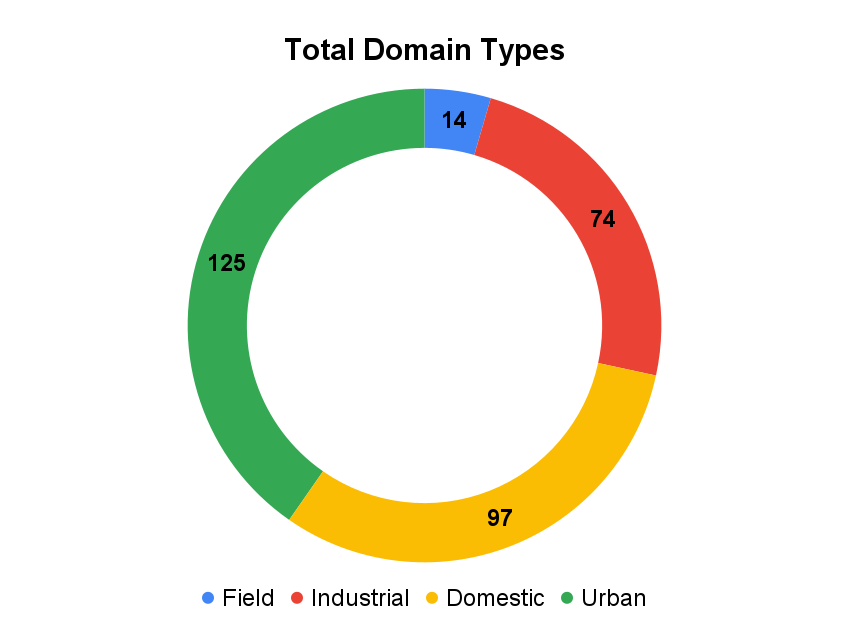}\label{fig:total_domain}}
\hfil
\subfloat[Robot Type Total]{\includegraphics[width=0.45\columnwidth]{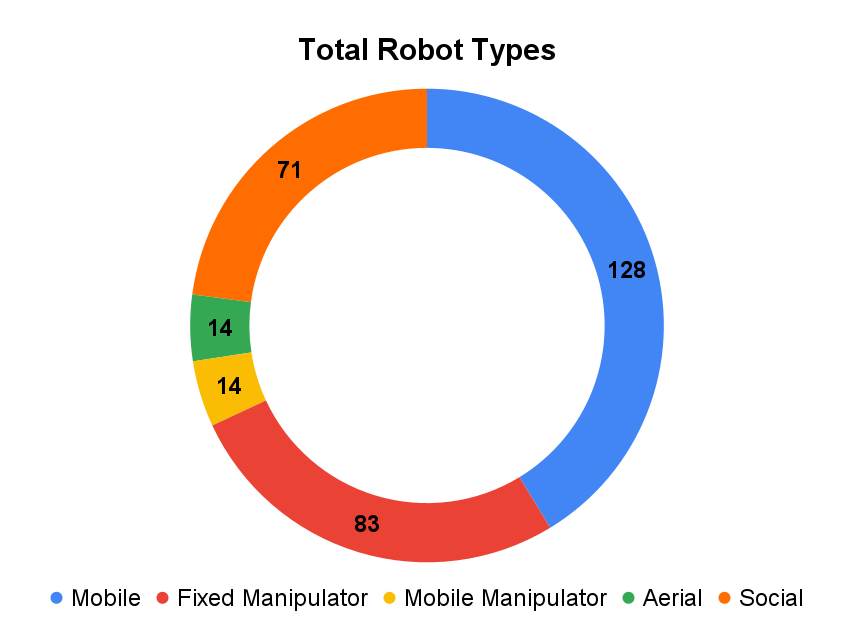}\label{fig:total_robot}}
\hfil
\caption{Statistical Summaries of Total Paper Counts and Trends}
\label{fig:total_all}
\end{figure}

This section presents general trend of robotic vision in human-robot collaboration and interaction in the last 10 years. \DIFaddbegin \DIFadd{Robotic vision for HRI/C had a moderate but steady increase, which might be attributed to several components, such as limited accessibility to robot platforms, integration challenges, the interdisciplinary nature of HRI/C, technical capacity for robots to operate consistently for robust use cases with people, limited engineering knowledge of human-robot testing, limited capacity to test robots in human spaces, and human-centred robotics representing a much smaller research field compared to its robotics and computer vision counterparts. }\DIFaddend Figure~\ref{fig:total_annual} depicts a modest increase in publications over the period, but a small decline from 2020/2021 which was predicted to be attributed to the COVID-19 pandemic. Figure~\ref{fig:total_theme} depicts the publication themes of robotic vision work, including interaction (human-robot interaction, human-machine systems), robotics (robotics, automation, mechatronics), sensors (sensors, vision, signal processing), engineering (engineering, systems, industry, control, science), and computers and artificial intelligence (AI). Figure~\ref{fig:total_venue} depicts the most relevant papers were published in conferences, journals and then book series. 

\begin{figure}[h!]
\centering
\hfil
\subfloat[Total Camera Types - Bar Chart Totals over Time]{\includegraphics[width=0.5\columnwidth]{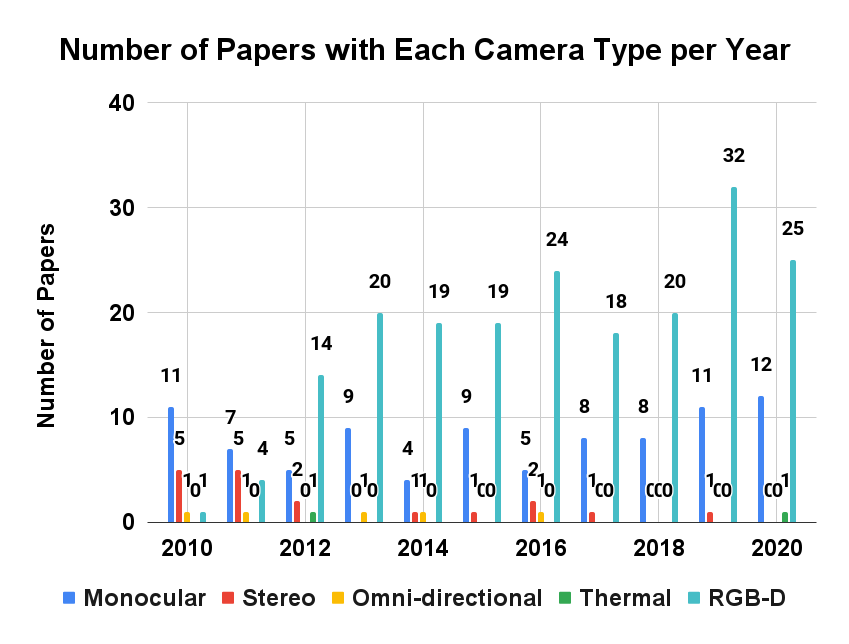}\label{fig:total_camera}}
\hfil
\subfloat[Total Camera Type - 100\% Stacked Area]{\includegraphics[width=0.5\columnwidth]{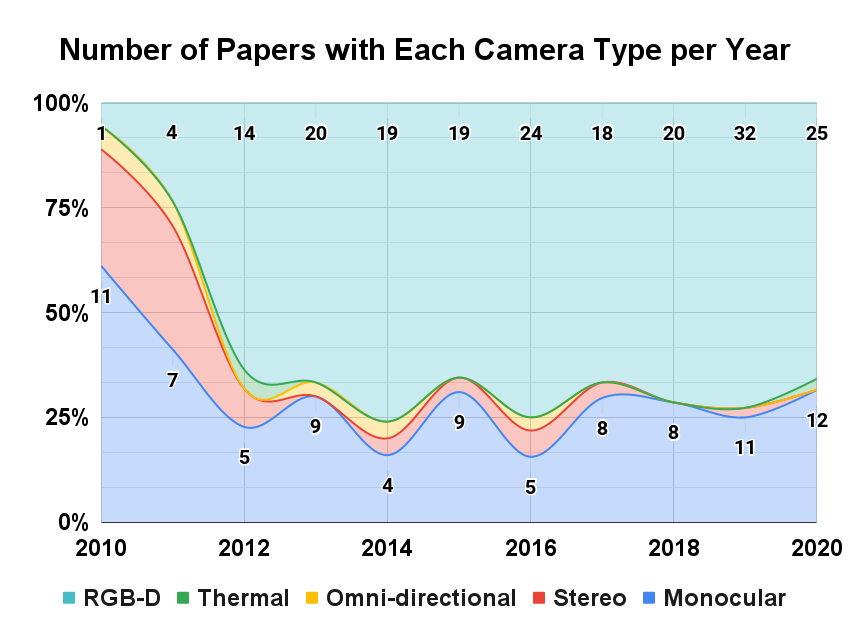}\label{fig:total_camerastack}}
\hfil
\subfloat[Camera Type and Robot Type]{\includegraphics[width=0.45\columnwidth]{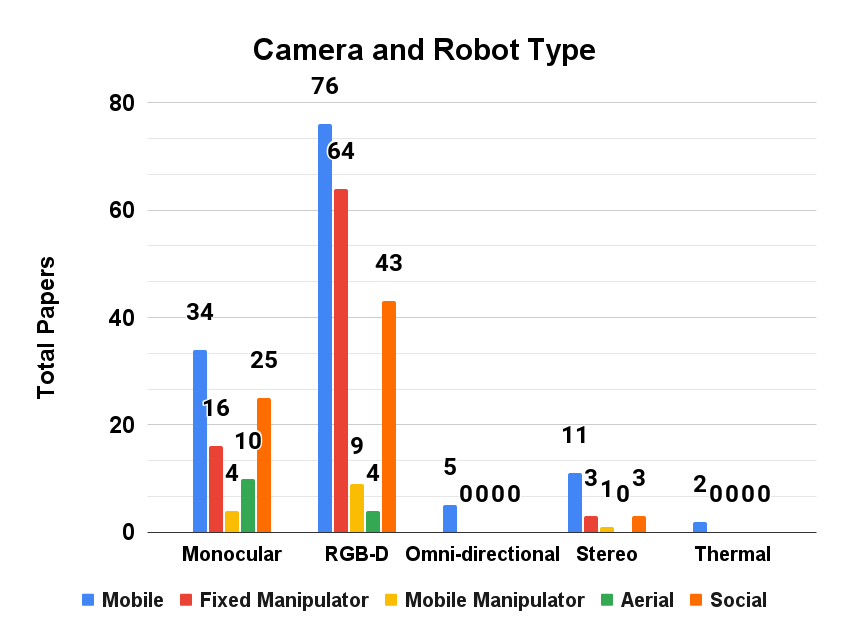}\label{fig:total_camrobot}}
\hfil
\subfloat[Camera Type and Domain]{\includegraphics[width=0.45\columnwidth]{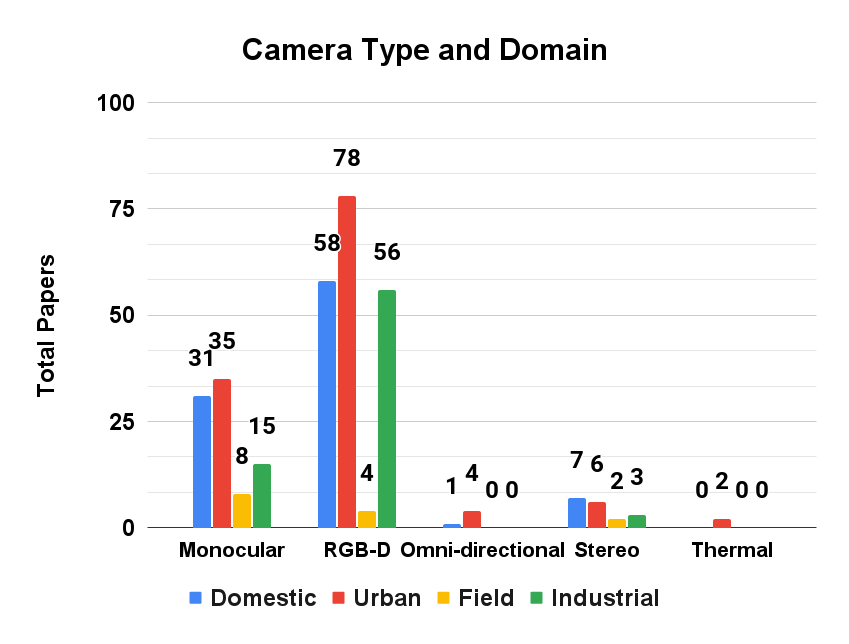}\label{fig:total_camdomain}}
\caption{Statistical Summary of Application, Domain and Robot Types}
\label{fig:papers_per_year}
\vspace{-3mm}
\end{figure}

\begin{figure}[hbt!]
\centering
\subfloat[Application Area]{\includegraphics[width=0.47\columnwidth]{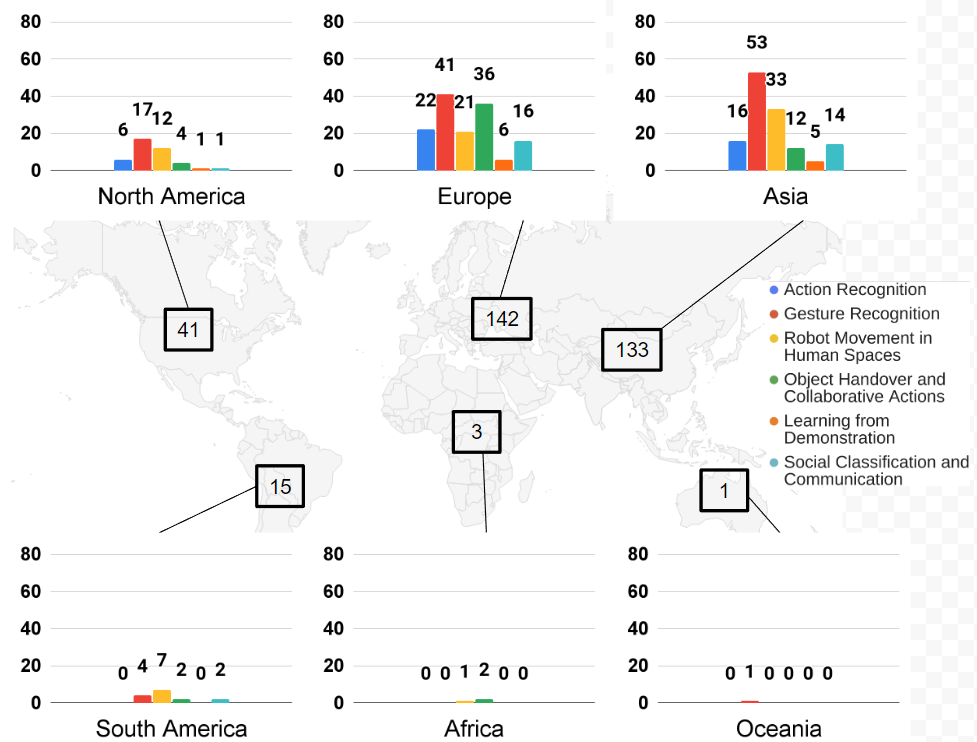}\label{fig:total_appglobal}}
\hfil
\subfloat[Robot Type]{\includegraphics[width=0.45\columnwidth]{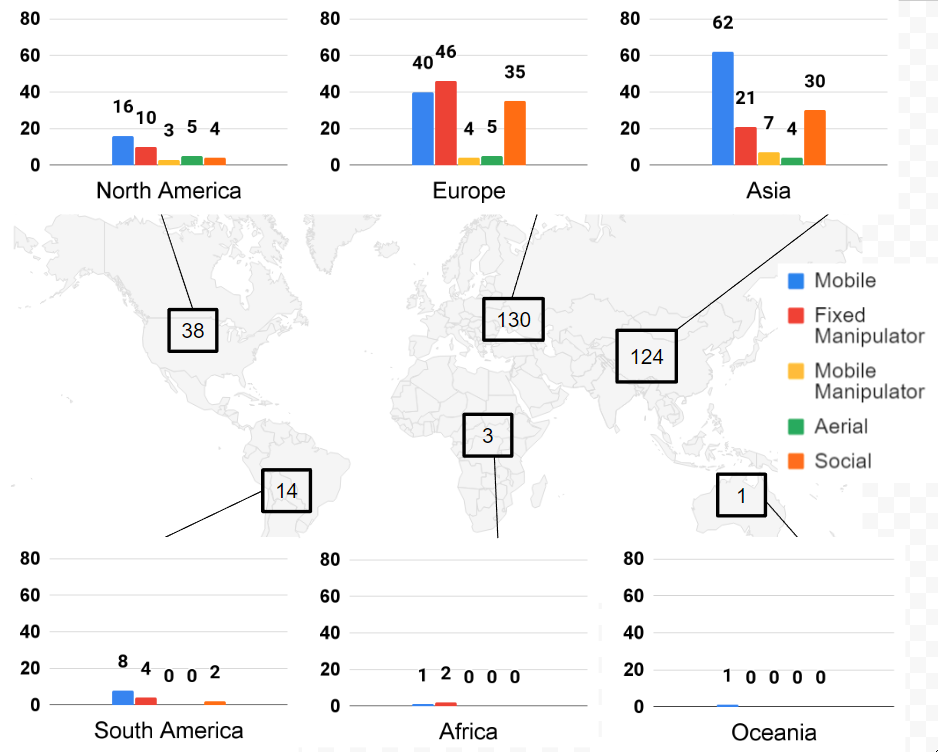}\label{fig:total_robotglobal}}
\vspace{-3mm}
\hfil
\caption{Statistical Summaries of Total Paper Counts and Trends per Continents}
\label{fig:papers_per_continent}
\vspace{-3mm}
\end{figure}

Figure~\ref{fig:total_application} depicts the most common application areas clustered into groups (\DIFdelbegin \DIFdel{$N=328$}\DIFdelend \DIFaddbegin \DIFadd{$N=335$}\DIFaddend ): action recognition (13\%), gesture recognition (\DIFdelbegin \DIFdel{36}\DIFdelend \DIFaddbegin \DIFadd{35}\DIFaddend \%), robot movement in human spaces (\DIFdelbegin \DIFdel{21}\DIFdelend \DIFaddbegin \DIFadd{22}\DIFaddend \%), object handover and collaborative actions (17\%), learning from demonstration (3\%) and social communication (\DIFdelbegin \DIFdel{9}\DIFdelend \DIFaddbegin \DIFadd{10}\DIFaddend \%). If a paper had more than one application, each area was included in the final total. Individual application breakdowns will be seen in the next section. Figure~\ref{fig:total_domain} depicts that common domain areas involved field, industrial (i.e. manufacturing and warehouses), domestic (i.e. home use), and urban settings (i.e. shopping centres, schools, restaurants, and hotel). \DIFaddbegin \DIFadd{Robots that work with and around humans were often proposed for domestic and urban environments. }\DIFaddend Figure~\ref{fig:total_robot} depicts common robot types being mobile robots, followed by fixed manipulators, social robots, mobile manipulators and aerial robots. If multiple robots were tested, only the first or most detailed test was reported in the total. \DIFaddbegin \DIFadd{Most works had a single focus on a specific vision application for a target purpose, and the intended outcome was often for robots to better integrate into human-populated environments in a direct (i.e. controlled via gesture) or non-direct way (i.e. following a person).

}\DIFaddend Figure~\ref{fig:total_camera} depicts that for camera type, RGB-D cameras such as the Kinect were the most frequently used, followed by monocular, stereo and omni-directional cameras. Figure~\ref{fig:total_camerastack} shows an increased uptake of RGB-D cameras. \DIFaddbegin \DIFadd{RGB-D cameras were extensively used across all use cases, environments and robot types, showing the value of this sensor capacity to provide critical visual information to improve robotic vision for robot tasks. }\DIFaddend Figure~\ref{fig:papers_per_continent} depicts global trends in domain and types. Figure~\ref{fig:total_appglobal} depicts the highest volume of work was conducted in gesture recognition or robot movement in human spaces, with the exception of Europe with a higher focus on object handover and collaborative actions. Figure~\ref{fig:total_robotglobal} depicts that the most common robot types was mobile robots and fixed manipulators across all continents. 

\DIFdelbegin \DIFdel{Robotic vision for HRI/C had a moderate but steady increase, which might be attributed to limited accessibility to robot platforms, challenges for robots to operate consistently for robust use cases with people, limited engineering laboratory knowledge of human-robot testing, limited capacity to test robots in human spaces, and human-centred robotics representing a much smaller research field compared to its robotics and computer vision counterparts. Robots that work with and around humans were often proposed for domestic and urban environments, which corresponds to where robots are most likely to bring value or support to the lives of humans. Most works had a single focus on a specific vision application for a target purpose, and the intended outcome was often for robots to better integrate into human-populated environments in a direct (i.e. controlled via gesture) or non-direct way (i.e. following a person). RGB-D cameras were extensively used across all use cases, environments and robot types, showing the value of this sensor capacity to provide critical visual information to improve robotic vision for robot tasks. 
}

\clearpage 
\section{RQ2. What are the most common application areas and domains for robotic vision in human-robot collaboration and interaction?}\label{sec:rq2}

This section provides a detailed breakdown of the following application areas: gesture and action recognition, robot movement in human spaces, object handover and collaborative actions, learning from demonstration, and social communication. If a paper had more than one application, each area was included in the final total. \DIFdelbegin \DIFdel{Published works often described the system architecture and the development process rather than system testing outside of laboratory settings and populations. Studies often had }\DIFdelend \DIFaddbegin \DIFadd{Papers often reported }\DIFaddend tasks and actions \DIFdelbegin \DIFdel{were often }\DIFdelend \DIFaddbegin \DIFadd{that were }\DIFaddend simplified or well-contained in \DIFdelbegin \DIFdel{a specific }\DIFdelend \DIFaddbegin \DIFadd{their relevant }\DIFaddend context or domain. \DIFdelbegin \DIFdel{There were fewer use case examples that included tasks or actions that could create high utility for the human, such as robots to support joint actions around household cleaning or more complex industrial tasks that require multiple actions. }\DIFdelend In addition, \DIFdelbegin \DIFdel{examples often had a focus on involving }\DIFdelend \DIFaddbegin \DIFadd{papers often focused on using }\DIFaddend the human to improve the robot's performance, such as \DIFdelbegin \DIFdel{gesture for greater control over }\DIFdelend \DIFaddbegin \DIFadd{human gestures for more control over the }\DIFaddend robot actions, \DIFdelbegin \DIFdel{improvement of }\DIFdelend \DIFaddbegin \DIFadd{humans to improve }\DIFaddend robot handover accuracy, and \DIFdelbegin \DIFdel{safety for mobile robot pathways. This could mean improving robotic vision for robots to more effectively operate around more people for a longer period of time in unpredictable scenarios, increasing their potential utility to the individual person and society. Therefore, there is a need to further develop robotic vision for robots for more diverse tasks with more comprehensive evaluations in relation to the person. In general, there was significant opportunity to draw from principles and concepts of computer vision to improve robot capacity to perceive and act upon visual information to improve human-robot collaboration. }\DIFdelend \DIFaddbegin \DIFadd{humans contributing to better mobile robot safety on pathways. }\DIFaddend A summary of the identified papers will be reviewed in Section~\ref{sec:rq2} and detailed exploration into the technical content of the papers discussed in Section~\ref{sec:rq7}. 

\subsection{Gesture Recognition}\label{sec:rq2:gesture}

\subsubsection{Overview}
\begin{figure}[ht!]
\centering
\subfloat[Total Papers]{\includegraphics[width=0.45\columnwidth]{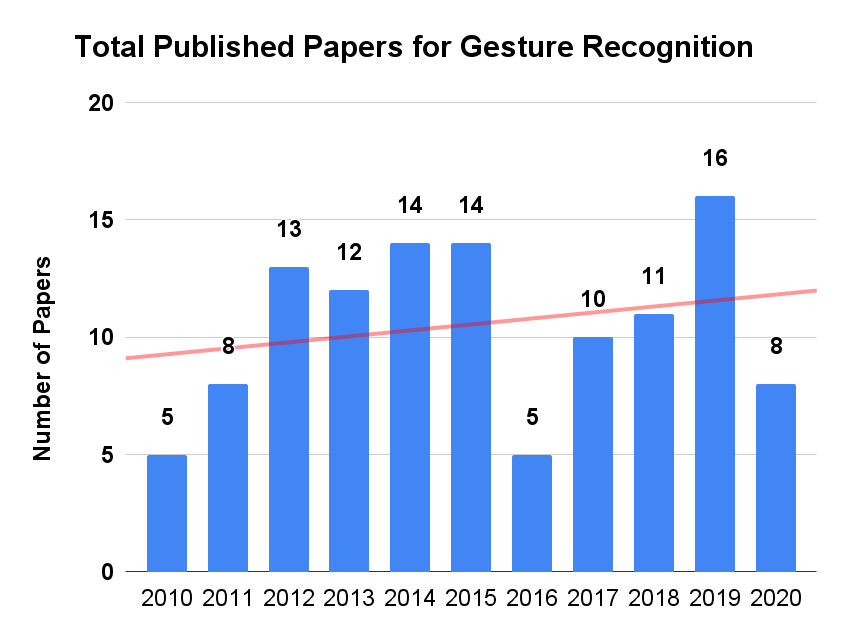}\label{fig:gesture:annual}}
\hfil
\subfloat[Domains]{\includegraphics[width=0.45\columnwidth]{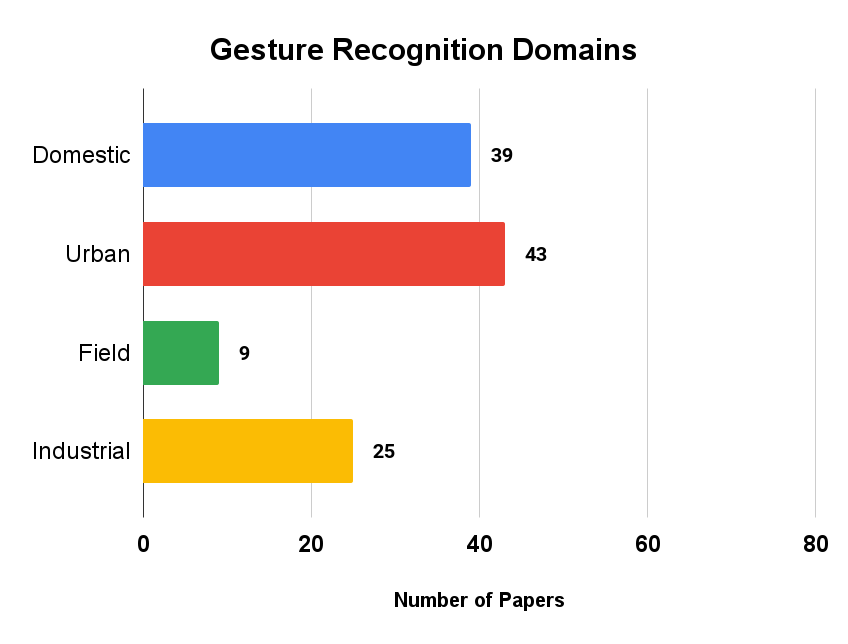}\label{fig:gesture:domainl}}
\hfil
\subfloat[Camera Types]{\includegraphics[width=0.45\columnwidth]{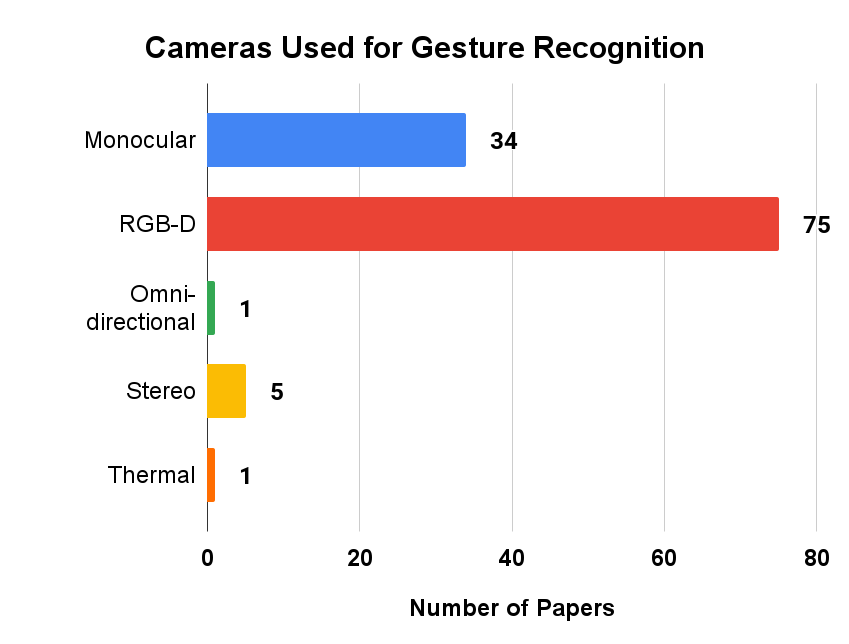}\label{fig:gesture:camera}}
\hfil
\subfloat[Robot Types]{\includegraphics[width=0.45\columnwidth]{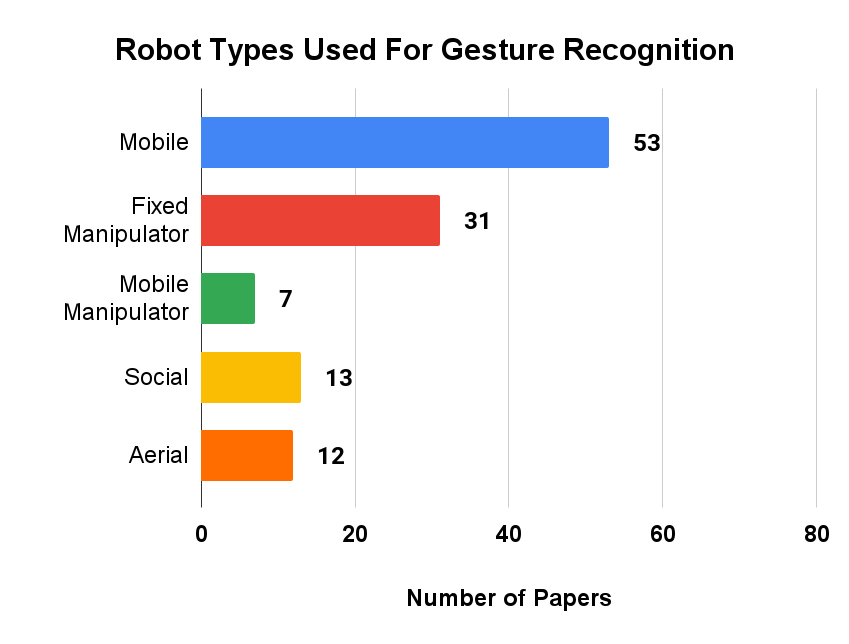}\label{fig:gesture:robot}}
\hfil
\subfloat[Gesture Types]{\includegraphics[width=0.45\columnwidth]{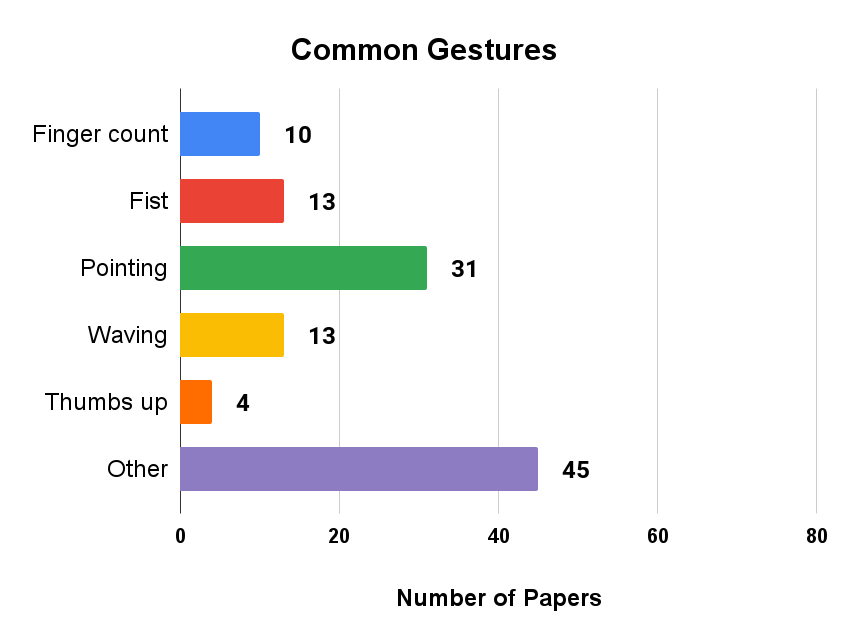}\label{fig:gesture:type}}
\hfil
\subfloat[Autonomy]{\includegraphics[width=0.45\columnwidth]{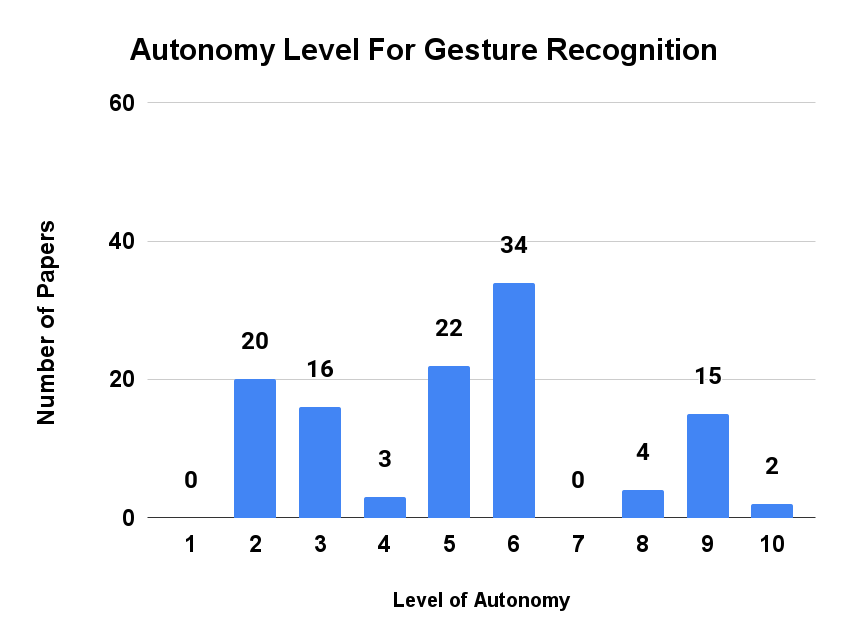}\label{fig:gesture:autonomy}}
\hfil
\caption{Gesture Recognition Totals and Summaries}
\label{fig:gesture_recognition}
\vspace{-3mm}
\end{figure}

Gestures were defined as a hand, arm, head or body movement intended to indicate, convey a message or send information. A total of \DIFdelbegin \DIFdel{117 papers (39}\DIFdelend \DIFaddbegin \DIFadd{116 papers (37}\DIFaddend \% of eligible total) were found to have at least one form of gesture recognition that used vision to identify and respond to the person. Figure~\ref{fig:gesture_recognition} depicts the number of gesture-related works, common domains, camera types, robot types, gesture types and level of autonomy. In 75 papers that used RGB-D cameras, 66 were the Kinect (88\%). From the total \DIFdelbegin \DIFdel{117 }\DIFdelend \DIFaddbegin \DIFadd{116 }\DIFaddend papers, 79 (68\%) involved gestures from body pose, and \DIFdelbegin \DIFdel{38 }\DIFdelend \DIFaddbegin \DIFadd{37 }\DIFaddend (32\%) involved a hand gesture. Most papers used static gestures (i.e., stationary human pose, $N=91$, 78\%) which did not require visual detection of movement. A smaller portion used dynamic methods (\DIFdelbegin \DIFdel{$N=26$}\DIFdelend \DIFaddbegin \DIFadd{$N=25$}\DIFaddend , 22\%), requiring multiple frames to classify the gesture. Some papers had a blended approach, for example, a static gesture to signify the start of a dynamic gesture, e.g.~\cite{xu_online_2014}. Gesture recognition to control robots was used for different robot types: industrial robot arms, e.g. ~\cite{fareed_gesture_2015, song_towards_2017,lima_real-time_2019, martin_real-time_2019}, mobile ground robots, e.g. ~\cite{zhang_interactive_2018, chen_human-following_2019, miller_self-driving_2019}, mobile manipulators, e.g. ~\cite{droeschel_towards_2011, canal_gesture_2015, moh_gesture_2019, sriram_mobile_2019}, and less commonly for social robots, e.g. ~\cite{kalidolda_towards_2018}, and aerial robots, e.g. ~\cite{lichtenstern_prototyping_2012}. In robot type, robots had little consistency across different categories with a wide variety of models used for gesture recognition. Continuous control was often used, such as to interact with mobile robots using hand position~\cite{paulo_vision-based_2012}, and small mobile robots with head positions~\cite{lam_real-time_2011}. Lastly, gestures were also used with teams of robots~\cite{milligan_selecting_2011,lichtenstern_prototyping_2012, nishiyama_human_2013, alonso-mora_human_2014, canal_gesture_2015}, and by multiple humans in the same scene~\cite{luo_tracking_2011}.    

\subsubsection{Use Case Examples: Mobile Robot Control}
In mobile ground robots, hand gestures were often used to control the robot to move forward, left, right, or stop, e.g. ~\cite{cicirelli_kinect-based_2015,fareed_gesture_2015,lalejini_evaluation_2015, ghandour_human_2017, long_kinect-based_2018, pentiuc_drive_2018, zhang_interactive_2018, chen_human-following_2019, miller_self-driving_2019, li_real-time_2019, chen_approaches_2010,manigandan_wireless_2010,faudzi_real-time_2012, lavanya_gesture_2018, waskito_wheeled_2020, xu_skeleton_2020}. Pointing gestures were often used to direct mobile robots to a specified location, e.g.~\cite{park_real-time_2011, van_den_bergh_real-time_2011, yoshida_evaluation_2011, abidi_human_2013, prediger_robot-supported_2014, chen_stereovision-only_2014, jevtic_comparison_2015, tolgyessy_foundations_2017}. 
Human body movements were used as the control signal, such as shoulder angle to control robot direction based on discrete angles~\cite{wang_wheeled_2013}. More dynamic motions were also used to control a robot to move forward, back, left and right by movement of an arm up and down, left to right, or in circles, e.g. ~\cite{fang_vehicle-mounted_2019}, or hand waving to signify a follow-me or goodbye command, e.g.~\cite{fujii_gesture_2014}. Body gestures were used to control an otherwise autonomous mobile robot to turn and stop by moving arms up or down~\cite{ghandour_human_2017}, or human shoulder position to control robot velocity~\cite{mao_medical_2018}. Lastly, a spherical vision system was used with a mobile robot with three omni-directional wheels to detect pointing gestures from a person wearing a red coat with a blue glove ~\cite{yoshida_evaluation_2011}. 

\subsubsection{Use Case Examples: Manipulator Robot Control} In manipulators, robots were controlled using hand gestures such as an open palm ~\cite{fareed_gesture_2015, lima_real-time_2019, martin_real-time_2019}). Hand gestures were used to command robot actions, such as to lift or lower the arm~\cite{fareed_gesture_2015, deepan_raj_static_2017, song_towards_2017, luo_human-robot_2019}, rotate the arm~\cite{choudhary_real_2015, song_towards_2017}, open or close the gripper~\cite{fareed_gesture_2015, lima_real-time_2019, martin_real-time_2019}, place an object into an open palm~\cite{arenas_deep_2017}, return to position when the palm is closed~\cite{arenas_deep_2017}, and to set positions for lifting and lowering~\cite{fareed_gesture_2015, deepan_raj_static_2017, luo_human-robot_2019}. Pointing was commonly used for selecting an object for grasping~\cite{quintero_visual_2015, moh_gesture_2019, valle_personalized_2019, sriram_mobile_2019}, including having the robot arm confirm object selection with the robot arm pointing at the object~\cite{quintero_visual_2015}. Hand gestures were also paired with other body movements for controlling manipulators~\cite{lima_real-time_2019, martin_real-time_2019}. Other works included more collaborative actions such as the robot helping to cook by dropping confirmed toppings over a pizza base~\cite{quintero_visual_2015}. A robot equipped with two arms, stereo vision, and tactile sensors could also pick up an object (sponge cube, wooden cube, ping-pong ball) that was selected by a hand pointing gesture from a human, and could release the object onto the palm of the person~\cite{ikai_robot_2016}.

\subsubsection{Use Case Examples: Mobile Manipulators and Aerial Robots} In mobile manipulators, pointing gestures were similarly used to select desired objects for the robot to pick up, e.g.~\cite{droeschel_towards_2011, qian_visually_2013, canal_gesture_2015, moh_gesture_2019, sriram_mobile_2019}. In one instance, a mobile manipulator responded to gestures (left and right hand) and user speech to identify, fetch, and handover objects such as a water bottle~\cite{burger_two-handed_2012}. Lastly, a mobile base with arms could wave back to a person waving at the robot and perform a behaviour as commanded by a dynamic gesture~\cite{li_real-time_2019}. Aerial examples include the use of body pose to control an aerial robot, such as right arm up to take off and right arm out to turn right \cite{sanna_kinect-based_2012} and pointing gestures to select an aerial robot and confirm  the selection by touching the right arm to the left hand \cite{lichtenstern_prototyping_2012}.  

\subsubsection{Use Case Examples: State Changes} Gestures were also used to signal the robot to commence state changes, e.g.~\cite{pereira_human-robot_2013, lalejini_evaluation_2015, ehlers_human-robot_2016, chen_human-following_2019}. Some examples include to initiate person guiding or following~\cite{pereira_human-robot_2013} or to indicate a path direction change for an otherwise autonomous robot~\cite{lalejini_evaluation_2015, ehlers_human-robot_2016}. Hand and body gestures were used to start/stop a walk action for a small humanoid~\cite{ratul_gesture_2016, shieh_fuzzy_2014}, body gestures to start/stop person following in an indoor environment~\cite{long_kinect-based_2018}, or left/right arm raised to change between robot following or parking behaviour~\cite{miller_self-driving_2019}. In one example, an autonomous mobile navigation robot explored a laboratory and asked humans for directions when a person was detected, translating pointing gestures to a goal in the robot's map~\cite{van_den_bergh_real-time_2011}. Gestures were also used in learning from demonstration to determine when a demonstration has commenced or concluded (hand~\cite{mazhar_real-time_2019} and body~\cite{stipancic_programming_2012}), or to update a robot's behaviour online~\cite{petric_online_2014, du_online_2018}. Further works on learning from demonstration will be discussed in Section~\ref{sec:rq2:learn}.

\subsubsection{Use Case Examples: Team-based Scenarios}
Gestures were also used in team-based scenarios, such as four mobile robots responding to gestures from a human operator~\cite{milligan_selecting_2011}. In this example, the human selected a group of robots by drawing a circle around robots, and directing the robots to go to a chosen location~\cite{milligan_selecting_2011}. Other team-based examples include the use of gesture-based interaction to signal to aerial and ground robot teams \cite{nagi_wisdom_2015}. Gestures were used to command a small swarm of mobile robots to move into a set configuration using body poses (i.e. arms out front, or above the persons head)~\cite{alonso-mora_human_2014}. Pointing gestures were often used to select a specific robot from a team of aerial robots~\cite{lichtenstern_prototyping_2012}, and to command a selected group of mobile robots~\cite{milligan_selecting_2011}. This included pointing to direct robot attention to other human targets~\cite{luo_tracking_2011}. Lastly, one example showed a multi-person interaction, including a mobile robot that identified a person by localising from an audio source, and then determining which person to track when they waved at the robot~\cite{nguyen_audio-visual_2014}.

\subsubsection{Use Case Examples: \DIFdelbegin \DIFdel{Social Action }\DIFdelend \DIFaddbegin \DIFadd{Implicit (Non-Verbal) Communication }\DIFaddend and \DIFaddbegin \DIFadd{Social }\DIFaddend Interactivity} \DIFdelbegin \DIFdel{Gestures were }\DIFdelend \DIFaddbegin \DIFadd{There were fewer papers around gestures being }\DIFaddend performed by anthropomorphic robots to mimic human gestures for \DIFaddbegin \DIFadd{the purpose of }\DIFaddend social interaction. An example includes \DIFaddbegin \DIFadd{hand }\DIFaddend waving from a humanoid robot in response to a human \DIFaddbegin \DIFadd{human }\DIFaddend wave~\cite{canal_gesture_2015}\DIFaddbegin \DIFadd{, helping to facilitate non-verbal communication}\DIFaddend . In another example, gesture recognition was used for humanoid robots (Pepper and NAO) to perform finger spelling gestures to communicate with hearing impaired individuals at a public service centre~\cite{kalidolda_towards_2018}. Social interactivity with robots also involved gesture-based games, such as paper-scissors-rock which required the robot to classify human pose to determine the result~\cite{katsuki_high_speed_2015, yuan_natural_2020}. Lastly, a game with the iCub robot required the robot to recognise each gesture performed by the person to participate in the game~\cite{gori_all_2012}.

\subsubsection{Included Papers} 
Papers related to gesture recognition are listed here:
~\cite{chen_integrated_2010,chen_approaches_2010,couture-beil_selecting_2010,manigandan_wireless_2010,broccia_gestural_2011,milligan_selecting_2011,uribe_mobile_2011,yoshida_evaluation_2011,celik_development_2012,de_luca_integrated_2012,faudzi_real-time_2012,gu_human_2012,konda_real_2012,lee_interactive_2012,lichtenstern_prototyping_2012,sanna_kinect-based_2012,baron_remote_2013,hafiane_3d_2013,hartmann_feasibility_2013,li_cyber-physical_2013,nishiyama_human_2013,pfeil_exploring_2013,qian_visually_2013,tao_multilayer_2013,monajjemi_hri_2013,wang_wheeled_2013,alonso-mora_human_2014,barros_real-time_2014,chao_robotic_2014,costante_personalizing_2014,farulla_real-time_2014,nguyen_audio-visual_2014,ozgur_natural_2014,quintero_interactive_2014,xiao_human-robot_2014,xu_online_2014,canal_gesture_2015,katsuki_high-speed_2015,monajjemi_uav_2015,nagi_wisdom_2015,saichinmayi_gesture_2015,manitsaris_fingers_2016,maraj_application_2016,han_human_2017,ju_integrative_2017,maher_realtime_2017,bras_gesture_2018,mohaimenianpour_hands_2018,shakev_autonomous_2018,azari_commodifying_2019,chen_online_2019,li_real-time_2019,yu_efficiency_2019,angani_human_2020,chen_design_2020,hsu_real-time_2020,medeiros_human-drone_2020,yue_human-robot_2020,martin_estimation_2010,droeschel_towards_2011,luo_tracking_2011,park_real-time_2011,van_den_bergh_real-time_2011,burger_two-handed_2012,cheng_design_2012,gori_all_2012,lambrecht_spatial_2012,abidi_human_2013,shieh_fuzzy_2014,choudhary_real_2015,cicirelli_kinect-based_2015,fareed_gesture_2015,gao_humanoid_2015,jevtic_comparison_2015,lalejini_evaluation_2015,quintero_visual_2015,vircikova_teach_2015,yang_real-time_2015,ehlers_human-robot_2016,ratul_gesture_2016,arenas_deep_2017,deepan_raj_static_2017,maurtua_natural_2017,song_towards_2017,tolgyessy_foundations_2017,kalidolda_towards_2018,lavanya_gesture_2018,mao_medical_2018,pentiuc_drive_2018,zhang_interactive_2018,fang_vehicle-mounted_2019,lima_real-time_2019,luo_human-robot_2019,martin_real-time_2019,moh_gesture_2019,sriram_mobile_2019,vysocky_interaction_2019,waskito_wheeled_2020,xu_skeleton_2020,yuan_natural_2020,petric_online_2014,valipour_incremental_2017,mazhar_real-time_2019,stipancic_programming_2012,du_online_2018,ikai_robot_2016,valle_personalized_2019,zhang_gesture-based_2019,pereira_human-robot_2013,fujii_gesture_2014,prediger_robot-supported_2014,ghandour_human_2017,long_kinect-based_2018,yuan_development_2018,chen_human-following_2019,miller_self-driving_2019}.

\subsection{Action Recognition}\label{sec:rq2:action}

\begin{figure}[ht!]
\centering
\subfloat[Total Annual Papers]{\includegraphics[width=0.45\columnwidth]{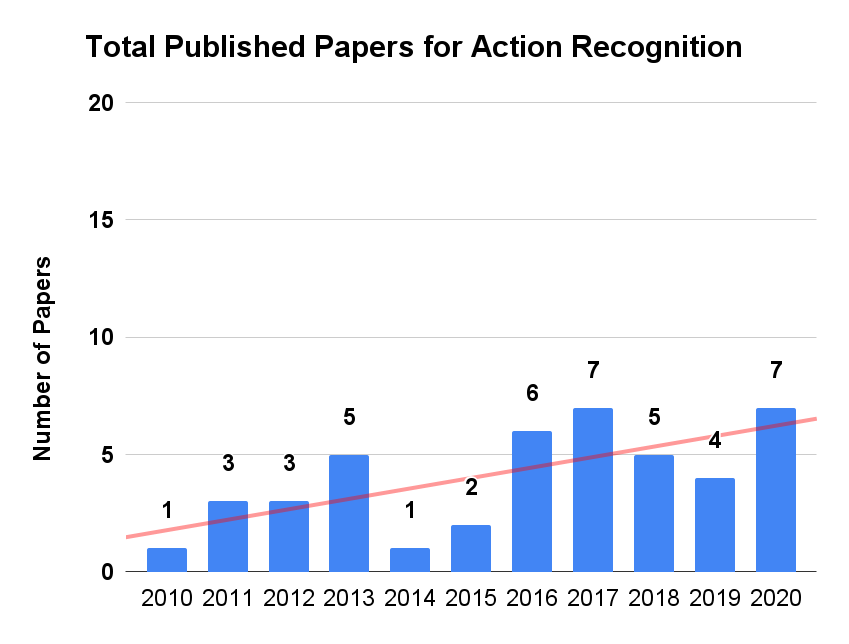}}
\hfil
\subfloat[Domains]{\includegraphics[width=0.45\columnwidth]{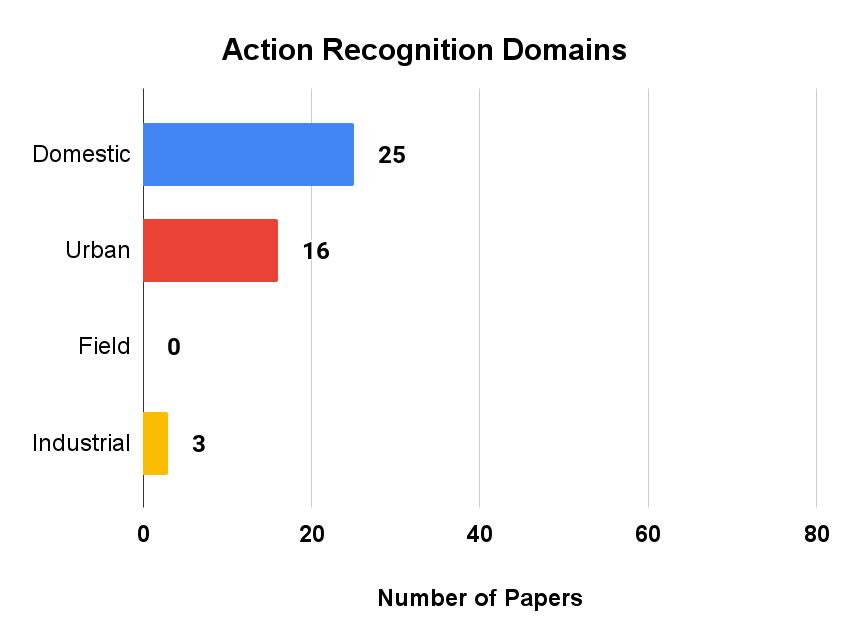}}
\hfil
\subfloat[Camera Types]{\includegraphics[width=0.45\columnwidth]{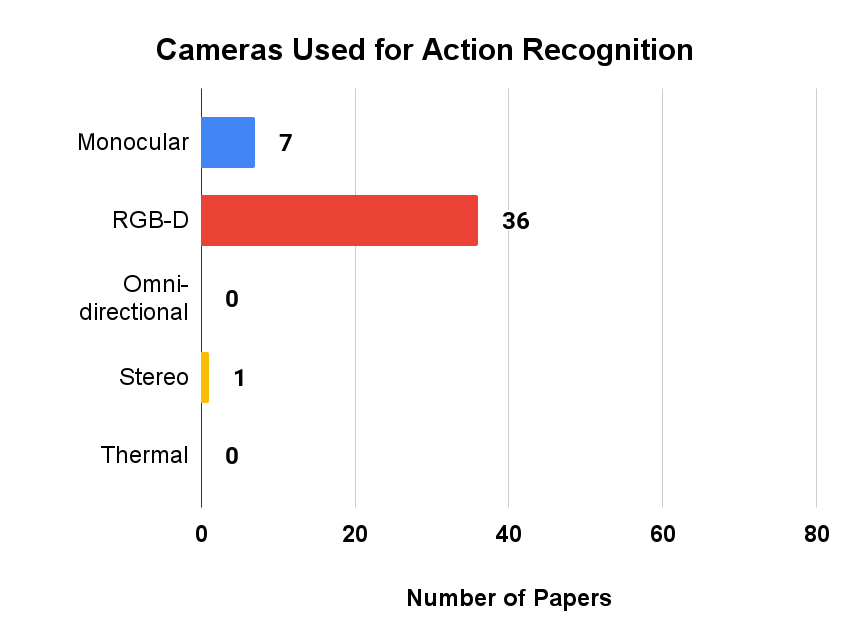}}
\hfil
\subfloat[Robot Types]{\includegraphics[width=0.45\columnwidth]{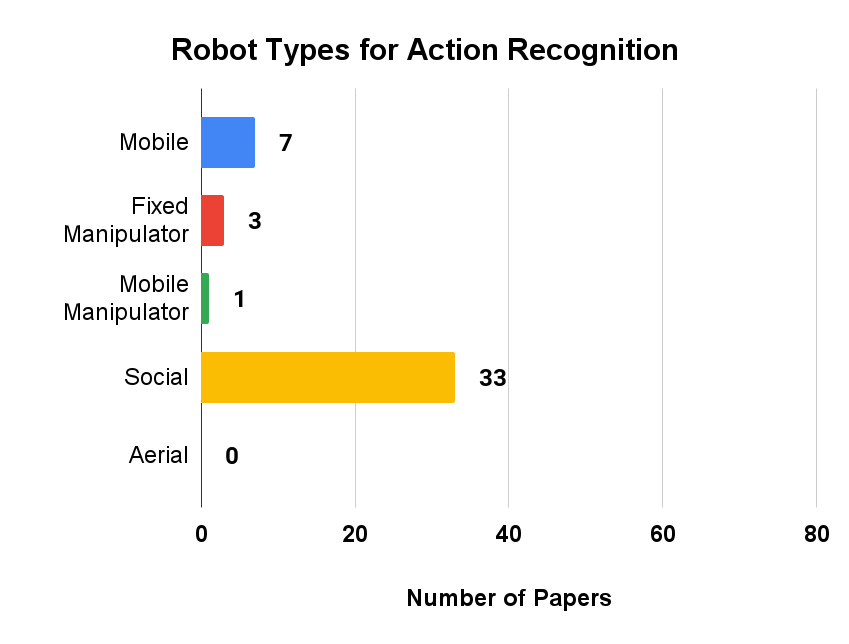}}
\hfil
\subfloat[Action Types]{\includegraphics[width=0.45\columnwidth]{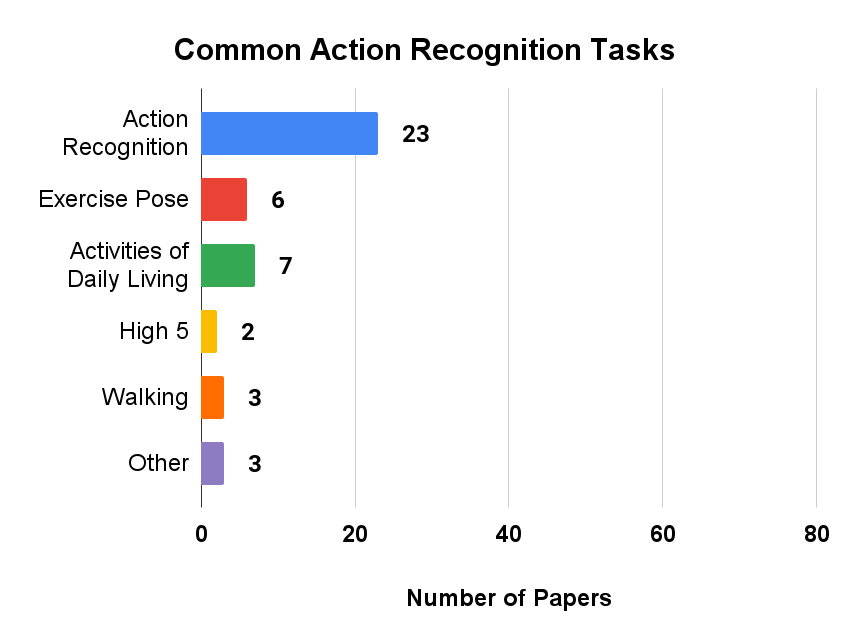}}
\hfil
\subfloat[Autonomy Level]{\includegraphics[width=0.45\columnwidth]{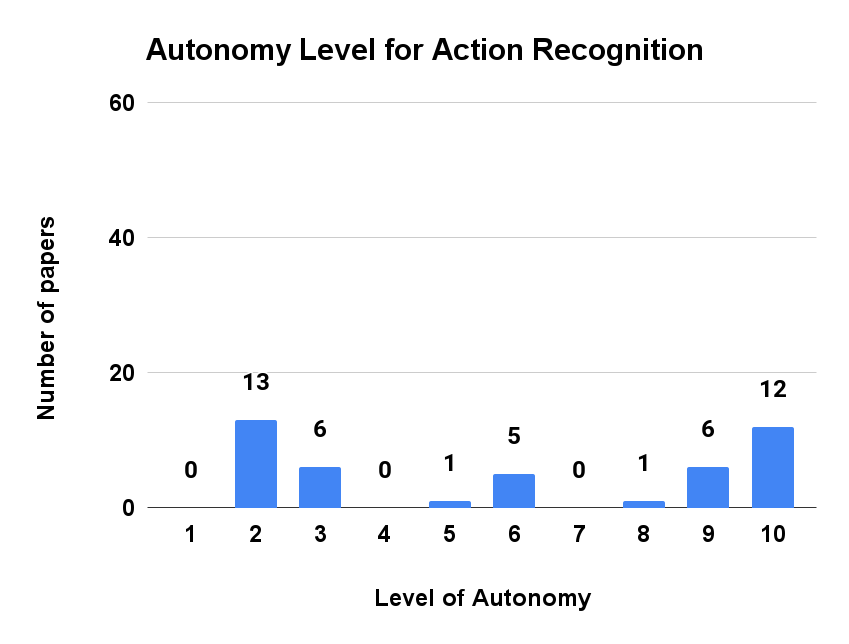}}
\hfil
\caption{Action Recognition Totals and Summaries}
\label{fig:action_recognition}
\vspace{-3mm}
\end{figure}

\subsubsection{Overview}
Action recognition was defined as the recognition of human actions or activities that were not related to explicit gestures. A total of \DIFdelbegin \DIFdel{43 }\DIFdelend \DIFaddbegin \DIFadd{44 }\DIFaddend papers (14\% of eligible total) involved some form of action or activity recognition. Figure~\ref{fig:action_recognition} shows the number of action recognition-related works, common domains, camera types, robot types, action types and level of autonomy. Of the \DIFdelbegin \DIFdel{35 }\DIFdelend \DIFaddbegin \DIFadd{36 }\DIFaddend that used RGB-D cameras, \DIFdelbegin \DIFdel{32 }\DIFdelend \DIFaddbegin \DIFadd{33 }\DIFaddend were the Kinect (\DIFdelbegin \DIFdel{91}\DIFdelend \DIFaddbegin \DIFadd{92}\DIFaddend \%). Action recognition often involved recognising the person's activities such as action recognition and response (\DIFdelbegin \DIFdel{$N=22$, 51}\DIFdelend \DIFaddbegin \DIFadd{$N=23$, 52}\DIFaddend \%), activities of daily living ($N=7$, 16\%), exercise pose ($N=6$, 14\%), and recognition of their walking motion ($N=3$, 7\%). Humanoid robots often used action recognition: NAO~\cite{werner_evaluation_2013, gorer_autonomous_2017, efthymiou_multi-_2018, avioz-sarig_robotic_2020}, Pepper~\cite{gui_teaching_2018, sorostinean_activity_2018, lang_research_2020}, other humanoids~\cite{potdar_learning_2016, avioz-sarig_robotic_2020} and mobile robots~\cite{zhang_automating_2018, vasquez_deep_2017, lee_real-time_2020}. 

\subsubsection{Use Case Examples}
Action recognition often involved the identification of states. For action recognition, a robot could make a decision on when to offer a person a footrest to rest their feet~\cite{zhang_automating_2018}, if the person had fallen down to ask them if they could call an ambulance~\cite{sorostinean_activity_2018}, or when to respond to a human that was handing a bottle to the robot~\cite{potdar_learning_2016}. This included to help robots to 
recognise multiple actions such as eating, brushing teeth, and making a phone call e.g.~\cite{lee_real-time_2020}. Action recognition was used to help the robot to predict human motion such as walking, eating, smoking, and to infer the remaining motion sequence after the camera was occluded~\cite{gui_teaching_2018}, as well as to predict human actions in a shared workspace, such as to avoid collision during tool use by recognising the use of a hammer or reaching for a cup~\cite{wang_collision-free_2017}. Action recognition was also used to allow a robot to detect other body actions (i.e. shake head, wave hand) and then perform a corresponding behaviour~\cite{lang_research_2020}. Other examples include to understand and copy human motions, such as joint positions~\cite{igorevich_behavioral_2011, csapo_multimodal_2012, torres_implementation_2012, indrajit_development_2013, yang_study_2013, taheri_social_2014, zhu_robust_2017, augello_towards_2020, lu_research_2020}, head positions~\cite{cazzato_real-time_2019}, facial expressions~\cite{sosnowski_mirror_2010, masmoudi_expressive_2011, taheri_social_2014, meghdari_real-time_2017}, or following continuous position of the persons hand~\cite{paulo_vision-based_2012}, or head~\cite{lam_real-time_2011}. This also involved more rigorous body motions such as humans performing physical activity and the robot could give feedback to the person on a chosen exercise about pose quality~\cite{werner_evaluation_2013, gorer_autonomous_2017, avioz-sarig_robotic_2020}. Other physical activity examples include recording pose count and signal to change exercises if the person waved their hand ~\cite{avioz-sarig_robotic_2020}, and the robot learning exercises through action recognition and response to a human demonstrator~\cite{gorer_autonomous_2017}. In a more applied environment, action recognition was used to detect walking ability for a service robot to be able to guide people of different walking capacities (i.e. wheelchairs, crutches, or walkers) to a suitable entrance~\cite{vasquez_deep_2017}. In particular, motion analysis helped the robot to track the persons position to adapt the motion of a robotic walking aid for different mobility levels~\cite{chalvatzaki_learn_2019}. For service assistance, action recognition was used to determine when to provide domestic chore assistance, such as filling a glass of water, opening a fridge door~\cite{koppula_anticipating_2016}, clearing a table, or pushing a trivet to the person~\cite{lee_learning_2017}. Lastly, some forms of action recognition were used in playful contexts, such as a child and humanoid robot taking turns to perform and recognise a pantomime action such as swimming, painting a wall, or digging a hole~\cite{efthymiou_multi-_2018}. 

\subsubsection{Included Papers} Papers related to action recognition are listed here:
~\cite{sosnowski_mirror_2010,igorevich_behavioral_2011,lam_real-time_2011,masmoudi_expressive_2011,paulo_vision-based_2012,torres_implementation_2012,cid_real_2013,indrajit_development_2013,mohammad_tele-operation_2013,werner_evaluation_2013,yang_study_2013,obo_robot_2015,saffar_context-based_2015,agrigoroaie_enrichme_2016,liu_interactive_2016,silva_mirroring_2016,zhu_real-time_2016,gorer_autonomous_2017,meghdari_real-time_2017,vasquez_deep_2017,vignolo_computational_2017,zhu_robust_2017,devanne_co-design_2018,efthymiou_multi-_2018,zhang_automating_2018,abiddin_development_2019,cazzato_real-time_2019,chalvatzaki_learn_2019,sripada_teleoperation_2019,augello_towards_2020,avioz-sarig_robotic_2020,lang_research_2020,lu_research_2020,koppula_anticipating_2016,lee_learning_2017,wang_collision-free_2017,gui_teaching_2018,sorostinean_activity_2018,lee_real-time_2020,potdar_learning_2016,yan_optimization_2020,csapo_multimodal_2012,taheri_social_2014,santos_copyrobot_2020} 

\subsection{Robot Movement in Human Spaces}\label{sec:rq2:movement}

\subsubsection{Overview}
\begin{figure}[ht!]
\centering
\subfloat[Total Papers]{\includegraphics[width=0.45\columnwidth]{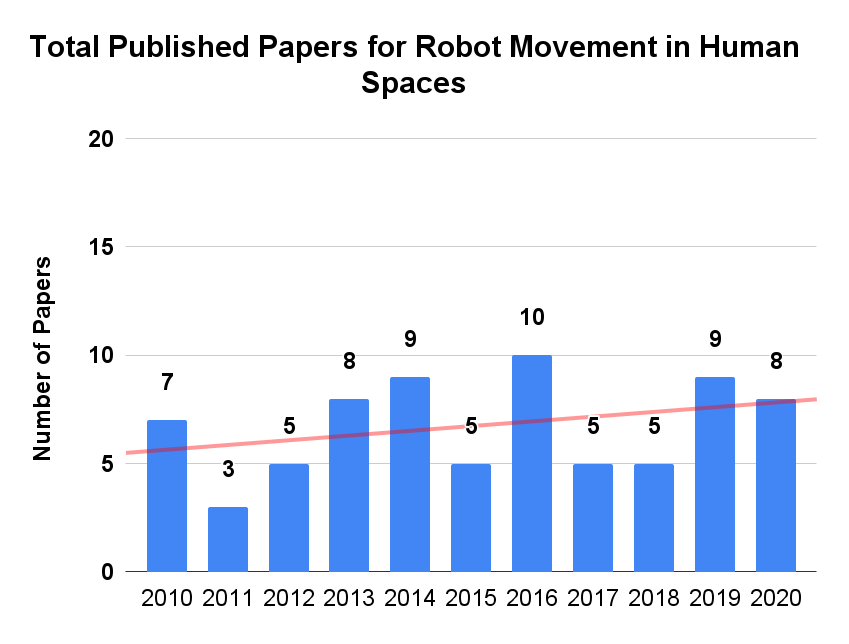}}
\hfil
\subfloat[Domains]{\includegraphics[width=0.45\columnwidth]{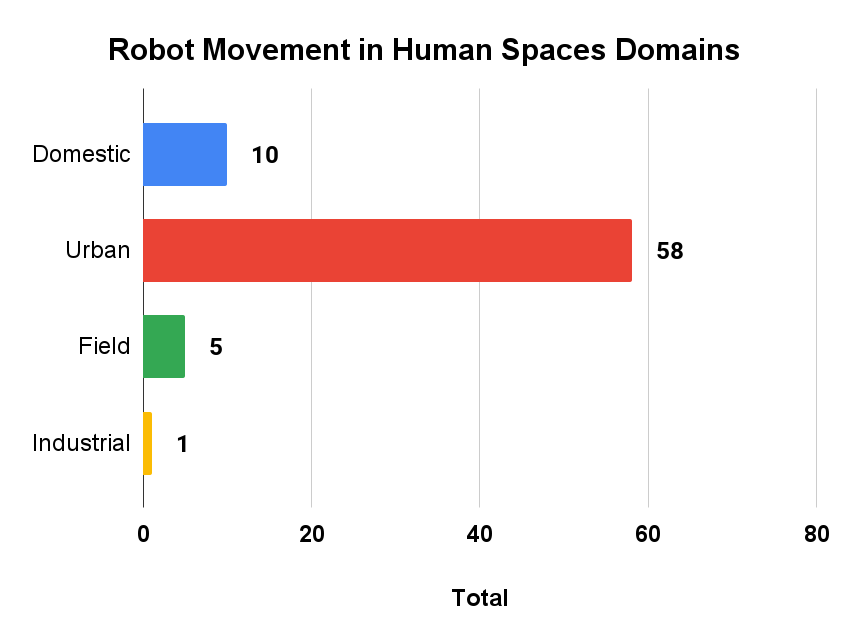}}
\hfil
\subfloat[Camera Types]{\includegraphics[width=0.45\columnwidth]{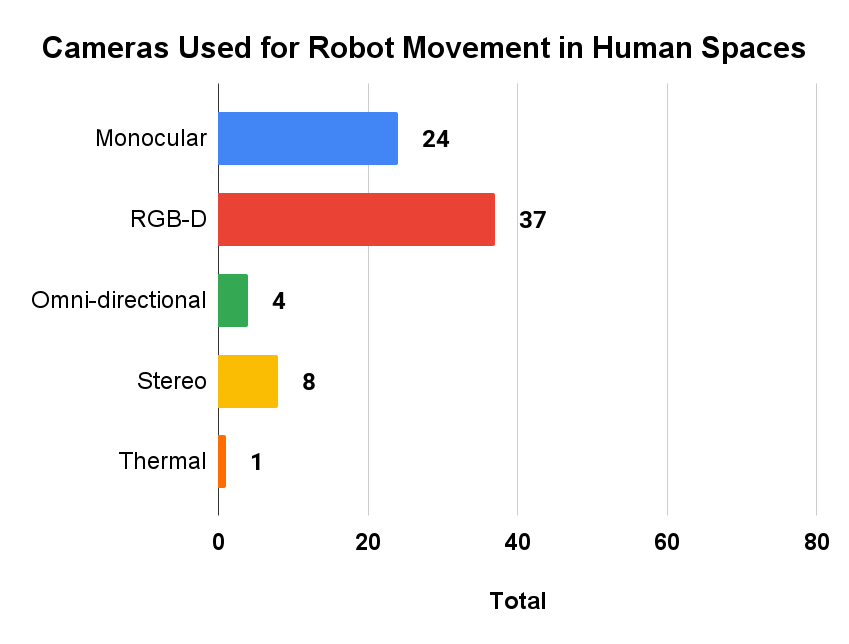}}
\hfil
\subfloat[Robot Types]{\includegraphics[width=0.45\columnwidth]{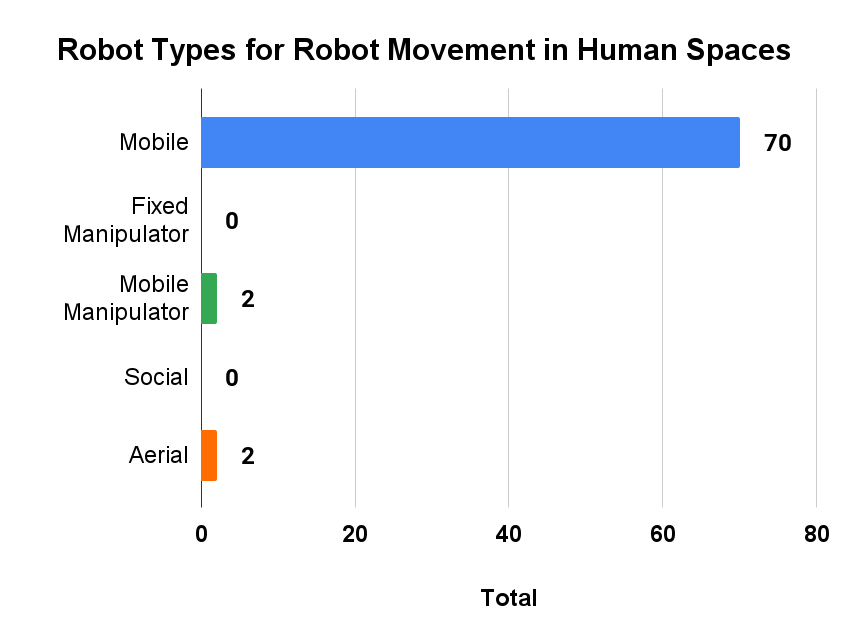}}
\hfil
\subfloat[Common Tasks]{\includegraphics[width=0.45\columnwidth]{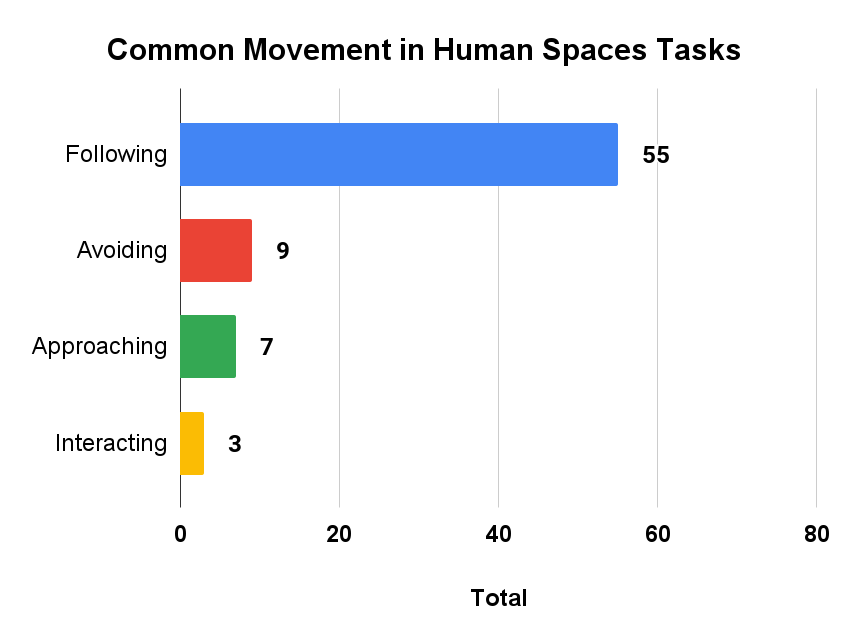}}
\hfil
\subfloat[Autonomy Level]{\includegraphics[width=0.45\columnwidth]{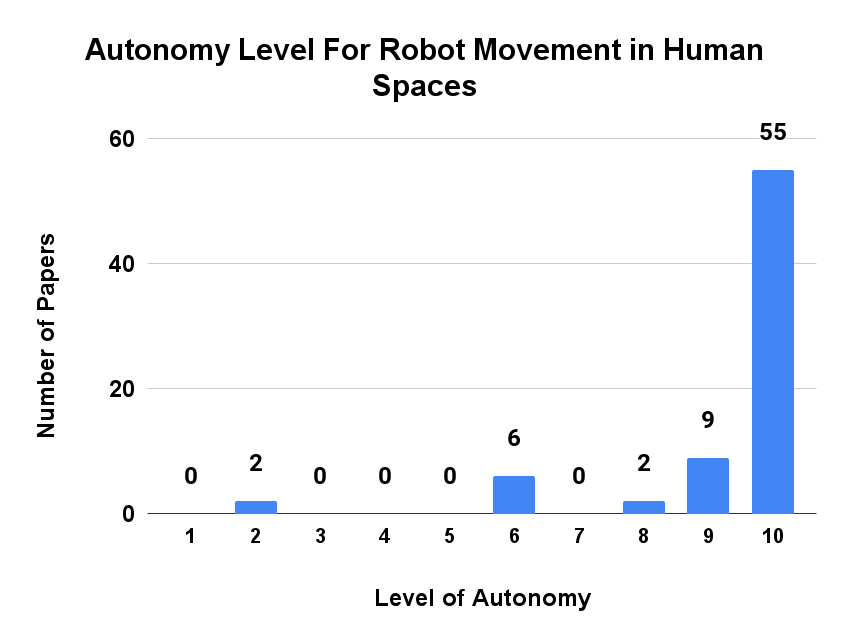}}
\hfil
\caption{Robot Movement in Human Spaces Totals and Summaries}
\label{fig:rmhs}
\vspace{-3mm}
\end{figure}

Robot movement in human spaces was classified if the robot had physical movement in a human-based environment and robot movement did not require the human to perform a set pose to signal movement commands to the robot\DIFaddbegin \DIFadd{, including if the robot was classified as a (semi-)autonomous vehicle}\DIFaddend . A total of \DIFdelbegin \DIFdel{69 papers (23}\DIFdelend \DIFaddbegin \DIFadd{74 papers (24}\DIFaddend \% of eligible total) used robot movement in human spaces. Figure~\ref{fig:rmhs} shows the number of action recognition-related works, common domains, camera types, robot types, common tasks, and level of autonomy. In the \DIFdelbegin \DIFdel{34 }\DIFdelend \DIFaddbegin \DIFadd{37 }\DIFaddend papers that used RGB-D cameras, \DIFdelbegin \DIFdel{26 }\DIFdelend \DIFaddbegin \DIFadd{29 }\DIFaddend were the Kinect (\DIFdelbegin \DIFdel{76}\DIFdelend \DIFaddbegin \DIFadd{78}\DIFaddend \%). Common robot tasks were following the person (\DIFdelbegin \DIFdel{$n=52$, 75}\DIFdelend \DIFaddbegin \DIFadd{$n=55$, 74}\DIFaddend \%), avoiding a person (\DIFdelbegin \DIFdel{$n=8$}\DIFdelend \DIFaddbegin \DIFadd{$n=9$}\DIFaddend , 12\%), and approaching one or more people (\DIFdelbegin \DIFdel{$n=6$}\DIFdelend \DIFaddbegin \DIFadd{$n=7$}\DIFaddend , 9\%). In total, body pose detection was the most common method for identifying a person in an image (\DIFdelbegin \DIFdel{$n=51$, 75}\DIFdelend \DIFaddbegin \DIFadd{$n=57$, 77}\DIFaddend \%), followed by face detection ($n=14$, \DIFdelbegin \DIFdel{21}\DIFdelend \DIFaddbegin \DIFadd{19}\DIFaddend \%). Other methods involved tracking clothing or detection of clothing~ \cite{wu_accompanist_2012, miyoshi_above_2014}. Mobile robots often had laser range sensors for person detection~\cite{luo_human_2010, alvarez-santos_feature_2012, pereira_human-robot_2013, weinrich_appearance-based_2013, hu_design_2014, kobayashi_people_2010}, obstacle avoidance~\cite{yun_robotic_2013, ali_improved_2015, yang_socially-aware_2019}, and for navigation (i.e. SLAM)~\cite{do_human-robot_2014, angonese_multiple_2017, yuan_development_2018, miller_self-driving_2019}. Depth images were also used for obstacle avoidance~\cite{bayram_audio-visual_2016} and SLAM~\cite{talebpour-board_2016}. Ultra-sonic sensors were also used for person following~\cite{hassan_computationally_2016}, and for navigation~\cite{chien_navigating_2019}, as well as audio to localise a person not in view~\cite{luo_human_2010, nguyen_audio-visual_2014, bayram_audio-visual_2016}. Re-identification when a person who had become occluded when following the person was addressed in several papers~\cite{luo_human_2010, wu_accompanist_2012, weinrich_appearance-based_2013, condes_person_2019, zhang_vision-based_2019}. Some papers used multimodal detectors such as laser or ultrasonic range sensors to identify a person (i.e. detecting legs ($n=9$) or shoulders ($n=1$)), and audio localisation to determine if a person was out of view ($n=4$). Others required minimal intervention from the person through gesture commands ($n=9$). Proxemics was often considered for appropriate social distance to approach~\cite{mead_probabilistic_2012}, and avoid people~\cite{talebpour-board_2016, vasconcelos_socially_2016, yang_socially-aware_2019, bellarbi_social_2017}, including velocity when a person is detected~\cite{vasconcelos_socially_2016}. Person following environments included both urban~\cite{do_human-robot_2014} and indoor settings~\cite{kobayashi_people_2010, wu_accompanist_2012, wu_improved_2019,long_kinect-based_2018,wu_toward_2020}. Commonly used mobile platforms included Pioneer mobile robots (\DIFdelbegin \DIFdel{$n=14$}\DIFdelend \DIFaddbegin \DIFadd{$n=16$}\DIFaddend ), SCITOS G5~\cite{weinrich_appearance-based_2013, weber_follow_2018}, iRobot create \cite{do_human-robot_2014}, a wheelchair robot ~\cite{wu_accompanist_2012}, Turtlebot~\cite{condes_person_2019}, and even a robotic blimp \cite{yao_monocular_2017}. 

\subsubsection{Use Case Examples \DIFaddbegin \DIFadd{- Social and Functional Navigation}\DIFaddend } 
There were several key examples for robots \DIFdelbegin \DIFdel{following people }\DIFdelend \DIFaddbegin \DIFadd{that followed people for both social and functional reasons~\cite{ long_kinect-based_2018, condes_person_2019, luo_real-time_2019, wu_improved_2019, zhang_vision-based_2019, anuradha_human_2020, hwang_interactions_2020, wu_toward_2020}. Many works }\DIFaddend required the robot to approach the person \DIFaddbegin \DIFadd{to commence navigation}\DIFaddend ~\cite{budiharto_indoor_2010, mead_probabilistic_2012, cheng_multiple-robot_2013, ferrer_robot_2013, chen_stereovision-only_2014}, \DIFdelbegin \DIFdel{as well as to avoid }\DIFdelend \DIFaddbegin \DIFadd{and with human detection and tracking to know where the person was to avoid colliding with }\DIFaddend them~\cite{talebpour-board_2016, vasconcelos_socially_2016, ghandour_human_2017, yang_socially-aware_2019, bellarbi_social_2017}. \DIFdelbegin \DIFdel{Others }\DIFdelend \DIFaddbegin \DIFadd{In addition, other works included integrating humans into a map for a mobile robot~\mbox{
\cite{angonese_multiple_2017}}\hspace{0pt}
. Other works }\DIFaddend had the robot approach and \DIFaddbegin \DIFadd{then }\DIFaddend interact with the person by initiating a dialogue~\cite{ferrer_robot_2013}, or \DIFdelbegin \DIFdel{used }\DIFdelend \DIFaddbegin \DIFadd{by creating a new goal through the use of human }\DIFaddend gestures to change \DIFdelbegin \DIFdel{robot states}\DIFdelend \DIFaddbegin \DIFadd{the robots state}\DIFaddend , such as to \DIFdelbegin \DIFdel{start a }\DIFdelend \DIFaddbegin \DIFadd{commence a variant of a }\DIFaddend person following behaviour~\cite{pereira_human-robot_2013, fujii_gesture_2014, prediger_robot-supported_2014, long_kinect-based_2018, yuan_development_2018, chen_human-following_2019, miller_self-driving_2019}. Other state changes included to guide \DIFaddbegin \DIFadd{the }\DIFaddend people avoiding the robot towards a particular path~\cite{ghandour_human_2017}. In \DIFdelbegin \DIFdel{approach-related examples}\DIFdelend \DIFaddbegin \DIFadd{works when the robot approached the person}\DIFaddend , a line following robot approached a person when their face was detected~\cite{budiharto_indoor_2010} and a PR2 robot approached a person using a real-time proxemic controller~\cite{mead_probabilistic_2012}\DIFaddbegin \DIFadd{, helping to bridge functional navigation methods towards social interaction points}\DIFaddend . In one example, \DIFdelbegin \DIFdel{robots }\DIFdelend \DIFaddbegin \DIFadd{a robot }\DIFaddend had an omni-directional camera for person detection, and a laser that is used to navigate through the environment using SLAM~\cite{do_human-robot_2014}. This included a social force model that allowed for appropriate social distance when passing people and to avoid having the \DIFdelbegin \DIFdel{robots }\DIFdelend \DIFaddbegin \DIFadd{robot }\DIFaddend cross into a person's personal or intimate space when passing a person from behind~\cite{yang_socially-aware_2019,bellarbi_social_2017}. \DIFdelbegin \DIFdel{Some robots had multiple tasks in the application. For instance, a mobile R2D2 robot with arms carried a smaller mobile robot with a gripper. The robot waves when it detects a person's face, and delivers a drink based on the distance to the target face. The main robot could also deploy a smaller robot with a gripper to pick up small objects in its path~\mbox{
\cite{cheng_multiple-robot_2013}}\hspace{0pt}
. }\DIFdelend Robot movement through human spaces was \DIFdelbegin \DIFdel{involved robot teams}\DIFdelend \DIFaddbegin \DIFadd{also tested in busy environments}\DIFaddend , such as an iRobot that could navigate through urban environments, detect human faces and report back to a supervisory person~\cite{do_human-robot_2014}. Multiple people were also detected and used as landmarks for integration into a SLAM solution for a Pioneer robot following a path~\cite{angonese_multiple_2017}. This included robust person following, even when multiple people were present in a scene and when the target became occluded~\cite{munaro_fast_2014}. The iRobot was also used to identify the face of a specific person, and keep the target person's face in view~\cite{zhang_optimal_2016}. In \DIFdelbegin \DIFdel{addition, other work included integrating humans into a map for a }\DIFdelend \DIFaddbegin \DIFadd{robot navigation for human environments, some robots also conducted multiple tasks together as part of its use case. For example, a }\DIFaddend mobile \DIFdelbegin \DIFdel{robot~\mbox{
\cite{angonese_multiple_2017}}\hspace{0pt}
}\DIFdelend \DIFaddbegin \DIFadd{R2D2 robot with arms carried a smaller mobile robot with a gripper. The robot could wave when it detected a person's face, and delivered a drink based on the distance to the target face. The main robot could also deploy a smaller robot with a gripper to pick up small objects in its path~\mbox{
\cite{cheng_multiple-robot_2013}}\hspace{0pt}
}\DIFaddend . Others had \DIFdelbegin \DIFdel{other }\DIFdelend \DIFaddbegin \DIFadd{different }\DIFaddend methods of person following, such as \DIFdelbegin \DIFdel{through the air. Two other }\DIFdelend \DIFaddbegin \DIFadd{following a person through airspace. Two }\DIFaddend examples involved using a two camera vision system with a small aerial robot to detect and hover above the hand of a person wearing a glove with the intention to pass the robot between people~\cite{miyoshi_above_2014}, and a monocular camera attached to an autonomous blimp robot to detect and follow a persons face~\cite{yao_monocular_2017}. 

\subsubsection{Included Papers} Papers related to robot movement in human spaces are listed here:
\cite{nguyen_audio-visual_2014,de_schepper_towards_2020,budiharto_indoor_2010,fahn_real-time_2010,kobayashi_people_2010,mendez-polanco_detection_2010,weiss_robots_2010,yun_robust_2010,jia_autonomous_2011,pustianu_mobile_2011,luo_human_2010,alvarez-santos_feature_2012,chen_person_2012,lee_human_2012,mead_probabilistic_2012,wu_accompanist_2012,cheng_multiple-robot_2013,ferrer_robot_2013,granata_human_2013,hegger_people_2013,pereira_human-robot_2013,weinrich_appearance-based_2013,yun_robotic_2013,zhao_interactive_2013,chen_stereovision-only_2014,do_human-robot_2014,fujii_gesture_2014,hu_design_2014,miyoshi_above_2014,munaro_fast_2014,prediger_robot-supported_2014,scheggi_human-robot_2014,ali_improved_2015,batista_probabilistic_2015,gupta_robust_2015,voisan_ros-based_2015,bayram_audio-visual_2016,hassan_computationally_2016,kahily_real-time_2016,mateus_human-aware_2016,talebpour-board_2016,vasconcelos_socially_2016,zhang_optimal_2016,angonese_multiple_2017,bellarbi_social_2017,ghandour_human_2017,rehman_target_2017,yao_monocular_2017,almonfrey_flexible_2018,long_kinect-based_2018,weber_follow_2018,yuan_development_2018,zhang_indoor_2018,ali_smart_2019,chen_human-following_2019,condes_person_2019,gemerek_video-guided_2019,luo_real-time_2019,miller_self-driving_2019,wu_improved_2019,yang_socially-aware_2019,zhang_vision-based_2019,anuradha_human_2020,hwang_interactions_2020,kawasaki_multimodal_2020,muller_multi-modal_2020,pang_efficient_2020,wu_toward_2020,chou_development_2020,brookshire_person_2010,pennisi_multi-robot_2015,gardel_wireless_2016,rosa_integration_2016,lovon-ramos_people_2016}. 

\subsection{Object Handover and Collaborative Actions}\label{sec:rq2:handover}

\subsubsection{Overview}
\begin{figure}[ht!]
\centering
\subfloat[Total Papers]{\includegraphics[width=0.45\columnwidth]{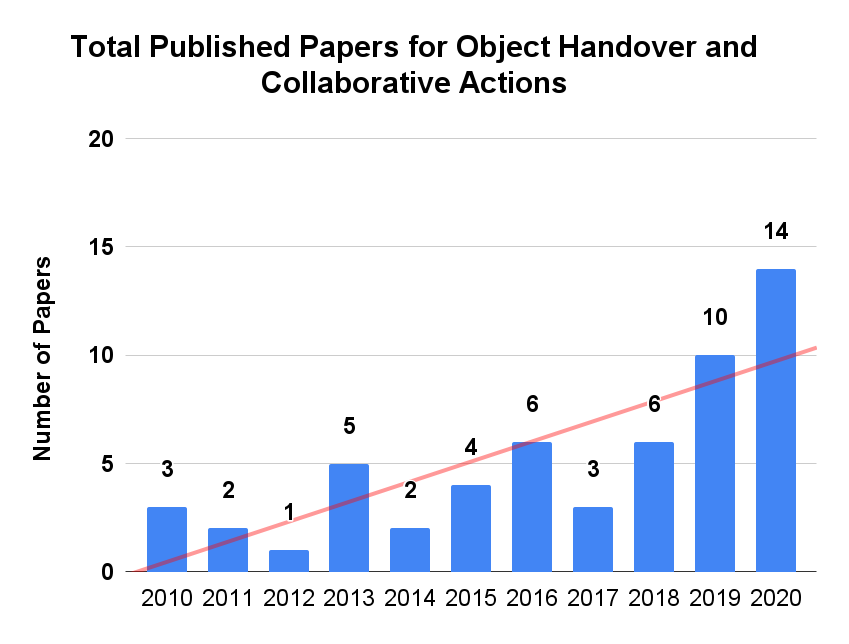}}
\hfil
\subfloat[Domains]{\includegraphics[width=0.45\columnwidth]{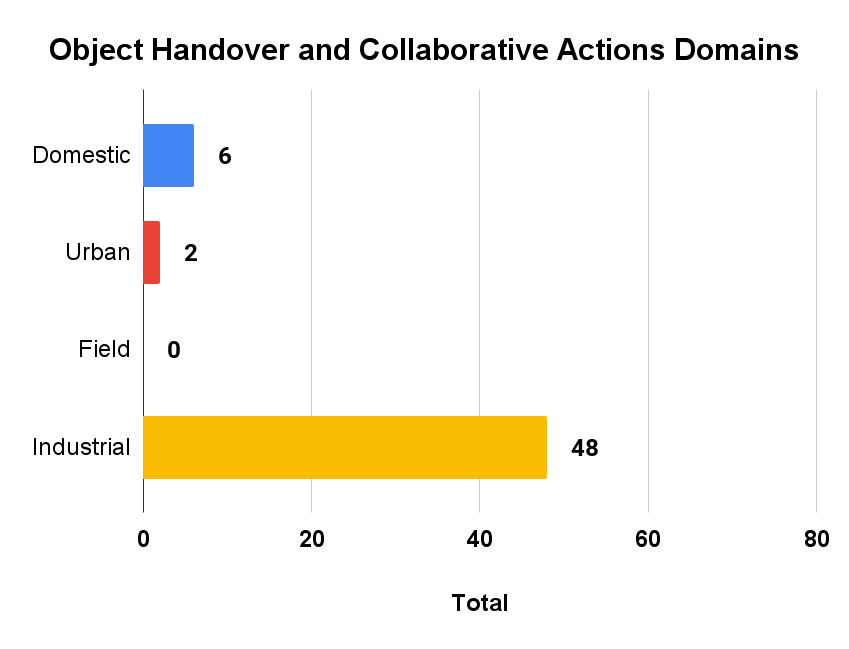}}
\hfil
\subfloat[Camera Types]{\includegraphics[width=0.45\columnwidth]{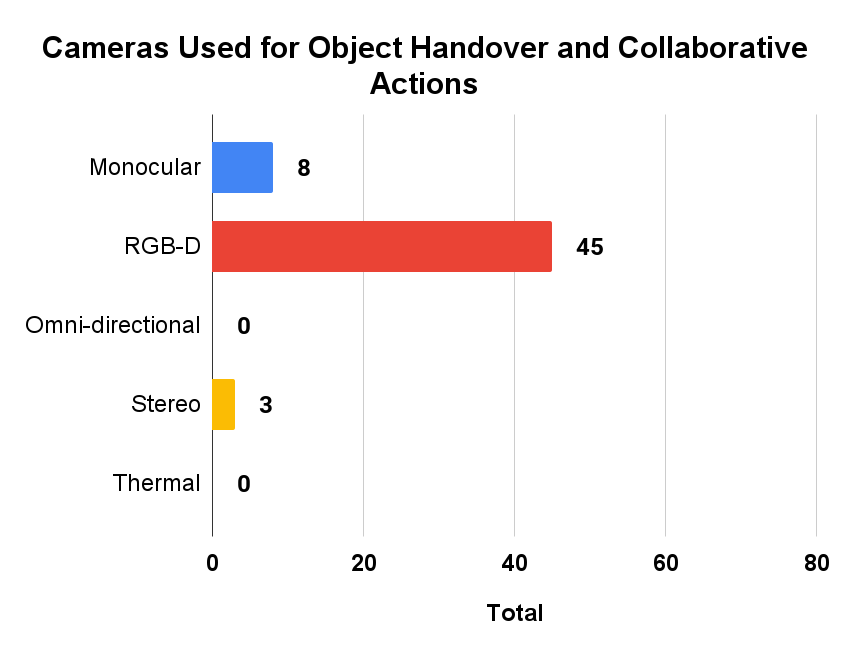}}
\hfil
\subfloat[Robot Types]{\includegraphics[width=0.45\columnwidth]{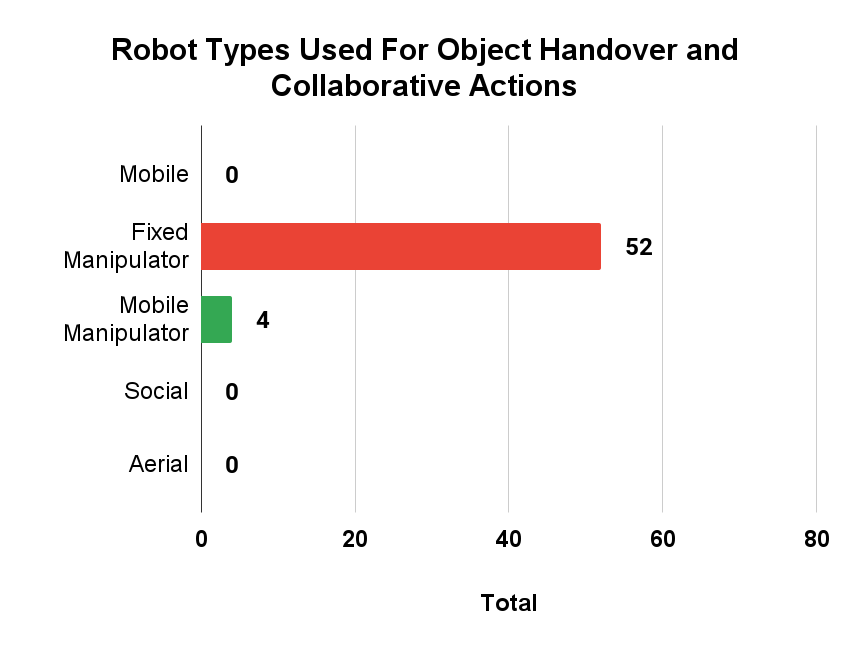}}
\hfil
\subfloat[Common Tasks]{\includegraphics[width=0.45\columnwidth]{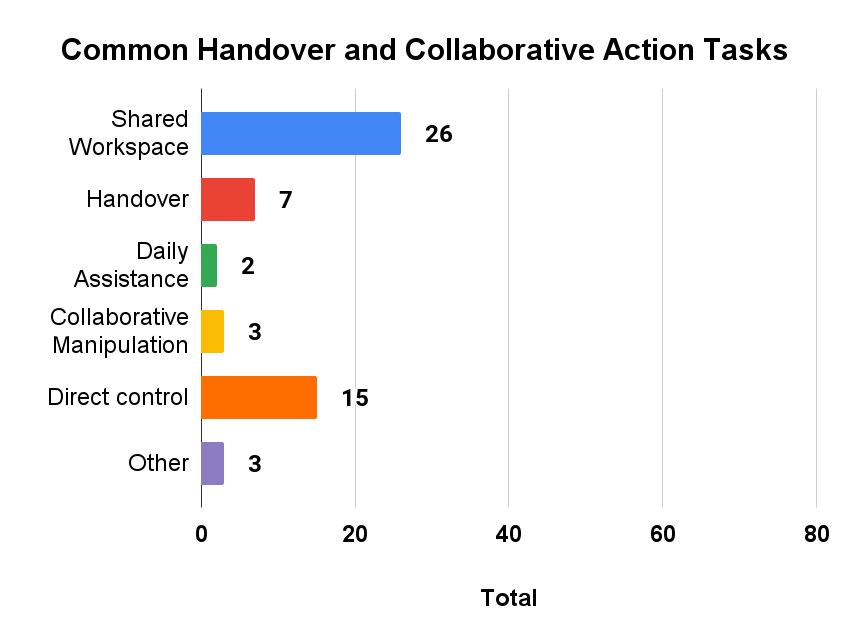}}
\hfil
\subfloat[Autonomy Level]{\includegraphics[width=0.45\columnwidth]{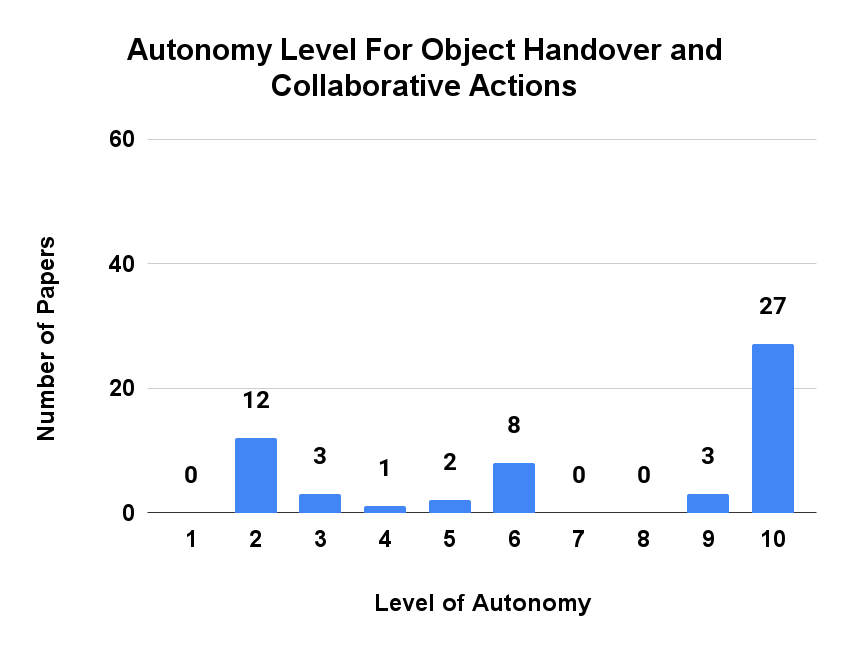}}
\hfil
\caption{Object Handover and Collaborative Actions Totals and Summaries}
\label{fig:ohca}
\vspace{-3mm}
\end{figure}

Object handover and collaborative action papers included a robot capable of manipulating objects while the interaction did not require the person to perform a set pose (i.e the person was detected without performing a gesture or action). A total of 56 papers (18\% of eligible total) involved an object handover or collaborative action. Figure~\ref{fig:ohca} shows the number of works, common domains, camera types, robot types, common tasks and level of autonomy. In the 45 papers that used RGB-D cameras, 34 were the Kinect (75.5\%). The most common use case involved a human-aware work space where the robot had to operate safely in a shared space ($N=26$, 46\%), followed by direct control of a robot arm ($N=15$, 27\%), object handover ($N=7$, 13\%) and collaborative manipulation ($N=3$, 5\%). For shared space work, most actions were to improve safety outcomes for the human. For instance, if a human was detected in the shared area, the robot would came to a halt~\cite{tan_safety_2010, nair_3d_2011, morato_toward_2014, shariatee_safe_2017, araiza-lllan_dynamic_2018}, slow down~\cite{zardykhan_collision_2019, bingol_practical_2020}, or change its trajectory to avoid contact~\cite{ding_optimizing_2011, moreno_path_2016, pasinetti_development_2018, landi_prediction_2019, chan_collision-free_2020, ferraguti_safety_2020, liu_dynamic_2020, mronga_constraint-based_2020, tarbouriech_bi-objective_2020}. In some shared work instances, the person was required to be in a safe standing pose before robot commands were accepted~\cite{haghighi_integration_2019}. In object handover cases, the handover process often involved both passing objects from robot-to-human~\cite{arumbakkam_multi-modal_2010, sisbot_synthesizing_2010, bdiwi_handing-over_2013, ikai_robot_2016} and from human-to-robot~\cite{scimmi_experimental_2019}. These handover actions often used force~\cite{bdiwi_handing-over_2013} or tactile sensors~\cite{ikai_robot_2016} to determine when to release the object. Another use case was robotic vision to control robot arms by matching tool center point with human hand positions~\cite{arumbakkam_multi-modal_2010, christiernin_interacting_2016, bolano_towards_2018, hong_interactive_2018, haghighi_integration_2019}. Lastly, other related works also used both robotic vision and an IMU~\cite{haghighi_integration_2019}.
Commonly used platforms were the Kuka ($N=9$), Universal Robots (UR) series ($N=8$), ABB Industrial ($N=4$), Franka Emika Panda ($N=3$), Stäubli ($N=2$), Baxter ($N=3$), Lynxmotion ($N=2$), Sawyer ($N=1$), WAM robot ($N=1$), and other types of arms ($N=19$).

\subsubsection{Use Case Examples}
Collaborative actions often involved the robot providing assistance during a specific task. In relation to specific tasks, robotic vision was used to track a person's hand, and a force sensor to sense contact in an screwing task~\cite{cherubini_unified_2015} and to match the end-effector position with hand position to pick up and pass an item~\cite{arumbakkam_multi-modal_2010}. Others included moving the tool centre point of a robot arm to follow a human hand to execute a grasp action when the human placed both hands out in front~\cite{bolano_towards_2018} or a robot performing cooperative sheet folding with a hand detected against the fabric corner with the robot arm picking up the opposite corner to perform the fold~\cite{koustoumpardis_human_2016}. Some collaborative actions took a mixed sensor approach such as joint manipulation of a wooden stick with a Franka Emika arm with the goal to keep the stick horizontal using data from a wearable IMU devices attached to the wrists, elbows and the hip fused with extracted skeleton information to provide the robot with an accurate human pose~\cite{zhang_human_2020}, a camera mounted to the end-effector of a Stäubli industrial arm to follow a human hand using visual servoing, and a force sensor to know when the robot should release the object~\cite{bdiwi_handing-over_2013}. Collaborative actions also had specific person-centered applications, such as robots to assist persons with disabilities through assistive dressing with a jacket  ~\cite{valle_personalized_2019}, soft robots to assist in bathing~\cite{dometios_real-time_2017}, and for a surgeon to instruct a robot to fetch an item during an operation~\cite{kogkas_free-view_2019}. Other applications were orientated around a specific outcome, such as to improve safety. Safety outcomes included robotic vision to identify a potential collision with a person and stop the robot in collaborative assembly task~\cite{morato_toward_2014}, to inform the robot to deactivate if a person was too close~\cite{nair_3d_2011}, to assist in consideration of proxemics when handing over a water bottle~\cite{sisbot_synthesizing_2010}, and image and torque sensing to help reduce robot arm speed based on human proximity and contact~\cite{bingol_practical_2020}. Other safety related functions included the robot using the shoulder and hand position to determine if the person was directed towards the task and if not, to pause until the person was again engaged or no person was present~\cite{bothe_effective_2018}, and robotic vision to help address ergonomics, such as to adjust the working height of the end-effector based on human pose (i.e. height and arm position)~\cite{van_den_broek_ergonomic_2020}. 

\subsubsection{Included Papers} Papers related to object handover and collaborative actions are listed here:
\cite{de_luca_integrated_2012,baron_remote_2013,tan_safety_2010,ding_optimizing_2011,nair_3d_2011,bdiwi_handing-over_2013,moe_real-time_2013,schmidt_contact-less_2013,wang_vision-guided_2013,morato_toward_2014,saveriano_safe_2014,benabdallah_kinect-based_2015,cherubini_unified_2015,fuad_skeleton_2015,nazari_simplified_2015,christiernin_interacting_2016,guo_control_2016,ikai_robot_2016,koustoumpardis_human_2016,moreno_path_2016,zhao_intuitive_2016,bouteraa_gesture-based_2017,dometios_real-time_2017,shariatee_safe_2017,araiza-lllan_dynamic_2018,bai_kinect-based_2018,bolano_towards_2018,bothe_effective_2018,hong_interactive_2018,pasinetti_development_2018,costanzo_multimodal_2019,haghighi_integration_2019,kogkas_free-view_2019,landi_prediction_2019,scimmi_experimental_2019,sun_visual_2019,svarny_safe_2019,valle_personalized_2019,zardykhan_collision_2019,zhang_gesture-based_2019,bingol_practical_2020,chan_collision-free_2020,de_schepper_towards_2020,fallahinia_comparison_2020,ferraguti_safety_2020,liu_dynamic_2020,mronga_constraint-based_2020,nascimento_collision_2020,sanchez-matilla_benchmark_2020,tarbouriech_bi-objective_2020,terreran_low-cost_2020,van_den_broek_ergonomic_2020,gao_user_2020,zhang_human_2020,arumbakkam_multi-modal_2010,sisbot_synthesizing_2010}. 
\subsection{Social Communication}\label{sec:rq2:social} 

\subsubsection{Overview}
\begin{figure}[ht!]
\centering
\subfloat[Total Papers]{\includegraphics[width=0.45\columnwidth]{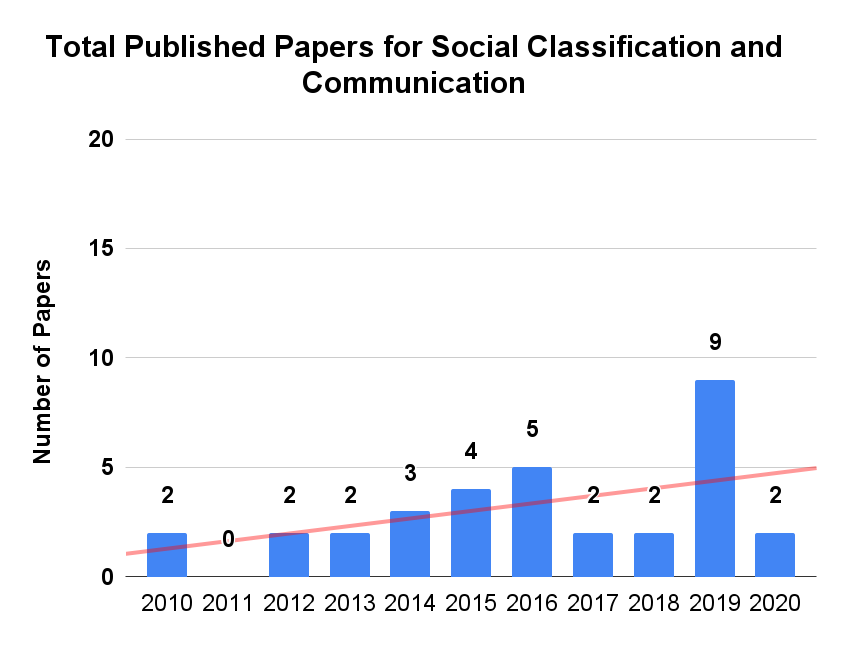}}
\hfil
\subfloat[Domains]{\includegraphics[width=0.45\columnwidth]{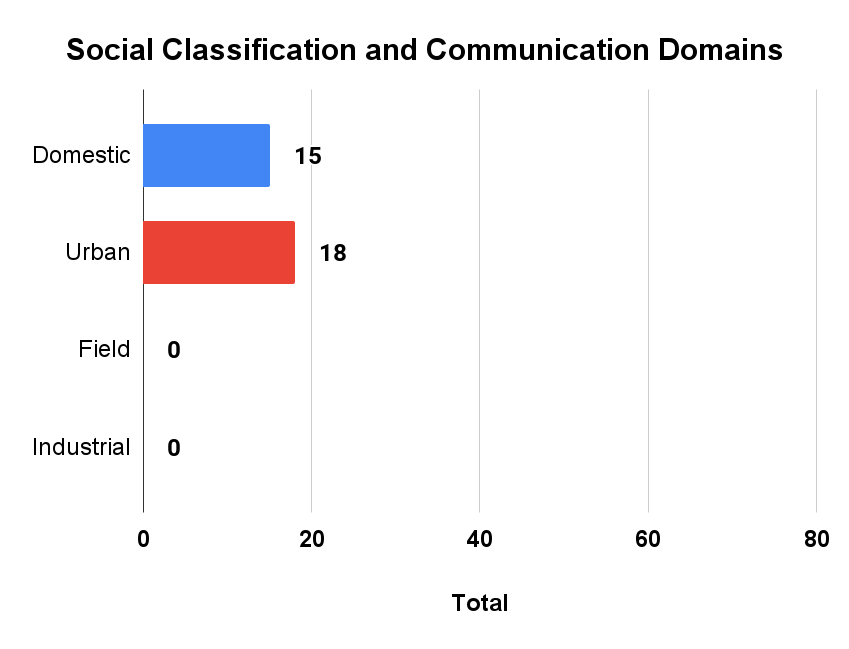}}
\hfil
\subfloat[Camera Types]{\includegraphics[width=0.45\columnwidth]{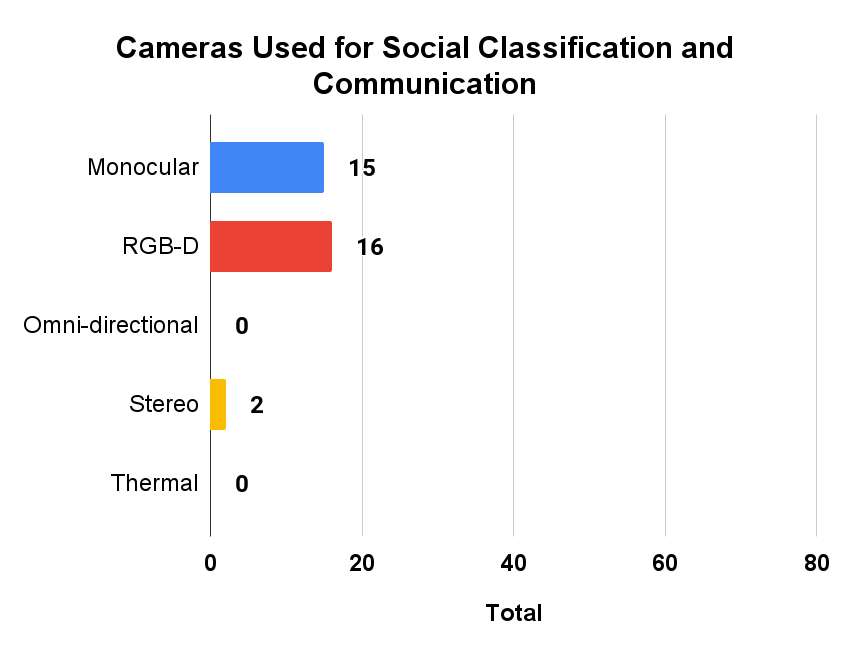}}
\hfil
\subfloat[Robot Types]{\includegraphics[width=0.45\columnwidth]{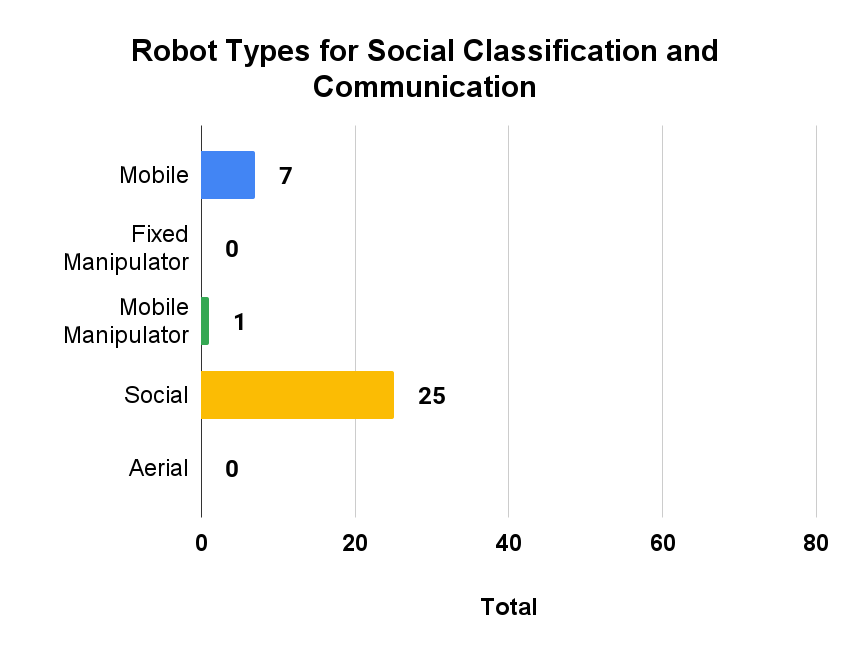}}
\hfil
\subfloat[Common Social Tasks]{\includegraphics[width=0.45\columnwidth]{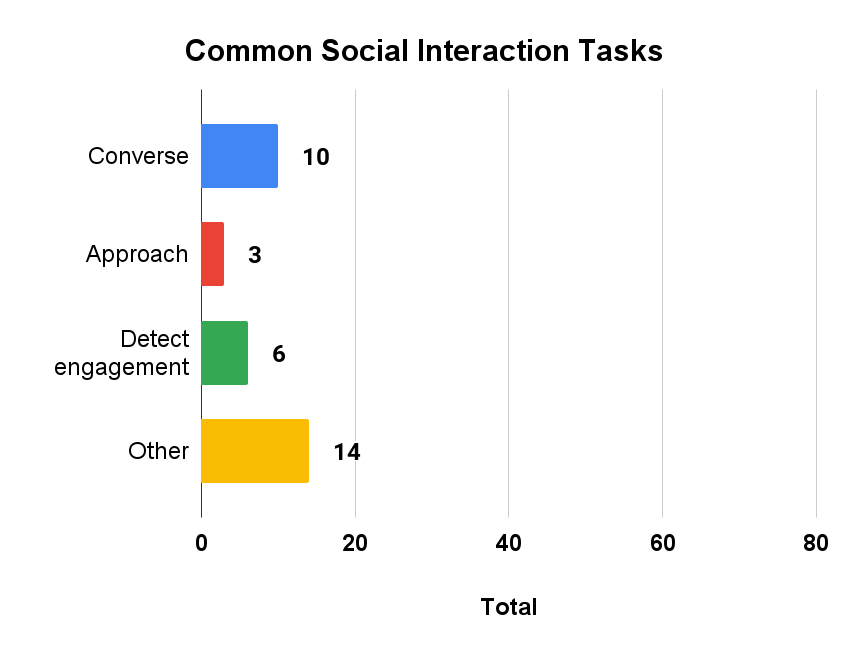}}
\hfil
\subfloat[Autonomy Level]{\includegraphics[width=0.45\columnwidth]{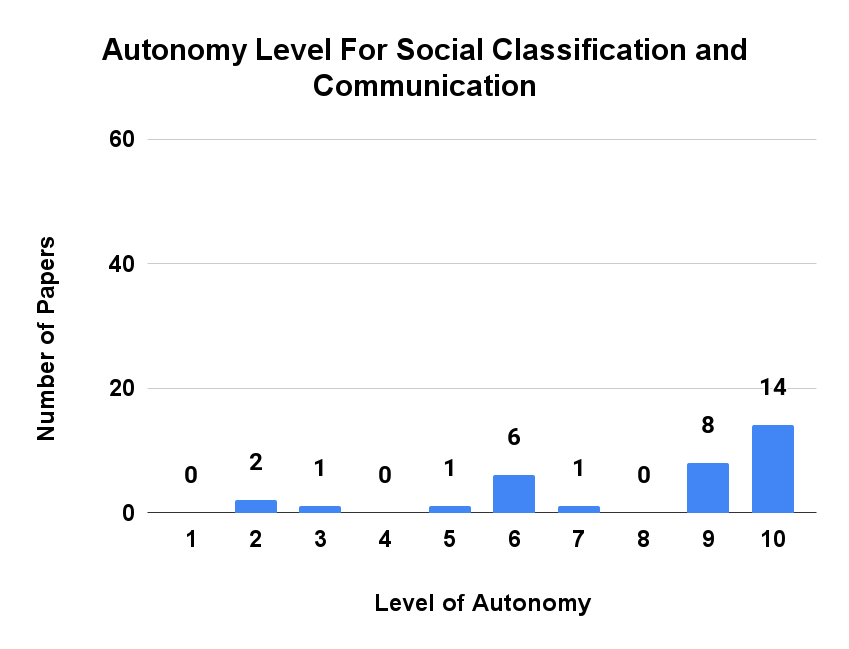}}
\hfil
\caption{Social Communication Totals and Summaries}
\label{fig:scac}
\vspace{-3mm}
\end{figure}

Categorization for social communication required that the robot needed to perform a social behaviour, or be capable of socially interacting with a person. A total of \DIFdelbegin \DIFdel{31 papers (10}\DIFdelend \DIFaddbegin \DIFadd{33 papers (11}\DIFaddend \% of the eligible total) involved a social interaction between a person and a robot. Figure~\ref{fig:scac} shows the number of social interaction works, common domains, camera types, robot types, common social tasks, and level of autonomy. In the \DIFdelbegin \DIFdel{15 }\DIFdelend \DIFaddbegin \DIFadd{16 }\DIFaddend papers that used RGB-D cameras, \DIFdelbegin \DIFdel{11 }\DIFdelend \DIFaddbegin \DIFadd{12 }\DIFaddend were the Kinect (\DIFdelbegin \DIFdel{73}\DIFdelend \DIFaddbegin \DIFadd{75}\DIFaddend \%). Common tasks required a robot to converse with a person ($N=10$, \DIFdelbegin \DIFdel{32}\DIFdelend \DIFaddbegin \DIFadd{30}\DIFaddend \%), detect social engagement ($N=6$, \DIFdelbegin \DIFdel{19}\DIFdelend \DIFaddbegin \DIFadd{18}\DIFaddend \%), or to approach people in a social way ($N=3$, \DIFdelbegin \DIFdel{10}\DIFdelend \DIFaddbegin \DIFadd{9}\DIFaddend \%). 

\subsubsection{Use Case Examples}
Social actions included having the robot face towards the person who was talking~\cite{csapo_multimodal_2012, li_visual_2015, cech_active-speaker_2015, phong_vietnamese_2020}, to detect the active speaker in a group of people using facial recognition and audio~\cite{cech_active-speaker_2015}, to commence a conversation when a person is detected~\cite{phong_vietnamese_2020}, to identify when the person had finished talking~\cite{bilac_gaze_2017, jarosz_detecting_2019}, to perform face detection and gestures during a conversation with a person~\cite{csapo_multimodal_2012}, to wave when a waving gesture was detected~\cite{gong_research_2018}, and to recognise engagement levels through facial expression and gaze detection~\cite{castellano_context-sensitive_2014, taheri_social_2014, saleh_nonverbal_2015} from head movements such as nodding or shaking~\cite{saleh_nonverbal_2015}. Other actions related to social interaction included classifying facial expressions~\cite{ke_vision_2016, li_cnn_2019}, gender and person identification~\cite{chien_navigating_2019}, as well as age and gender estimation using facial cues~\cite{cech_active-speaker_2015}. Social communication was also used for mobile robot situations, such as to detect if the person wanted to interact with it~\cite{vaufreydaz_starting_2016, mollaret_multi-modal_2016, li_inferring_2019}, and to determine social group configurations to enact appropriate social distance conventions~\cite{tseng_multi-human_2014}. Multi-person applications were also explored, such as speech and vision sensing with an iCat robot head with two arms to greet people, take orders and serve drinks from multiple people~\cite{foster_two_2012}, and Pepper robot in a restaurant where the robot was required to point to seating locations, repeat bar orders, relocate a person, and deliver an item to a person even if they had moved from their original position~\cite{lee_visual_2020}. This included other tests for if there was more than one human in the robot's field of view e.g.~\cite{tseng_multi-human_2014, li_inferring_2019}. 

\subsubsection{Included Papers} Papers related to social communication are listed here:
\cite{yu_interactive_2019,hasanuzzaman_adaptation_2010,jindai_small-size_2010,csapo_multimodal_2012,foster_two_2012,anzalone_multimodal_2013,das_attracting_2013,castellano_context-sensitive_2014,taheri_social_2014,tseng_multi-human_2014,cech_active-speaker_2015,li_visual_2015,saleh_nonverbal_2015,zhang_adaptive_2015,ke_vision_2016,li_id-match_2016,simul_support_2016,vaufreydaz_starting_2016,barz_evaluating_2017,bilac_gaze_2017,gong_research_2018,lathuiliere_deep_2018,belo_facial_2019,chien_navigating_2019,jarosz_detecting_2019,li_cnn_2019,paetzel_let_2019,lee_visual_2020,phong_vietnamese_2020,mollaret_multi-modal_2016,li_inferring_2019,bastos_robot-assisted_2019,12_intelligent_2019}. 

\subsection{Learning from Demonstration}\label{sec:rq2:learn}

\subsubsection{Overview}
\begin{figure}[ht!]
\centering
\subfloat[Total Papers]{\includegraphics[width=0.45\columnwidth]{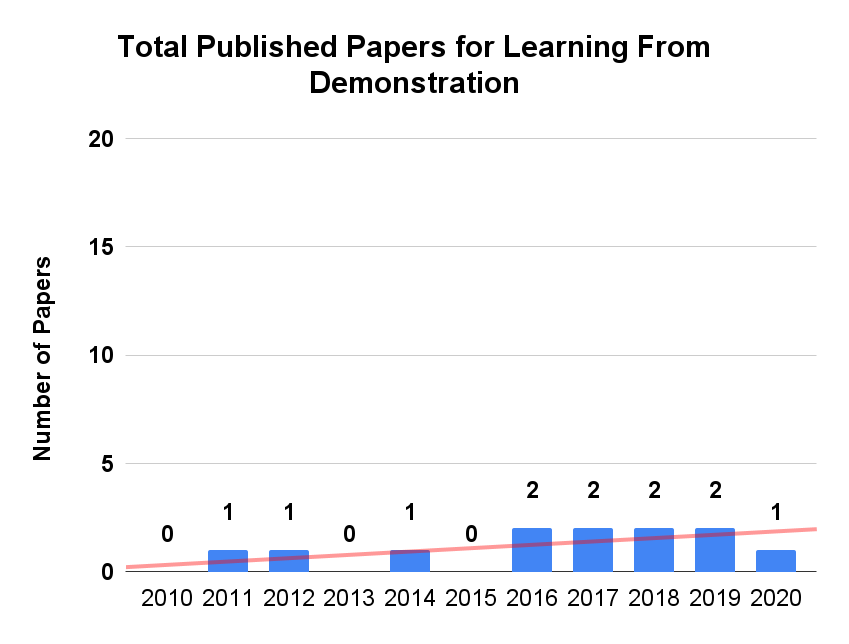}}
\hfil
\subfloat[Domains]{\includegraphics[width=0.45\columnwidth]{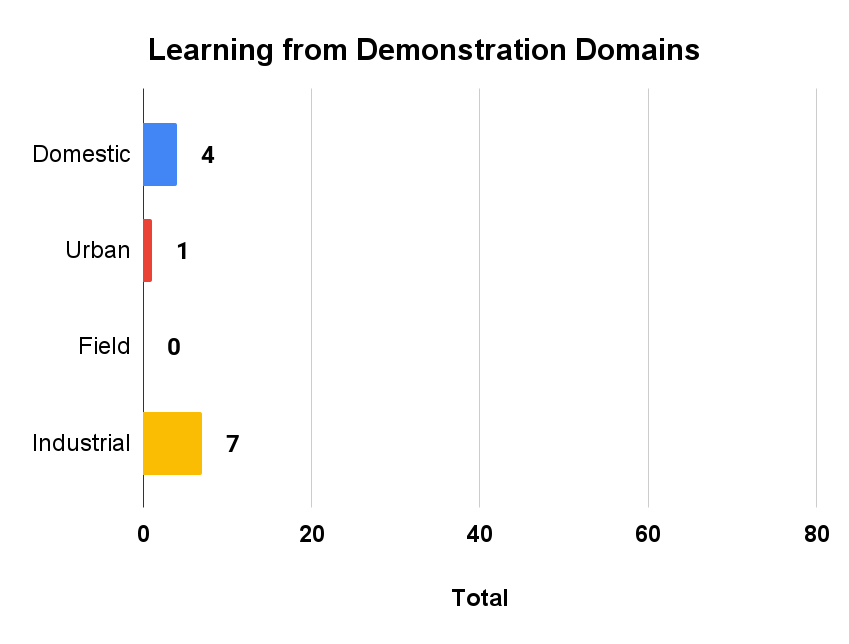}}
\hfil
\subfloat[Camera Types]{\includegraphics[width=0.45\columnwidth]{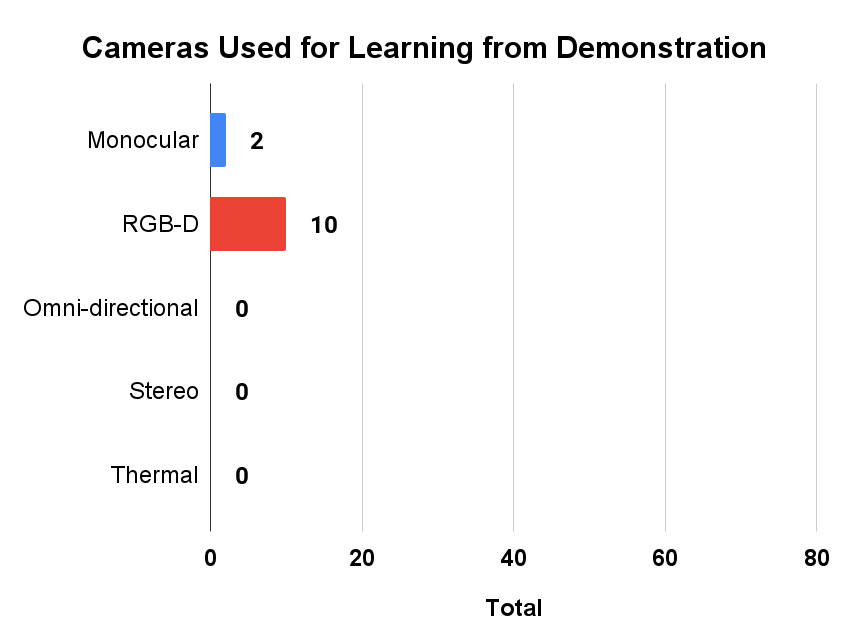}}
\hfil
\subfloat[Robot Types]{\includegraphics[width=0.45\columnwidth]{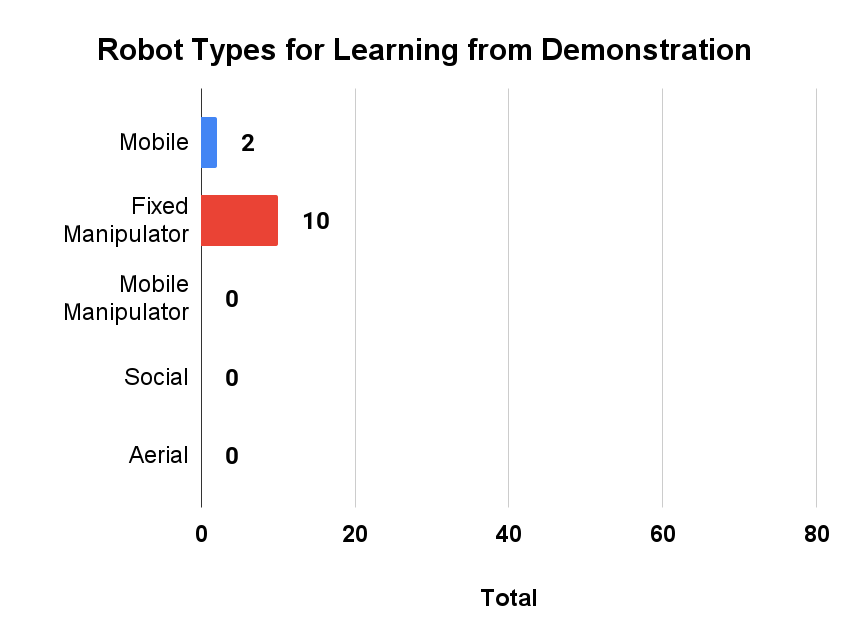}}
\hfil
\subfloat[Common Tasks]{\includegraphics[width=0.45\columnwidth]{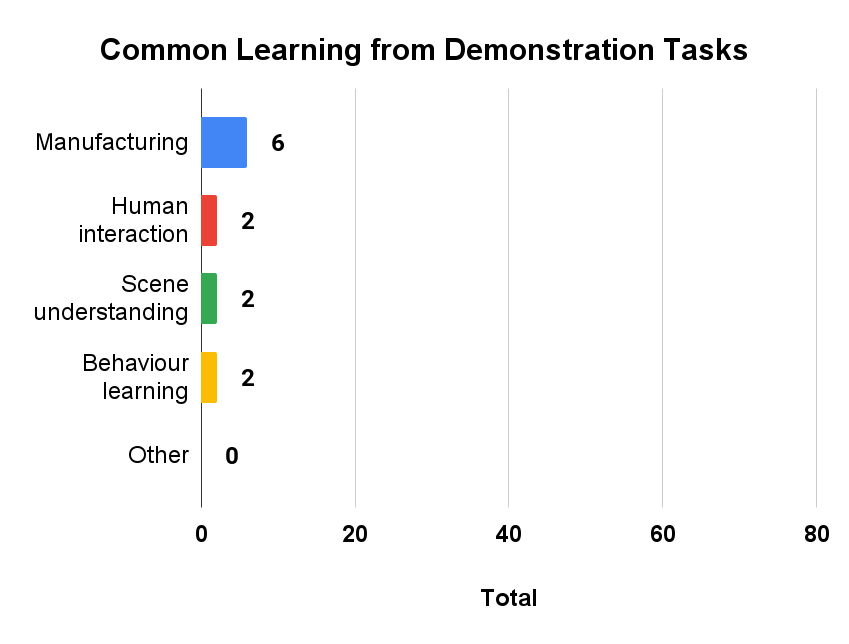}}
\hfil
\subfloat[Autonomy Level]{\includegraphics[width=0.45\columnwidth]{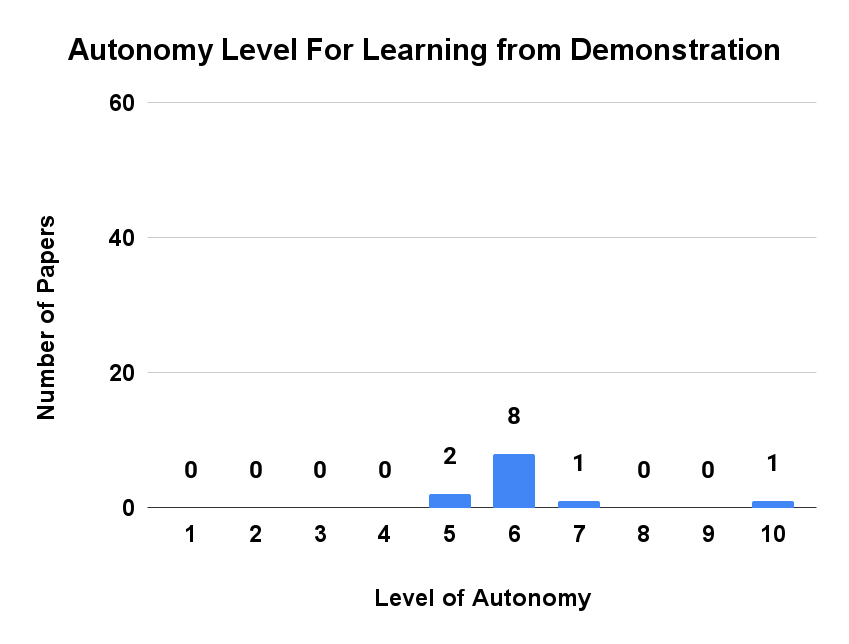}}
\hfil
\caption{Learning from Demonstration.}
\label{fig:lfd}
\vspace{-3mm}
\end{figure}

A total of 12 papers (4\% of the eligible total) involved some form of learning from demonstration. Figure~\ref{fig:lfd} shows the number of learning from demonstration works, common domains, camera types, robot types, common tasks, and level of autonomy. In the 10 papers that used RGB-D cameras, 9 were the Kinect (90\%). Learning from demonstration tasks included manufacturing assistance ($N=6$, 25\%), human interaction ($N=2$, 16\%), scene understanding ($N=2$, 16\%), and behaviour learning ($N=2$, 16\%). Robots could learn by watching a person perform a task, or through collecting data from an interaction between two people. Gestures were often used to enter a demonstration mode, including when a human would move the robot ($N=3$, 25\%), or to provide instructions ($N=2$, 16\%). Robots were often humanoid robots ($N=7$) which included the iCub ($N=2$)~\cite{saegusa_cognitive_2011, zambelli_multimodal_2016}, Pepper ($N=1$)~\cite{yu_interactive_2019}, SARCOS humanoid ($N=1$)~\cite{petric_online_2014}, imNeu ($N=1$)~\cite{li_learning_2018}, a small humanoid ($N=1$)~\cite{potdar_learning_2016}, and an anthropomorphic robot head ($N=1$)~\cite{yoo_gaze_2017}. There were 5 industrial robot arms: Kuka ($N=3$)~\cite{du_online_2018, mazhar_real-time_2019, yan_optimization_2020}, WAM robot ($N=1$)~\cite{valipour_incremental_2017}, and FANUC ($N=1$)~\cite{stipancic_programming_2012}). 

\subsubsection{Use Case Examples}
Gestures were often used in learning from demonstration tasks, such as hand and body gestures to signal the beginning and end of a demonstration~\cite{stipancic_programming_2012, mazhar_real-time_2019}, or teach the robot online~\cite{petric_online_2014, du_online_2018}. Other examples involved pointing and speech to show an industrial arm where to work~\cite{du_online_2018}, teaching the robot to perform a peg grasping task~\cite{li_learning_2018}, and to follow actions in a watering task from visual gaze~\cite{yoo_gaze_2017}. Learning methods included programming the robot to become compliant once a hand gesture command had been received to move the end effector for a chosen action, with a second gesture to signal the completion of programming, and for the robot to begin executing a new task~\cite{mazhar_real-time_2019}. Other methods included fine tuning actions when the robot end effector was close to the desired position~\cite{du_online_2018}, to change the periodic motion of a humanoid end effector~\cite{petric_online_2014} and change the motion of its hand in response to a human coaching gesture~\cite{petric_online_2014}. Learning from demonstration also included humans teaching a small humanoid~\cite{potdar_learning_2016}, or from mirrored human examples such as doing a task with a bottle by observing related actions (hold, place, and take), and learning to replicate it ~\cite{saegusa_cognitive_2011}. 

\subsubsection{Included Papers} Papers related to learning from demonstrations are listed here:
\cite{saegusa_cognitive_2011,petric_online_2014,potdar_learning_2016,zambelli_multimodal_2016,valipour_incremental_2017,yoo_gaze_2017,li_learning_2018,mazhar_real-time_2019,yu_interactive_2019,yan_optimization_2020,stipancic_programming_2012,du_online_2018}.

\section{RQ3. What is the human-robot interaction taxonomy for robotic vision in human robot collaboration and interaction?}\label{sec:rq3}
\subsection{Summary}\label{sec:rq3:summary}
\begin{figure}[h!]
\centering
\subfloat[Task Criticality]{\includegraphics[width=0.45\columnwidth]{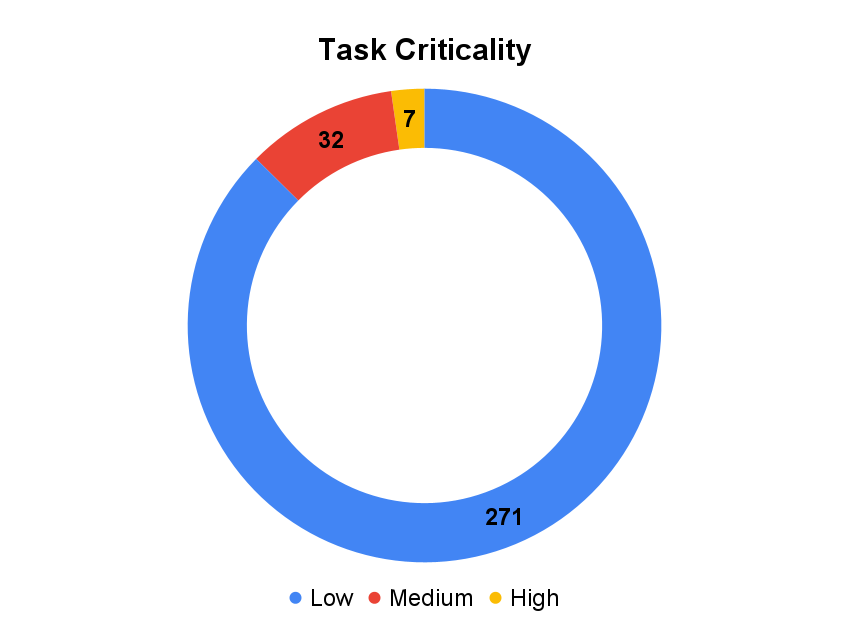}}
\hfil
\subfloat[Robot Morphology]{\includegraphics[width=0.45\columnwidth]{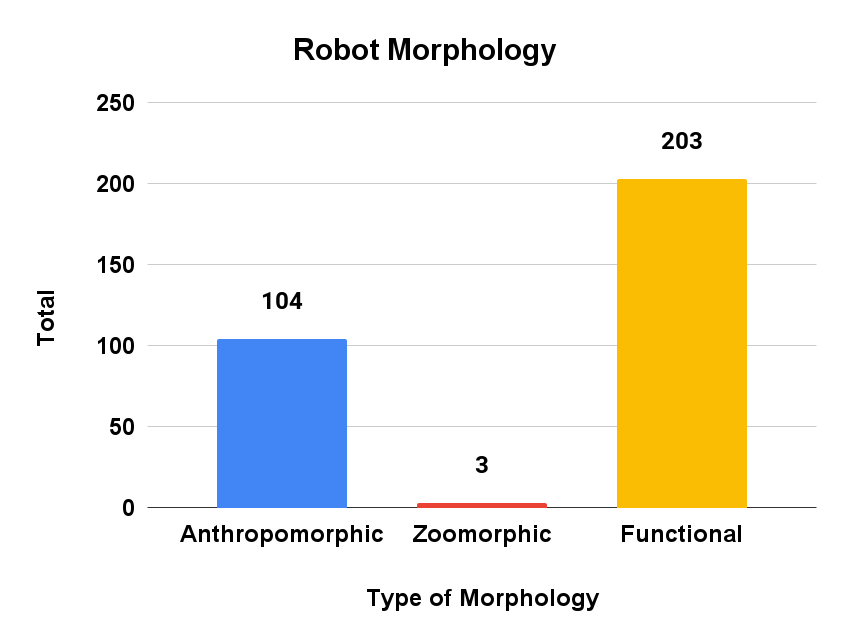}}
\hfil
\subfloat[Task Criticality and Robot Domain]{\includegraphics[width=0.45\columnwidth]{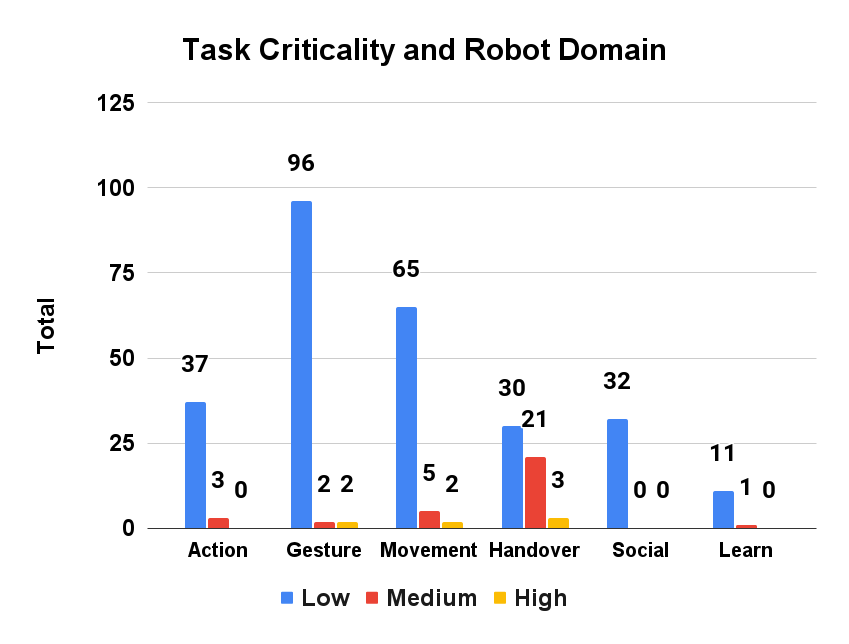}}
\hfil
\subfloat[Robot Interaction Formation]{\includegraphics[width=0.45\columnwidth]{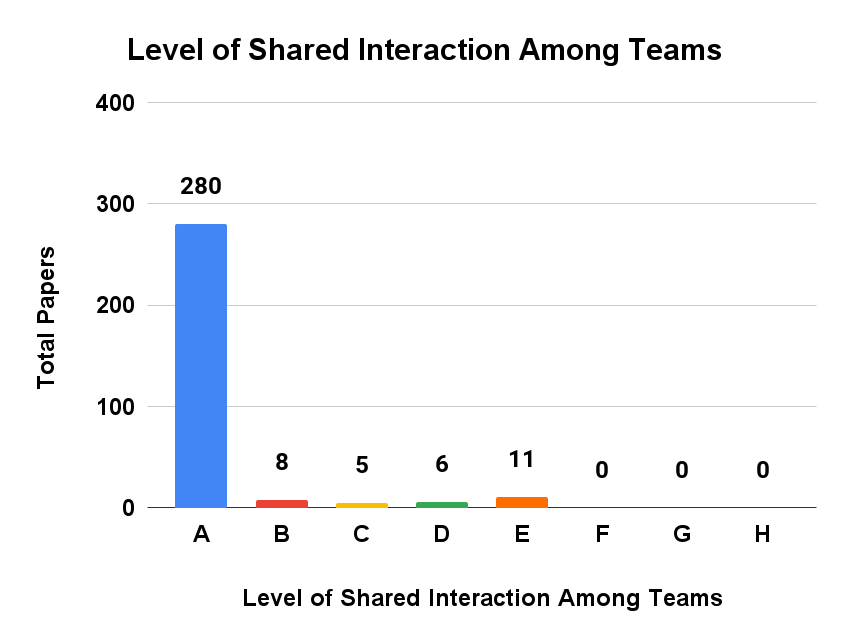}}
\hfil
\subfloat[Human-Robot Proximity]{\includegraphics[width=0.45\columnwidth]{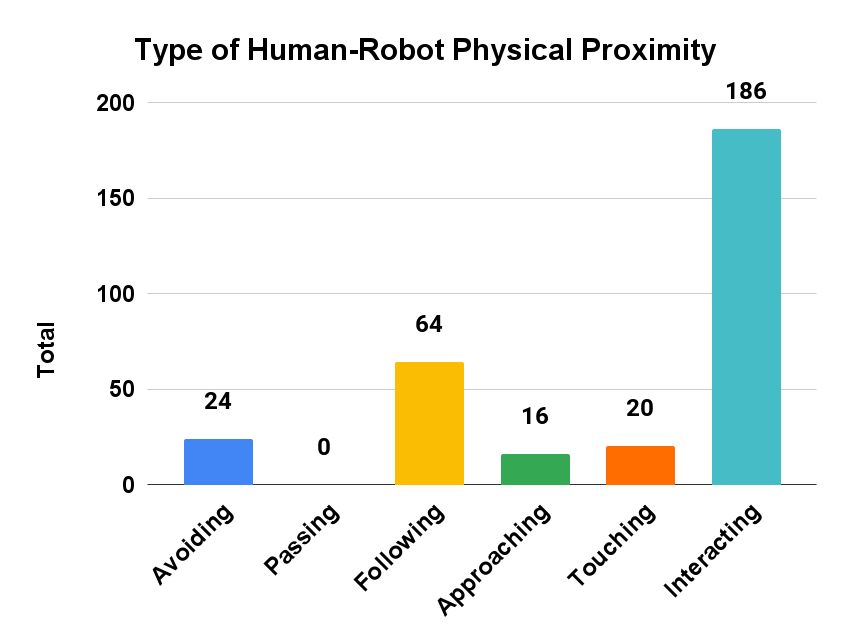}}
\hfil
\subfloat[Type of Interaction]{\includegraphics[width=0.45\columnwidth]{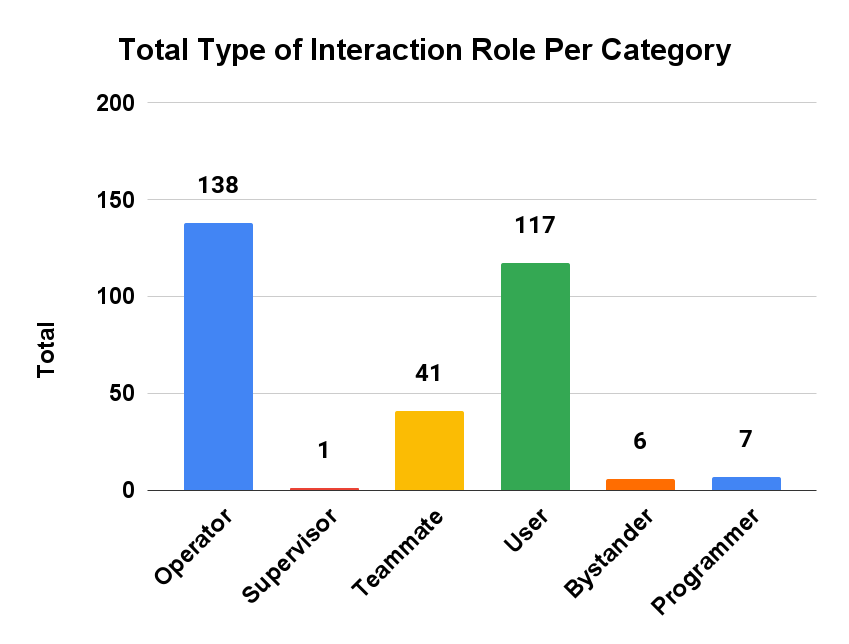}}
\hfil
\caption{Taxonomy Data Summary}
\label{fig:taxonomy}
\vspace{-3mm}
\end{figure}

\begin{figure}[h!]
\centering
\subfloat[Metrics and Domain]{\includegraphics[width=0.45\columnwidth]{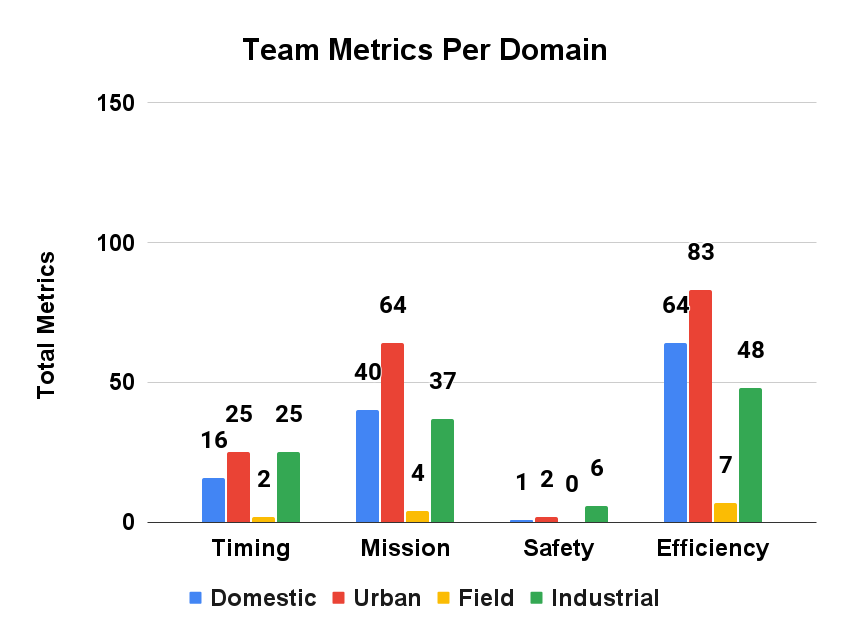}}
\hfil
\subfloat[Metrics and Robot
Type]{\includegraphics[width=0.45\columnwidth]{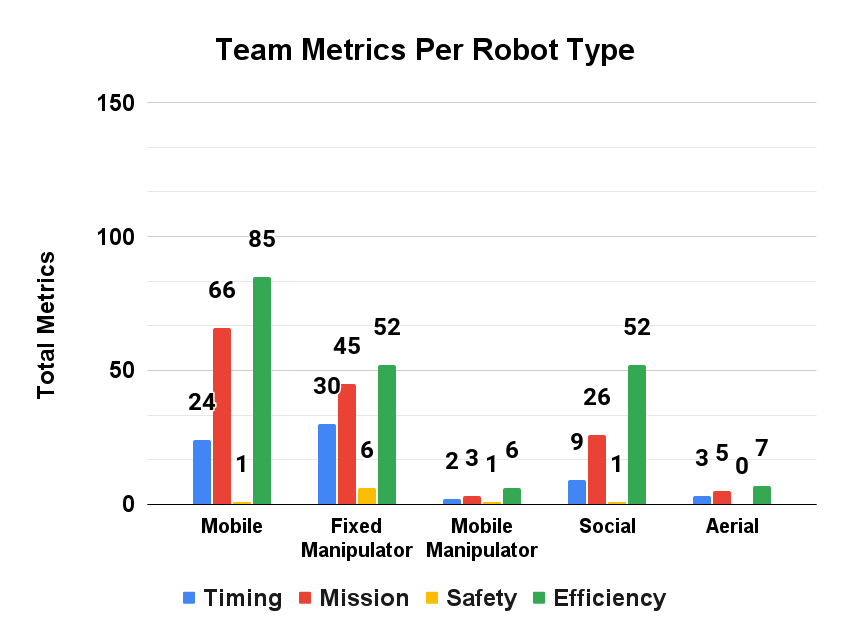}}
\hfil
\caption{Taxonomy Data Summary - Team Metrics}
\label{fig:taxonomyteam}
\vspace{-3mm}
\end{figure}

\begin{figure}[tb!]
\centering
\subfloat[Autonomy Level by Year]{\includegraphics[width=0.45\columnwidth]{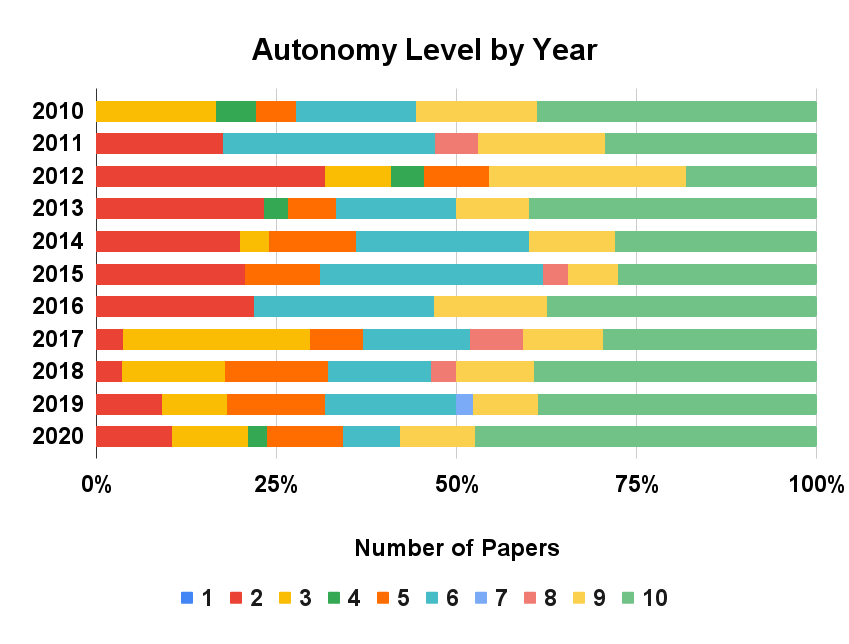}\label{fig:autoyear}}
\hfil
\subfloat[Autonomy Level by Domain]{\includegraphics[width=0.45\columnwidth]{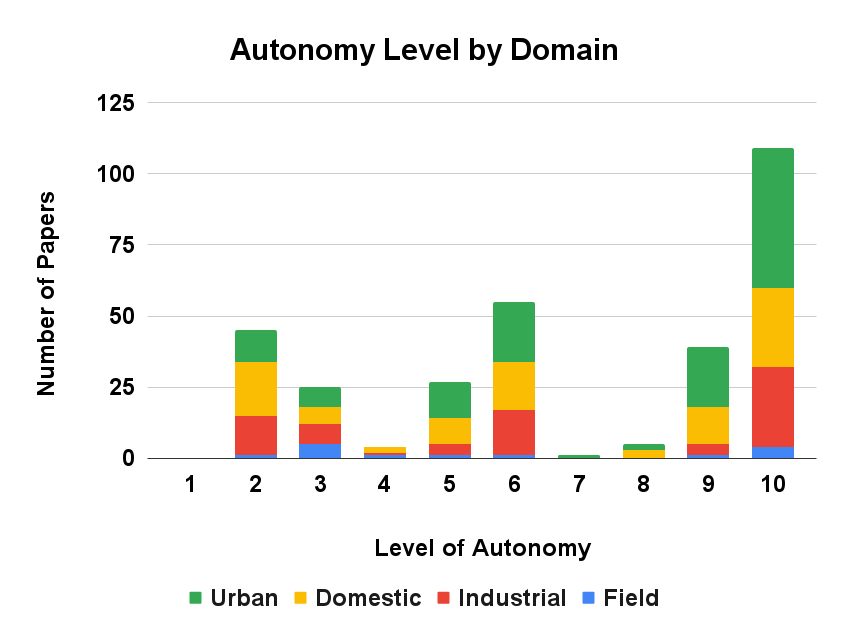}\label{fig:autodomain}}
\hfil
\subfloat[Autonomy level by Robot Type]{\includegraphics[width=0.45\columnwidth]{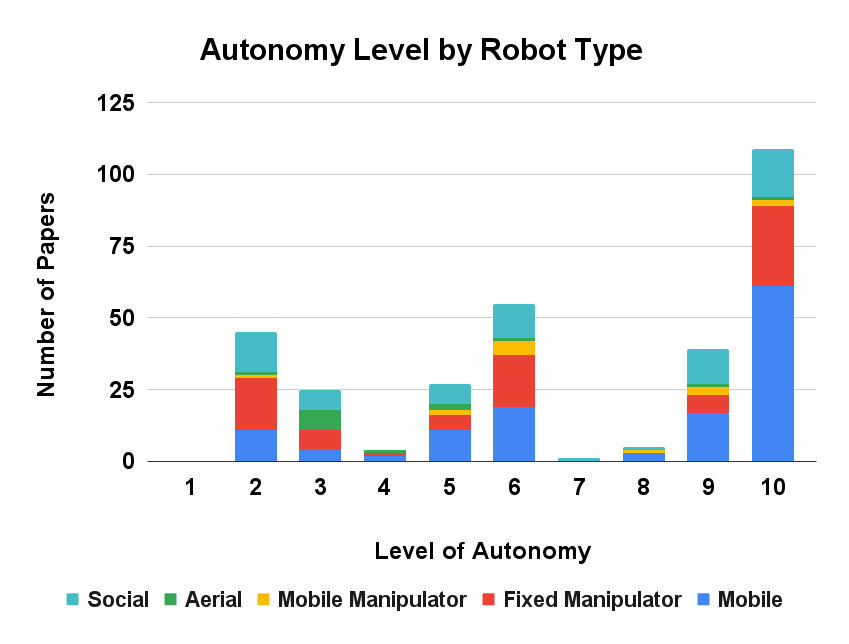}\label{fig:autorobot}}
\hfil
\subfloat[Autonomy Level by Camera Type]{\includegraphics[width=0.45\columnwidth]{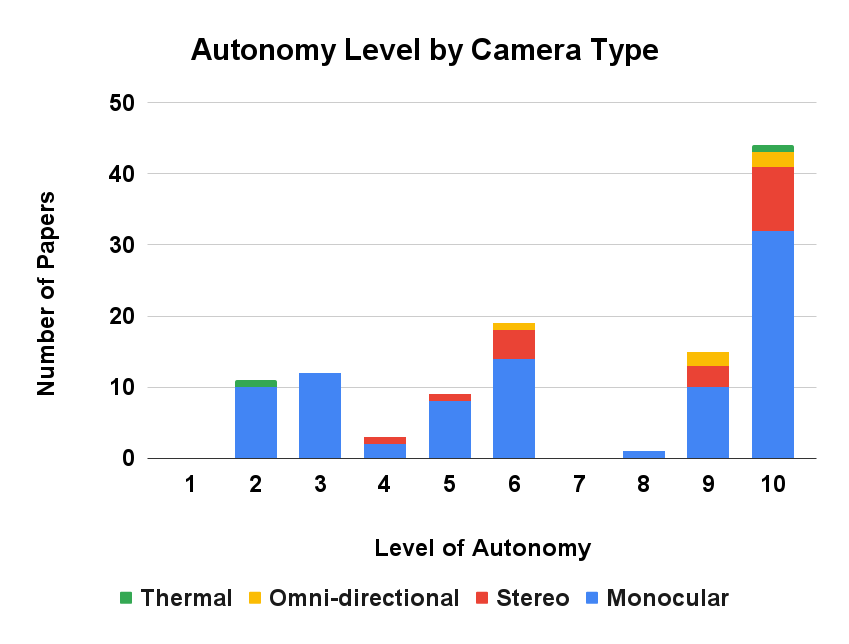}\label{fig:autocamera}}
\hfil
\caption{Autonomy Level Trends and Summaries}
\label{fig:autonomy_level}
\vspace{-3mm}
\end{figure}

This section will explore the human-robot interaction taxonomy data (See Fig~\ref{fig:taxonomy}  and \ref{fig:taxonomyteam}) as informed by human-robot interaction and robot classification taxonomies, e.g. \cite{yanco2004classifying, beer2014toward}. Detailed explanation of taxonomy hierarchy and their relevant classification labels can be found in \cite{yanco2004classifying, beer2014toward}, and a brief summary of labels has been listed in the review information and categorisation section (See Section~\ref{category}). Task type was relatively broad with limited consistency between studies, and has been reported in individual sections listed above this section (See Section \ref{sec:rq2}). Task criticality for most robot use cases was identified as low criticality (\DIFdelbegin \DIFdel{$N=265$}\DIFdelend \DIFaddbegin \DIFadd{$N=271$, 88\%}\DIFaddend ) compared to medium (\DIFdelbegin \DIFdel{$N=31$}\DIFdelend \DIFaddbegin \DIFadd{$N=32$, 10\%}\DIFaddend ) or high ($N=7$, 2\%) classification, which may further support the emergent nature of robot roles in easier use cases as a first application. There was also a link between task difficulty and frequency, where less difficult tasks are more commonly investigated, and harder tasks are less represented. These classification patterns were similar to our own custom metric on a single score for overall task evaluation using task complexity, risk, importance, and robot complexity: low (\DIFdelbegin \DIFdel{$N=242$}\DIFdelend \DIFaddbegin \DIFadd{$N=249$, 80\%}\DIFaddend ), medium (\DIFdelbegin \DIFdel{$N=49$}\DIFdelend \DIFaddbegin \DIFadd{$N=51$, 17\%}\DIFaddend ) and high ($N=10$, 3\%). Robot morphology was categorised as anthropomorphic (human-like), zoomorphic (animal-like) and functional (neither humanlike nor animal-like, but related to function). Most systems were functional (\DIFdelbegin \DIFdel{$N=198$}\DIFdelend \DIFaddbegin \DIFadd{$N=203$, 65\%}\DIFaddend ) compared to anthropomorphic (\DIFdelbegin \DIFdel{$N=102$,}\DIFdelend \DIFaddbegin \DIFadd{$N=104$, 34\%}\DIFaddend ) or zoomorphic ($N=3$, 1\%). Functional robots were more likely to be used in medium and high task \DIFdelbegin \DIFdel{critiality }\DIFdelend \DIFaddbegin \DIFadd{criticality }\DIFaddend studies compared to anthropomorphic or zoomorphic robots. A high volume of works had a 1:1 human to robot ratio (\DIFdelbegin \DIFdel{$N=274$, 90}\DIFdelend \DIFaddbegin \DIFadd{$N=281$, 91}\DIFaddend \%) with an overall mean of 1.08, 9 with more than one robot, 20 with more than one human, a maximum reported ratio as 5 \cite{li_id-match_2016} and minimum reported ratio between 0.1 \cite{alonso-mora_human_2014} and 0.07 \cite{nagi_wisdom_2015}. Human-robot teams often had 1:1 human-robot team compositions, showing that the methods and team setups focused on a single human to potentially assist in robustness and utility of robotic vision in the collaborative scenario. Homogeneous teams were used for \DIFdelbegin \DIFdel{300 }\DIFdelend \DIFaddbegin \DIFadd{307 }\DIFaddend (99\%) of the reported studies, with only 3 (1\%) of studies using heterogeneous robot team compositions, e.g. \cite{lee_visual_2020,cheng_multiple-robot_2013,lichtenstern_prototyping_2012}. Level of shared interaction among teams was high in the 'A' formation (\DIFdelbegin \DIFdel{$N=274$}\DIFdelend \DIFaddbegin \DIFadd{$N=280$}\DIFaddend , 90\%) as predicated on the earlier reported ratio of people to robots (\DIFdelbegin \DIFdel{$N=274$}\DIFdelend \DIFaddbegin \DIFadd{$N=281$}\DIFaddend , 90\%). There were 8 papers with a `B' formation (one human with multiple robots using a single interaction), \DIFdelbegin \DIFdel{4 }\DIFdelend \DIFaddbegin \DIFadd{5 }\DIFaddend with a `C' formation (one human with multiple robots using a separate interaction for each robot), 6 with a `D' formation (multiple humans with one robot, where the robot interacts with the humans through a single interaction), and 11 with an `E' formation (multiple humans with one robot, where each human interacts with the robot separately). In terms of the type of human-robot physical proximity, interacting (\DIFdelbegin \DIFdel{$N=184$, 61}\DIFdelend \DIFaddbegin \DIFadd{$N=186$, 60}\DIFaddend \%) was the highest followed by following (\DIFdelbegin \DIFdel{$N=61$, 20}\DIFdelend \DIFaddbegin \DIFadd{$N=64$, 21}\DIFaddend \%) and then avoiding (\DIFdelbegin \DIFdel{$N=23$}\DIFdelend \DIFaddbegin \DIFadd{$N=24$}\DIFaddend , 8\%). No studies that used tele-operation were eligible in this review, but for the remainder of eligible studies, nearly all (\DIFdelbegin \DIFdel{$N=302$}\DIFdelend \DIFaddbegin \DIFadd{$N=309$}\DIFaddend , 99\%) had the robot as synchronous (same time) and collocated (same place) with the exception of one study that was non-collocated, e.g. \cite{lalejini_evaluation_2015}. Autonomy level scoring by \cite{beer2014toward} was used, but no scores were classified on level 1 due to exclusion criteria that the robot must not be manually operated by the human. For the remainder, robots were often high on autonomy, which may have been skewed by initial entry criteria that required the robot to use robotic vision to perform an action or response. Figure~\ref{fig:autoyear} depicts papers over the last 10 years often had level 2 (tele-operation: robot prompted to assist but sensing and planning left to the human), level 6 (shared control with human initiative: robot senses the environment, develops plans/goals, and implements actions while the human monitors the robot’s progress) or level 10 (full autonomy: robot performs all task aspects autonomously without human intervention). Figure~\ref{fig:autodomain} depicts that there was a relatively even spread of autonomy level across the four domains, and Figure~\ref{fig:autorobot} depicts mobile and fixed manipulators were most often used with level 10 autonomy, with similar trends seen across the autonomy levels and Figure~\ref{fig:autocamera} depicts camera types per robot autonomy level. 

\subsubsection{Included Papers} Papers where a mobile robot was used are listed here:
\cite{lam_real-time_2011,paulo_vision-based_2012,gorer_autonomous_2017,vasquez_deep_2017,zhang_automating_2018,chalvatzaki_learn_2019,lee_real-time_2020,chen_integrated_2010,chen_approaches_2010,couture-beil_selecting_2010,manigandan_wireless_2010,milligan_selecting_2011,uribe_mobile_2011,yoshida_evaluation_2011,faudzi_real-time_2012,gu_human_2012,konda_real_2012,lee_interactive_2012,nishiyama_human_2013,tao_multilayer_2013,wang_wheeled_2013,alonso-mora_human_2014,nguyen_audio-visual_2014,ozgur_natural_2014,xu_online_2014,canal_gesture_2015,saichinmayi_gesture_2015,manitsaris_fingers_2016,maraj_application_2016,ju_integrative_2017,azari_commodifying_2019,li_real-time_2019,angani_human_2020,hsu_real-time_2020,yue_human-robot_2020,martin_estimation_2010,droeschel_towards_2011,park_real-time_2011,van_den_bergh_real-time_2011,abidi_human_2013,cicirelli_kinect-based_2015,jevtic_comparison_2015,lalejini_evaluation_2015,ehlers_human-robot_2016,tolgyessy_foundations_2017,lavanya_gesture_2018,mao_medical_2018,pentiuc_drive_2018,zhang_interactive_2018,fang_vehicle-mounted_2019,waskito_wheeled_2020,xu_skeleton_2020,budiharto_indoor_2010,fahn_real-time_2010,kobayashi_people_2010,mendez-polanco_detection_2010,weiss_robots_2010,yun_robust_2010,jia_autonomous_2011,pustianu_mobile_2011,luo_human_2010,alvarez-santos_feature_2012,chen_person_2012,lee_human_2012,mead_probabilistic_2012,wu_accompanist_2012,ferrer_robot_2013,granata_human_2013,hegger_people_2013,pereira_human-robot_2013,weinrich_appearance-based_2013,yun_robotic_2013,zhao_interactive_2013,chen_stereovision-only_2014,do_human-robot_2014,fujii_gesture_2014,hu_design_2014,munaro_fast_2014,prediger_robot-supported_2014,scheggi_human-robot_2014,ali_improved_2015,batista_probabilistic_2015,gupta_robust_2015,voisan_ros-based_2015,bayram_audio-visual_2016,hassan_computationally_2016,kahily_real-time_2016,mateus_human-aware_2016,talebpour-board_2016,vasconcelos_socially_2016,zhang_optimal_2016,angonese_multiple_2017,bellarbi_social_2017,ghandour_human_2017,rehman_target_2017,almonfrey_flexible_2018,long_kinect-based_2018,weber_follow_2018,yuan_development_2018,zhang_indoor_2018,ali_smart_2019,chen_human-following_2019,condes_person_2019,gemerek_video-guided_2019,luo_real-time_2019,miller_self-driving_2019,wu_improved_2019,yang_socially-aware_2019,zhang_vision-based_2019,anuradha_human_2020,hwang_interactions_2020,kawasaki_multimodal_2020,muller_multi-modal_2020,pang_efficient_2020,wu_toward_2020,chou_development_2020,hasanuzzaman_adaptation_2010,tseng_multi-human_2014,vaufreydaz_starting_2016,chien_navigating_2019,phong_vietnamese_2020,mollaret_multi-modal_2016,li_inferring_2019,brookshire_person_2010,pennisi_multi-robot_2015,gardel_wireless_2016,rosa_integration_2016,lovon-ramos_people_2016}. Papers where a fixed manipulator was used are listed here:
\cite{lee_learning_2017,wang_collision-free_2017,celik_development_2012,de_luca_integrated_2012,baron_remote_2013,hartmann_feasibility_2013,li_cyber-physical_2013,chao_robotic_2014,farulla_real-time_2014,quintero_interactive_2014,katsuki_high-speed_2015,bras_gesture_2018,yu_efficiency_2019,chen_design_2020,lambrecht_spatial_2012,choudhary_real_2015,quintero_visual_2015,yang_real-time_2015,arenas_deep_2017,deepan_raj_static_2017,maurtua_natural_2017,song_towards_2017,lima_real-time_2019,martin_real-time_2019,vysocky_interaction_2019,saegusa_cognitive_2011,petric_online_2014,valipour_incremental_2017,li_learning_2018,mazhar_real-time_2019,yan_optimization_2020,stipancic_programming_2012,du_online_2018,tan_safety_2010,ding_optimizing_2011,nair_3d_2011,bdiwi_handing-over_2013,moe_real-time_2013,schmidt_contact-less_2013,wang_vision-guided_2013,morato_toward_2014,saveriano_safe_2014,benabdallah_kinect-based_2015,cherubini_unified_2015,fuad_skeleton_2015,nazari_simplified_2015,christiernin_interacting_2016,ikai_robot_2016,koustoumpardis_human_2016,moreno_path_2016,zhao_intuitive_2016,bouteraa_gesture-based_2017,dometios_real-time_2017,shariatee_safe_2017,araiza-lllan_dynamic_2018,bai_kinect-based_2018,bolano_towards_2018,bothe_effective_2018,hong_interactive_2018,pasinetti_development_2018,costanzo_multimodal_2019,haghighi_integration_2019,kogkas_free-view_2019,landi_prediction_2019,scimmi_experimental_2019,sun_visual_2019,svarny_safe_2019,valle_personalized_2019,zardykhan_collision_2019,zhang_gesture-based_2019,bingol_practical_2020,chan_collision-free_2020,fallahinia_comparison_2020,ferraguti_safety_2020,liu_dynamic_2020,mronga_constraint-based_2020,nascimento_collision_2020,sanchez-matilla_benchmark_2020,tarbouriech_bi-objective_2020,terreran_low-cost_2020,van_den_broek_ergonomic_2020,gao_user_2020,zhang_human_2020}. Papers that used a mobile manipulator are listed here:
\cite{koppula_anticipating_2016,broccia_gestural_2011,qian_visually_2013,burger_two-handed_2012,fareed_gesture_2015,luo_human-robot_2019,moh_gesture_2019,sriram_mobile_2019,guo_control_2016,de_schepper_towards_2020,arumbakkam_multi-modal_2010,sisbot_synthesizing_2010,cheng_multiple-robot_2013,lee_visual_2020}. Papers that used an aerial robot are listed here:
\cite{lichtenstern_prototyping_2012,sanna_kinect-based_2012,pfeil_exploring_2013,monajjemi_hri_2013,costante_personalizing_2014,monajjemi_uav_2015,nagi_wisdom_2015,maher_realtime_2017,mohaimenianpour_hands_2018,shakev_autonomous_2018,chen_online_2019,medeiros_human-drone_2020,miyoshi_above_2014,yao_monocular_2017}. Papers that used a social robot are listed here:
\cite{sosnowski_mirror_2010,igorevich_behavioral_2011,masmoudi_expressive_2011,torres_implementation_2012,cid_real_2013,indrajit_development_2013,mohammad_tele-operation_2013,werner_evaluation_2013,yang_study_2013,obo_robot_2015,saffar_context-based_2015,agrigoroaie_enrichme_2016,liu_interactive_2016,silva_mirroring_2016,zhu_real-time_2016,meghdari_real-time_2017,vignolo_computational_2017,zhu_robust_2017,devanne_co-design_2018,efthymiou_multi-_2018,abiddin_development_2019,cazzato_real-time_2019,sripada_teleoperation_2019,augello_towards_2020,avioz-sarig_robotic_2020,lang_research_2020,lu_research_2020,gui_teaching_2018,sorostinean_activity_2018,hafiane_3d_2013,barros_real-time_2014,xiao_human-robot_2014,han_human_2017,luo_tracking_2011,cheng_design_2012,gori_all_2012,shieh_fuzzy_2014,gao_humanoid_2015,vircikova_teach_2015,ratul_gesture_2016,kalidolda_towards_2018,yuan_natural_2020,potdar_learning_2016,zambelli_multimodal_2016,yoo_gaze_2017,yu_interactive_2019,jindai_small-size_2010,csapo_multimodal_2012,foster_two_2012,anzalone_multimodal_2013,das_attracting_2013,castellano_context-sensitive_2014,taheri_social_2014,cech_active-speaker_2015,li_visual_2015,saleh_nonverbal_2015,zhang_adaptive_2015,ke_vision_2016,li_id-match_2016,simul_support_2016,barz_evaluating_2017,bilac_gaze_2017,gong_research_2018,lathuiliere_deep_2018,belo_facial_2019,jarosz_detecting_2019,li_cnn_2019,paetzel_let_2019,bastos_robot-assisted_2019,12_intelligent_2019,santos_copyrobot_2020}.

\section{RQ4. What are the vision techniques and tools used in human-robot collaboration and interaction?}\label{sec:rq4}

This section provides a detailed review of robotic vision techniques used in selected papers, including methods, algorithms, data sets, cameras and methods to allow robots to provide information or take action. Common techniques are discussed in detail to provide clear trends on how methods and techniques have been adapted from computer vision to robotic vision problems\DIFdelbegin \DIFdel{, but }\DIFdelend \DIFaddbegin \DIFadd{. However, many }\DIFaddend emergent techniques from computer vision \DIFdelbegin \DIFdel{were still }\DIFdelend \DIFaddbegin \DIFadd{are }\DIFaddend yet to be seen in HRI/C \DIFdelbegin \DIFdel{tests}\DIFdelend \DIFaddbegin \DIFadd{works}\DIFaddend . The use of these techniques may create many new opportunities for robots to help people in domains that have not yet been explored due to technical challenges, and to the speed or accuracy needed to be useful to the person\DIFaddbegin \DIFadd{, as discussed in Section~\ref{sec:future}}\DIFaddend . However, this level of development and testing will likely experience a translation delay from the advances seen in computer vision, given the need for human study approvals, extensive testing on hardware components that are subject to error and malfunction, and robust results that can meet peer-review standards for publication. 

\DIFaddbegin \DIFadd{This research question is important to the field of HRI/C because understanding the visual world is fundamental for interacting with the environment and its actors. It is important for the HRI/C researcher to understand the computer vision techniques and tools that have been used so far, since these are often robust, well-understood, and well-suited to HRI/C applications. More significantly, these patterns of usage allow us to identify areas for improvement, where an over-reliance on traditional or proprietary techniques might be inhibiting progress.}

\DIFaddend To begin, robot vision requires the robot to perceive a human and/or their actions to be able to provide a function, task or service to the person during human-robot collaboration and interaction. Robotic vision is often performed in a two-step process. First, localisation detects where the human is located in the robot's field of view, often to the granularity of the position of specific parts of the body, such as the hands. Second, classification determines what gesture, action, or expression is being shown by the person. This can include tracking across image frames to help resolve actions that are ambiguous or reliant on motion cues, and to facilitate continuous interaction between the person and the robot. This visual information can then provide the robot with important information that can help the robot to determine its next action or movement, making the vision process central to the function and utility of the robot to the person. Multiple sensors are also commonly used to provide different types of relevant information, especially \DIFdelbegin \DIFdel{both }\DIFdelend \DIFaddbegin \DIFadd{the combination of }\DIFaddend colour and depth measurements. The next section will discuss implemented solutions for human detection and pose estimation, gesture classification, action classification, tracking, and multi-sensor fusion. A summary of camera types including relevant information and examples are listed in Table\DIFdelbegin \DIFdel{1. 
}\DIFdelend \DIFaddbegin \DIFadd{~\ref{tab:cameras_table}. 
}\DIFaddend 

\begin{table}[!htp]\centering
\caption{Camera types represented in the corpus of papers and their properties.}\label{tab:cameras_table}
\scriptsize
\begin{tabular}{p{0.15\columnwidth} | p{0.15\columnwidth} | p{0.15\columnwidth} | p{0.1\columnwidth} | p{0.11\columnwidth} | p{0.05\columnwidth} | p{0.11\columnwidth}}\toprule
\textbf{Camera} &\textbf{RGB resolution} &\textbf{Depth resolution} &\textbf{Depth range} &\textbf{Field-of-view (degrees)} &\textbf{Frame rate} &\textbf{Examples} \\\midrule

Microsoft Kinect & 1920 $\times$ 1080 & 512 $\times$ 424 & 0.5--4.5 m & 70 $\times$ 60 & 30 fps & \cite{lu_research_2020, liu_dynamic_2020, zhang_human_2020}\\

Intel RealSense & 1920 $\times$ 1080 & 1280 $\times$ 720 & 0.3--3 m & 69 $\times$ 42 & 30 fps & \cite{silva_mirroring_2016, azari_commodifying_2019, paetzel_let_2019}\\

PrimeSense & 1280 $\times$ 960 & 640 $\times$ 480 & 0.35--3 m & 54 $\times$ 45 & 30 fps & \cite{talebpour-board_2016, zhang_gesture-based_2019,talebpour-board_2016}\\

Asus Xtion Pro Live  & 1920 $\times$ 1080 & 640 $\times$ 480 & 0.5--6 m & 57 $\times$ 44  & 30 fps & \cite{agrigoroaie_enrichme_2016, mronga_constraint-based_2020, tarbouriech_bi-objective_2020}\\

Logitech c9xx & 1920 $\times$ 1080 & -- & -- & 78 & 30 fps & \cite{van_den_broek_ergonomic_2020, barros_real-time_2014, castellano_context-sensitive_2014}\\

\bottomrule
\end{tabular}
\end{table}

\subsection{Human Detection and Pose Estimation}

Detection of the location of the person or people in a stream of visual data is often the first step for reasoning about them. Pose estimation techniques go beyond coarse detection (e.g., of bounding boxes) to find the location of body parts and their connections (i.e., skeleton estimation).

\subsubsection{Commercial Software}
The most common approach from the corpus of selected papers was to apply skeleton extraction and tracking software to depth images from RGB-D cameras (\DIFdelbegin \DIFdel{$N=106$}\DIFdelend \DIFaddbegin \DIFadd{$N=107$}\DIFaddend ), which combines detection and pose estimation. This software was primarily sourced from the Microsoft SDK for the Kinect camera or OpenNI for PrimeSense cameras. This approach has the advantage of using commercial-grade software that is easily available, real-time, and robust. Example papers include~\cite{broccia_gestural_2011, lichtenstern_prototyping_2012, cicirelli_kinect-based_2015,fang_vehicle-mounted_2019, chen_human-following_2019, ghandour_human_2017, long_kinect-based_2018, yuan_development_2018, miller_self-driving_2019, wu_improved_2019, yang_socially-aware_2019, anuradha_human_2020, christiernin_interacting_2016, wang_vision-guided_2013, morato_toward_2014, cherubini_unified_2015, koustoumpardis_human_2016, moreno_path_2016, araiza-lllan_dynamic_2018, bothe_effective_2018, hong_interactive_2018, landi_prediction_2019, scimmi_experimental_2019, valle_personalized_2019, liu_dynamic_2020, zhang_human_2020,  chan_collision-free_2020, ferraguti_safety_2020, stipancic_programming_2012, petric_online_2014, potdar_learning_2016, li_learning_2018, du_online_2018, gorer_autonomous_2017, sorostinean_activity_2018, zhang_automating_2018, lang_research_2020, igorevich_behavioral_2011, torres_implementation_2012, csapo_multimodal_2012, indrajit_development_2013, yang_study_2013, taheri_social_2014, tseng_multi-human_2014, mollaret_multi-modal_2016, meghdari_real-time_2017, augello_towards_2020, lu_research_2020, phong_vietnamese_2020, zhu_robust_2017, bellarbi_social_2017}.

\subsubsection{Face Detection}
The next most common method (especially for earlier papers) was to detect a person by first detecting their face using the Viola--Jones method~\cite{viola_robust_2004} ($N=34$). This approach uses Haar filters to extract features from an image, followed by AdaBoost~\cite{freund1997decision} to make predictions using a cascade of classifiers. This approach has the advantage of being extremely fast and performance, without requiring depth information. Depth can be used if available to disambiguate false positives based on the realistic size of a face~\cite{zhang_optimal_2016}. Example papers that use this method include~\cite{couture-beil_selecting_2010, martin_estimation_2010, droeschel_towards_2011, lam_real-time_2011, van_den_bergh_real-time_2011,pereira_human-robot_2013, budiharto_indoor_2010, luo_human_2010, kobayashi_people_2010, wu_accompanist_2012, ferrer_robot_2013, do_human-robot_2014, ali_improved_2015, bayram_audio-visual_2016, yao_monocular_2017, condes_person_2019, arumbakkam_multi-modal_2010, yoo_gaze_2017, masmoudi_expressive_2011, csapo_multimodal_2012, saleh_nonverbal_2015, ke_vision_2016, gui_teaching_2018, li_cnn_2019}. Once the face location is identified, this was sometimes used to help detect other body parts, such as hands \cite{droeschel_towards_2011, luo_tracking_2011, milligan_selecting_2011}.

\subsubsection{Segmentation}
Colour segmentation is a simple and fast method to distinguish regions containing skin from a video stream. However, colour segmentation is less robust than many other methods to different skin tones, scene colours, and occlusions such as clothing and hair. Segmentation is typically performed in HSV ~\cite{manigandan_wireless_2010, lambrecht_spatial_2012, sriram_mobile_2019}, or YCrCb~\cite{luo_tracking_2011, choudhary_real_2015, xu_skeleton_2020} colour spaces. This includes segmentation using skin threshold values~\cite{chen_approaches_2010, luo_tracking_2011, choudhary_real_2015, foster_two_2012}, or by using histogram analysis from a skin sample~\cite{manigandan_wireless_2010, lambrecht_spatial_2012, xu_skeleton_2020}. For example, Xu et al.~\cite{xu_skeleton_2020} used colour segmentation thresholds tuned for human skin tones and incorporated depth to remove segmentation errors. Accurate colour segmentation facilitated by using coloured items was also used, such as clothing~\cite{chen_integrated_2010, yoshida_evaluation_2011, faudzi_real-time_2012, hassan_computationally_2016}, or coloured tape around the person's body~\cite{lavanya_gesture_2018}. For example, Miyoshi et al.~\cite{miyoshi_above_2014} had an aerial robot follow a glove of a known colour that can be easily segmented. Given a segmented image, morphology and edge detection operations can be used to approximately extract the person from the image~\cite{shariatee_safe_2017}. Depth segmentation is an alternative that relies on some assumptions about the scene and setup~\cite{droeschel_towards_2011, paulo_vision-based_2012, mazhar_real-time_2019, xu_skeleton_2020}. For example, Paulo et al.~\cite{paulo_vision-based_2012} segmented hand gestures by setting minimum and maximum distances to the sensor, and Mazhar et al.~\cite{mazhar_real-time_2019} expanded a hand keypoint to a hand segmentation using depth.

\subsubsection{Region of Interest}
Regions of interest, such as a bounding box around a body, face or hands, can also be acquired using deep learning techniques. The most prevalent models include the region-based convolutional neural network (R-CNN) family of architectures~\cite{girshick_fast_rcnn}, and the single shot \DIFdelbegin \DIFdel{detector (SSD) network~\mbox{
\cite{liu_ssd_2016}}\hspace{0pt}
}\DIFdelend \DIFaddbegin \DIFadd{detectors~\mbox{
\cite{liu_ssd_2016, redmon2016you}}\hspace{0pt}
}\DIFaddend . These approaches are fast, robust, and are able to detect multiple people in a single image. \DIFaddbegin \DIFadd{The R-CNN family are two-stage networks that generate many object proposals, before filtering and classifying them, whereas single-stage networks predict the final bounding boxes in one go. This saves computation time at the expense of accuracy.
}\DIFaddend From the corpus of selected papers, Vasquez et al.~\cite{vasquez_deep_2017} use Fast R-CNN to extract a human bounding box from an image of a person using a walking aid. The SSD network is used in other papers to locate bodies~\cite{condes_person_2019, hwang_interactions_2020} or faces~\cite{weber_follow_2018, li_inferring_2019, cazzato_real-time_2019} in an image. For example, Weber et al.~\cite{weber_follow_2018} used an SSD network for face detection, fit faces to a deformable template model and track the detections using large-scale direct monocular SLAM (LSD-SLAM)~\cite{fleet_lsd-slam_2014}.
\DIFaddbegin \DIFadd{While there are many other methods for detecting humans that have been deployed on robots, such as detection from 2D range data \mbox{
\cite{arras2007using,beyer2018deep,jia2020dr}}\hspace{0pt}
, these approached were not present in the corpus of papers.
}\DIFaddend 

\subsubsection{Pose Estimation}
Deep learning is also commonly used for \DIFaddbegin \DIFadd{human }\DIFaddend pose estimation from RGB images. \DIFdelbegin \DIFdel{This method is used to extract human }\DIFdelend \DIFaddbegin \DIFadd{Image-based pose estimation usually refers to extracting }\DIFaddend keypoints and skeletons from \DIFdelbegin \DIFdel{images without }\DIFdelend \DIFaddbegin \DIFadd{an image of one or more people, without any additional }\DIFaddend depth information. \DIFaddbegin \DIFadd{Out of the HRI/C corpus surveyed, }\DIFaddend Convolutional Pose Machines (CPM)~\cite{wei_convolutional_2016} and OpenPose~\cite{Cao_2017_CVPR} were used most often ($N=15$) for robotic vision. The OpenPose approach trains a network to predict the location of all joints in the image, and then assembles skeletons using learned part affinity fields in a post-processing step. The main advantage of this bottom-up approach is that it runs in real-time, and runtime is independent of the number of people in the image, making it particularly suitable for human-robot interaction. \DIFaddbegin \DIFadd{An additional explanation for its prevalence among HRI/C papers is that it has a well-known, well-maintained, and high-quality codebase that is user-friendly and publicly available. }\DIFaddend Example papers can be seen here~\cite{martin_real-time_2019, mazhar_real-time_2019, miller_self-driving_2019, zardykhan_collision_2019, van_den_broek_ergonomic_2020, chalvatzaki_learn_2019, gui_teaching_2018, lee_real-time_2020}. 

\subsection{Gesture Classification}

For human-robot interaction or collaboration, the human is often required to perform a specific hand or body configuration to interact with the robot. Once the region of interest or skeleton is extracted, hand and body gestures are classified through a variety of methods, depending on the dynamic or static nature of the gestures being performed. For static hand gestures, classification was often performed from segmented images by determining the contours (boundary pixels of a region) and the convex hull (smallest convex polygon to contain the region) as a method of counting the number of fingers being held up~\cite{lambrecht_spatial_2012,  choudhary_real_2015, deepan_raj_static_2017,  sriram_mobile_2019, xu_skeleton_2020}. The distance of each contour point to the centroid of the segmented image was also used for finger counting~\cite{manigandan_wireless_2010, luo_tracking_2011}. From skeleton information, gestures were often classified from the 3D joint positions of the human skeleton using the angular configuration of the joints~\cite{cheng_design_2012, lichtenstern_prototyping_2012, nishiyama_human_2013, wang_vision-guided_2013, fujii_gesture_2014, yang_real-time_2015, fareed_gesture_2015, han_human_2017, miller_self-driving_2019, moh_gesture_2019, werner_evaluation_2013, gorer_autonomous_2017, avioz-sarig_robotic_2020, lang_research_2020}. For example, Ghandour et al.~\cite{ghandour_human_2017} trained a neural network classifier to recognise body pose gestures using joint positions from a single image as input. In contrast, dynamic gestures, which consist of a sequence of poses, require multiple image frames for classification. Common algorithms used to track and classify dynamic gestures include Hidden Markov Models (HMM)~\cite{park_real-time_2011, droeschel_towards_2011, burger_two-handed_2012, tao_multilayer_2013, fujii_gesture_2014, xu_online_2014}, particle filters~\cite{park_real-time_2011, burger_two-handed_2012, prediger_robot-supported_2014, du_online_2018}, and Dynamic Time Warping (DTW)~\DIFdelbegin \DIFdel{\mbox{
\cite{qian_visually_2013, canal_gesture_2015, vircikova_teach_2015, pentiuc_drive_2018, chen_wristcam_2019}}\hspace{0pt}
}\DIFdelend \DIFaddbegin \DIFadd{\mbox{
\cite{qian_visually_2013, canal_gesture_2015, vircikova_teach_2015, pentiuc_drive_2018}}\hspace{0pt}
}\DIFaddend . For example, Tao et al.~\cite{tao_multilayer_2013} used a HMM to classify hand waves in various directions, \DIFdelbegin \DIFdel{Chen et al.~\mbox{
\cite{chen_wristcam_2019} }\hspace{0pt}
used DTW to classify dynamic hand gestures from a wrist mounted camera, }\DIFdelend Cicirelli et al.~\cite{cicirelli_kinect-based_2015} used a neural network to recognise gestures from an input of Fourier-transformed joint positions, and Li et al.~\cite{li_real-time_2019} reason about temporal information using an LSTM to determine the intention of the person. While traditional machine learning techniques have been used for gesture and pose classification, such as k-nearest neighbours~\DIFdelbegin \DIFdel{\mbox{
\cite{pereira_human-robot_2013, gao_humanoid_2015, chen_wristcam_2019}}\hspace{0pt}
}\DIFdelend \DIFaddbegin \DIFadd{\mbox{
\cite{pereira_human-robot_2013, gao_humanoid_2015}}\hspace{0pt}
}\DIFaddend , support vector machines~\cite{chen_integrated_2010, chen_approaches_2010, gori_all_2012, ehlers_human-robot_2016}, and multi-layered perceptrons~\cite{martin_estimation_2010, tao_multilayer_2013}, deep neural networks have become increasingly prevalent.
Convolutional neural networks (CNNs) have been used to classify gestures directly from colour images~\cite{arenas_deep_2017, mazhar_real-time_2019, xu_skeleton_2020}, or depth images~\cite{kalidolda_towards_2018, lima_real-time_2019}.
Other architectures have been used to classify gestures from intermediate representations such as skeletal information~\cite{cicirelli_kinect-based_2015, ghandour_human_2017, li_real-time_2019}, and other sensory inputs~\cite{martin_estimation_2010, tao_multilayer_2013}.
Practitioners can fine-tune pre-trained models on task-specific data~\cite{mazhar_real-time_2019}, or train from scratch with a custom dataset~\cite{xu_skeleton_2020, waskito_wheeled_2020}.
As previously indicated, neural networks have been used to classify both static~\cite{ghandour_human_2017} and dynamic~\cite{cicirelli_kinect-based_2015, li_cnn_2019} body poses.

\subsection{Non-Gestural Action Classification}

Human action classification involves the identification of specific types of human motions from video streams. For this section, gestural actions are considered to involve gestures where the person is explicitly trying to communicate with the robot, and non-gestural actions involve actions such as walking, eating, running, and sitting down. Non-gestural actions can be important for robots to recognise and facilitate contextual understanding, but the robot may not require an immediate response from the person, unlike a gesture action. Several action classification methods operated on pre-detected human keypoints~\cite{gorer_autonomous_2017, sorostinean_activity_2018, zhang_automating_2018, lang_research_2020,chalvatzaki_learn_2019, gui_teaching_2018, lee_real-time_2020}. For example, Gorer et al.~\cite{gorer_autonomous_2017} compared the keypoint position of motion identifier joints on an elderly user performing an exercise pose with those of a human demonstrator to identify any disparities. In another example, Vasquez et al.~\cite{vasquez_deep_2017} classify the category and estimate the 3D position and velocity of a person with a walking aid using a ResNet-50~\cite{he_deep_2016} network and a hidden Markov model. Efthymiou et al.~\cite{efthymiou_multi-_2018} showed that dense trajectories~\cite{wang_action_2011} provide better features that result in more accurate action classifications than CNNs, when there is a mismatch between training and testing data (for example, identifying the actions of children from models trained on large action recognition data sets where adults are more prevalent). Lastly, Gui et al~\cite{gui_teaching_2018} trained a Generative Adversarial Network (GAN) from a sequence of skeleton keypoints over time, extracted using OpenPose~\cite{Cao_2017_CVPR}, to generate plausible motion predictions.

\subsection{Social Classification}

This section explores the classification of social factors that were not directly associated with a specific gesture or action, including facial expression, level of engagement, and intent to interact with the robot. For example, Saleh et al.~\cite{saleh_nonverbal_2015} classify head movements, such as nodding and shaking, with an SVM, using the direction and magnitude patterns of depth pixels as features.
Facial expressions are classified by fitting an active appearance model~\cite{masmoudi_expressive_2011}, by using Gabor filters with PCA and an SVM~\cite{ke_vision_2016}, or using a neural network classifier on facial keypoints~\cite{meghdari_real-time_2017}. Gender classification was performed from principle component analysis of face regions \cite{ke_vision_2016}, or using an SVM with local binary patterns and histogram-of-gradient features \cite{chien_navigating_2019}. Intention to interact with a robot was identified using random forest regression on facial expression features~\cite{li_inferring_2019}, or from a combination of the user's line-of-sight to the robot, shoulder orientation, and speech activity \cite{mollaret_multi-modal_2016}. Lastly, level of engagement was estimated using gaze analysis techniques to determine if a person was averting their gaze during conversation with a social robot \cite{bilac_gaze_2017}.

\subsection{Human Motion Tracking}

To interact with a human, robots are often required to track the person through multiple frames, which includes detection, tracking their motion, and re-acquiring the intended person if they have been occluded or fall out-of-frame. Human motion tracking for the purpose of robotic vision has been performed using particle filters~\cite{fahn_real-time_2010, kobayashi_people_2010, ali_improved_2015, prediger_robot-supported_2014, arumbakkam_multi-modal_2010, nair_3d_2011, morato_toward_2014, mollaret_multi-modal_2016}, optical flow~\cite{wu_improved_2019}, or simultaneous localisation and mapping (SLAM)~\cite{angonese_multiple_2017, bellarbi_social_2017, weber_follow_2018}. In specific examples, Fahn et al.~\cite{fahn_real-time_2010} used a particle filter to track a face from the centre of the image using its colour properties, while Nair et al.~\cite{nair_3d_2011} tracked multiple people with a particle filter by first segmenting out the static background and then tracking bounding boxes in the foreground. Image keypoints and features, such as SURF features~\cite{bay_speeded-up_2008}, has be used to find correspondences across frames~\cite{wu_accompanist_2012}, where the track can be initialised by providing a known pattern worn by the target~\cite{chen_person_2012} or automatically identifying features on a detected person's clothing~\cite{wu_accompanist_2012}. Kernalized correlation filters (KCF)~\cite{henriques_high-speed_2015} \DIFdelbegin \DIFdel{has }\DIFdelend \DIFaddbegin \DIFadd{have }\DIFaddend been used for efficient tracking~\cite{luo_real-time_2019, hwang_interactions_2020} and Kalman filters have also been frequently used to reason about and reduce tracking errors from noisy sensors and odometry~\cite{foster_two_2012, morato_toward_2014, gupta_robust_2015, talebpour-board_2016, luo_real-time_2019, long_kinect-based_2018, vasquez_deep_2017}.
For example, Foster et al.~\cite{foster_two_2012} propagated a set of pixel hypotheses for segmented skin-coloured blobs with a Kalman filter. Lastly, human motion prediction~\cite{hu_design_2014, gupta_robust_2015, chen_human-following_2019, morato_toward_2014, landi_prediction_2019}, and robot kinematic models~\cite{wang_vision-guided_2013, zhang_vision-based_2019} have been used to perform better tracking of the person. To demonstrate, Landi et al.~\cite{landi_prediction_2019} used a neural network to predict hand positions to avoid collisions, and inverse kinematics to plan a trajectory that could avoid the human~\cite{moreno_path_2016, zardykhan_collision_2019}. While not expressly used in the corpus of HRI/C papers, there is a significant body of work on techniques for detecting and tracking groups of people, which is likely to be used in future work \mbox{
\cite{lau2010multi, linder2014multi, munaro2014fast, jafari2014real, vazquez2015parallel, vazquez2016maintaining, taylor2016robot, taylor2020robot}}\hspace{0pt}
. For example, Lau et al. \mbox{
\cite{lau2010multi} }\hspace{0pt}
and Linder et al. \mbox{
\cite{linder2014multi} }\hspace{0pt}
cast group detection and tracking as a multi-hypothesis model selection problem, a probabilistic model that allows for the splitting and merging of clusters. RGB-D sensors are frequently used by robots for multi-person tracking \mbox{
\cite{munaro2014fast, jafari2014real, taylor2020robot}}\hspace{0pt}
. The latter \mbox{
\cite{taylor2020robot}}\hspace{0pt}
, predicts social groups from egocentric RGBD by reasoning about joint motion and proximity estimates. These approaches have significant potential for use in HRI/C applications, since reasoning about group behaviour us likely to be critical for robots in social settings.


\DIFaddend \subsection{Multiple Sensors} 
In the selected papers, robotic vision was often paired with other sensors to enhance the robots capacity to perceive and respond to the human. Other sensors often involved microphones for speech recognition~\cite{burger_two-handed_2012, fujii_gesture_2014, vysocky_interaction_2019, yuan_development_2018, du_online_2018, maurtua_natural_2017}, laser range sensors~\cite{luo_human_2010, alvarez-santos_feature_2012, pereira_human-robot_2013, weinrich_appearance-based_2013, hu_design_2014, kobayashi_people_2010} and ultrasonic sensors~\cite{hassan_computationally_2016} to help determine distance, audio sensors for locating the active speaker~\cite{cech_active-speaker_2015, nguyen_audio-visual_2014}) and to relocate people who were out-of-view~\cite{luo_human_2010, nguyen_audio-visual_2014, bayram_audio-visual_2016}, Leap Motion sensors~\cite{kalidolda_towards_2018} and inertial measurement units to help track movement and orientation~\cite{du_online_2018, haghighi_integration_2019, zhang_human_2020}\DIFdelbegin \DIFdel{, as well as wrist cameras to assist in hand gesture detection~\mbox{
\cite{chen_wristcam_2019}}\hspace{0pt}
}\DIFdelend . Force sensors were used with vision sensors during applied tasks such as object handover~\cite{bdiwi_handing-over_2013}, collaborative manipulation~\cite{cherubini_unified_2015, koustoumpardis_human_2016}, and to determine contact in a safe workspace scenario~\cite{bingol_practical_2020}. 
Tactile sensors were also used in object handover to assist physical interaction with the environment~\cite{ikai_robot_2016}. Humans who operated the robot (\DIFdelbegin \DIFdel{$N=139$}\DIFdelend \DIFaddbegin \DIFadd{$N=138$}\DIFaddend ) were often provided additional information from other sensors or information sources. Operators made informed decisions by information provided from sources such as different LED colours~\cite{couture-beil_selecting_2010, milligan_selecting_2011, lichtenstern_prototyping_2012, alonso-mora_human_2014, gao_humanoid_2015} feedback on a display screen~\cite{fujii_gesture_2014, bolano_towards_2018, lavanya_gesture_2018, chen_human-following_2019} video feedback~\cite{lalejini_evaluation_2015, sriram_mobile_2019}, augmented reality~\cite{lambrecht_spatial_2012}, haptic feedback~\cite{scheggi_human-robot_2014}, spoken response from the robot~\cite{fujii_gesture_2014, gao_humanoid_2015, zhang_optimal_2016, mazhar_real-time_2019}, or robot movement~\cite{zhang_optimal_2016, moh_gesture_2019} to signal or confirm that the command had been received by the robot. In most examples, the position or configuration of the robot was sufficient (\DIFdelbegin \DIFdel{$N=97$}\DIFdelend \DIFaddbegin \DIFadd{$N=96$}\DIFaddend , e.g. ~\cite{arenas_deep_2017, augello_towards_2020}). 

\section{RQ5. What are the datasets and models that have been used for robotic vision in human-robot interaction and collaboration?}\label{sec:rq5}

Common datasets and model choices are summarised to identify open source choices for robotic vision methods and to provide recommendations for future use (See Table 2). Custom datasets were often collected by researchers using the platform under investigation. Some examples include~\DIFdelbegin \DIFdel{\mbox{
\cite{martin_estimation_2010, droeschel_towards_2011,park_real-time_2011, pereira_human-robot_2013, tao_multilayer_2013, cicirelli_kinect-based_2015, arenas_deep_2017, ghandour_human_2017, lima_real-time_2019, chen_wristcam_2019, xu_skeleton_2020, meghdari_real-time_2017, chien_navigating_2019}}\hspace{0pt}
}\DIFdelend \DIFaddbegin \DIFadd{\mbox{
\cite{martin_estimation_2010, droeschel_towards_2011,park_real-time_2011, pereira_human-robot_2013, tao_multilayer_2013, cicirelli_kinect-based_2015, arenas_deep_2017, ghandour_human_2017, lima_real-time_2019, xu_skeleton_2020, meghdari_real-time_2017, chien_navigating_2019}}\hspace{0pt}
}\DIFaddend . Some papers built on previous work but details of the data set used for training were not provided~\cite{couture-beil_selecting_2010, quintero_visual_2015, ehlers_human-robot_2016, maurtua_natural_2017, martin_real-time_2019, li_real-time_2019, saleh_nonverbal_2015}. Only three papers (1.5\%) released their datasets~\cite{mazhar_real-time_2019, munaro_fast_2014, lima_real-time_2019}: Mazhar et al~\cite{mazhar_real-time_2019} released OpenSign, a dataset of hand gestures from 10 volunteers who recorded 10 static gestures and Lima et al~\cite{lima_real-time_2019} released a dataset of 160,000 samples from men and women in different poses who performed open and closed hand gestures. In action classification, there was a need to use large dataset collections of different actions~\cite{wang_collision-free_2017, gui_teaching_2018, sorostinean_activity_2018, vasquez_deep_2017, lee_real-time_2020}. For instance, the NTU RGB-D dataset~\cite{shahroudy_ntu_2016} contained 60 action classes from 56,880 video samples with RGB videos, depth map sequences, 3D skeletal data, and infrared videos for each sample. This dataset was used by~\cite{sorostinean_activity_2018, lee_real-time_2020}. The manipulation action dataset (MAD)~\cite{fermuller_prediction_2018} contained five different objects with five distinct actions where each action was repeated five times for a total number of 625 recordings. It was used by~\cite{wang_collision-free_2017}. The H3.6M dataset~\cite{ionescu_dataset_2014} had 3.6 million 3D human poses (5 female, 6 male from 4 view points) and was used by~\cite{gui_teaching_2018} to train a motion prediction GAN. Vasquez et al~\cite{vasquez_deep_2017} made available their dataset of various mobility levels (wheelchair, walker, walking stick) with over 17,000 annotated RGB-D images. People detection datasets included the INRIA and the Market-1501 datasets. Lee et al~\cite{lee_visual_2020} trained the twin convolutional neural network using the Market-1501~\cite{zheng_scalable_2015} dataset, containing 32,000 annotated bounding boxes and 500,000 distractor images. Munaro et al~\cite{munaro_fast_2014} released the Kinect Tracking Precision (KTP) dataset, collected from the Microsoft Kinect and containing 8,475 frames with a total of 14,766 instances of people. Weber et al~\cite{weber_follow_2018} trained a face detection model on the HollywoodHeads Dataset~\cite{vu15heads}, containing annotated head bounding box regions for 224,740 video frames from Hollywood movies. The Cohn-Kanade dataset~\cite{kanade_comprehensive_2000} and its extended form (CK+)~\cite{lucey_extended_2010} included 123 subjects and 593 image sequences and was used to train facial expression models for the following papers~\cite{sosnowski_mirror_2010, ke_vision_2016, li_cnn_2019}. The AffectNet Database~\cite{mollahosseini_affectnet_2019} contained 1,000,000 facial images from Internet queries and was used by~\cite{li_cnn_2019}. The WIDER FACE dataset~\cite{yang2016wider} contained 32,203 images, with 393,703 bounding regions of faces and used by~\cite{li_inferring_2019}. The Aberdeen Facial Database~\cite{noauthor_aberdeen_nodate} (687 faces), GUFD Facial Database~\cite{noauthor_gufd_nodate} (6,000 images of faces), Utrecht ECVP Facial Database~\cite{noauthor_utrecht_nodate} were used, along with images of laboratory personnel to train a face classifier~\cite{chien_navigating_2019}. Other datasets included the FERET database \cite{phillips1998feret} for gender identification as used by \cite{ke_vision_2016}, Inter-ACT corpus \cite{castellano2010inter} as used by \cite{castellano_context-sensitive_2014}, and the Cohn-Kanade Facial Experession Database \cite{kanade2000comprehensive} and MMI Face Database \cite{pantic2005web} as both used by \cite{sosnowski_mirror_2010}. 

\begin{table}[!ht]\centering
\caption{Datasets used in Robotic Vision for Human-Robot Interaction/Collaboration}\label{tab:datasets_table}
\scriptsize
\begin{tabular}{p{0.25\columnwidth} | p{0.17\columnwidth} | p{0.2\columnwidth} | p{0.15\columnwidth} |p{0.08\columnwidth}}\toprule

\textbf{Name of dataset} &\textbf{Type of data} &\textbf{Volume of data} &\textbf{Usage} &\textbf{Used by} \\\midrule

NTU RGB-D dataset~\cite{shahroudy_ntu_2016} & RGB-D images and 3D skeletal data & 60 action classes, 56,880 video samples & Action recognition &\cite{sorostinean_activity_2018, lee_real-time_2020}\\

Manipulation action dataset (MAD)~\cite{fermuller_prediction_2018} & Video of object / action samples (e.g. cup | drink, pound, shake, move) & 625 recordings & Action recognition  &\cite{wang_collision-free_2017}\\

H3.6M dataset~\cite{ionescu_dataset_2014} & RGB-D images with bounding regions, 3D pose data & 3.6 million 3D human poses & Human motion prediction &\cite{gui_teaching_2018}\\

Market-1501 dataset~\cite{zheng_scalable_2015}& RGB images & 32,000 annotated bounding boxes & Person detection &\cite{lee_visual_2020}\\

INRIA dataset~\cite{dalal_histograms_2005} & RGB images & 1805 images of humans & Person detection &\cite{alvarez-santos_feature_2012, munaro_fast_2014}\\

HollywoodHeads dataset~\cite{vu15heads} & RGB video frames & Annotated head regions for 224,740 video frames & Head detection &\cite{weber_follow_2018} \\

The extended Cohn-Kanade (CK+) dataset~\cite{kanade_comprehensive_2000,lucey_extended_2010} & RGB images & 593 image sequences & Facial expression recognition &\cite{sosnowski_mirror_2010, ke_vision_2016, li_cnn_2019} \\

The AffectNet Database~\cite{mollahosseini_affectnet_2019} & RGB images & 1,000,000 facial images & Facial expression recognition &\cite{li_cnn_2019}\\

The WIDER FACE dataset~\cite{yang2016wider} & RGB images & 32,203 images, with 393,703 bounding regions & Facial expression recognition &\cite{li_inferring_2019}\\

The Aberdeen Facial Database~\cite{noauthor_aberdeen_nodate} & RGB images & 687 faces & Face detection &\cite{chien_navigating_2019}\\

Glasgow Unfamiliar Face Database (GUFD)~\cite{noauthor_gufd_nodate} & RGB images & 6,000 images of faces & Face detection &\cite{chien_navigating_2019}\\

Utrecht ECVP Facial Database~\cite{noauthor_utrecht_nodate} & RGB images & 131 images & Face detection &\cite{chien_navigating_2019}\\

ChaLearn Looking at People Challenge~\cite{agapito_chalearn_2015} & RGB-D images & 14,000 images of hand gestures & Hand gestures & \cite{bras_gesture_2018} \\


Kinect Tracking Precision (KTP) dataset ~\cite{munaro_fast_2014} & RGB-D images & 8,475 frames, 14,766 instances of people & Person detection &\cite{munaro_fast_2014}\\

Annotated hospital dataset~\cite{vasquez_deep_2017} & RGB-D images & 17,000 annotated images & Mobility detection &\cite{vasquez_deep_2017}\\

OpenSign~\cite{mazhar_real-time_2019} & RGB-D images & 20,950 images & Hand gestures &\cite{mazhar_real-time_2019} \\

Dataset by Lima et al~\cite{lima_real-time_2019}& RGB images & 160,000 images of open and closed hands & Gesture recognition &\cite{lima_real-time_2019} \\

\bottomrule
\end{tabular}
\end{table}

\section{RQ6. What has been the main participant sample, and how is robotic vision in human-robot collaboration and interaction evaluated?} \label{sec:rq6}

\subsection{Main Participant Sample}
\begin{figure}[h]
\centering
\subfloat[Team Metric Categories]{\includegraphics[width=0.45\columnwidth]{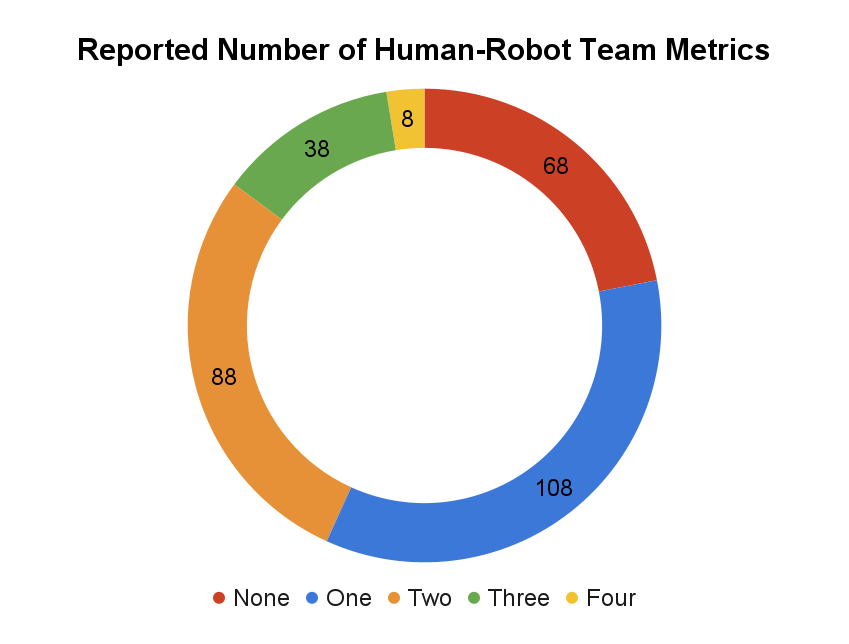}\label{team:hr}}
\subfloat[Team Metric Report]{\includegraphics[width=0.45\columnwidth]{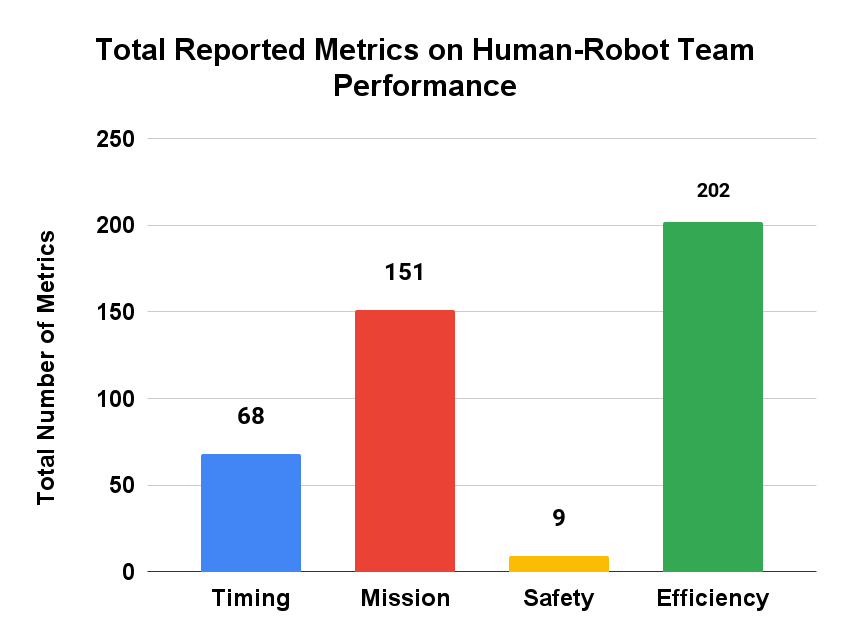}\label{team:general}}
\hfil
\subfloat[Application Area]{\includegraphics[width=0.45\columnwidth]{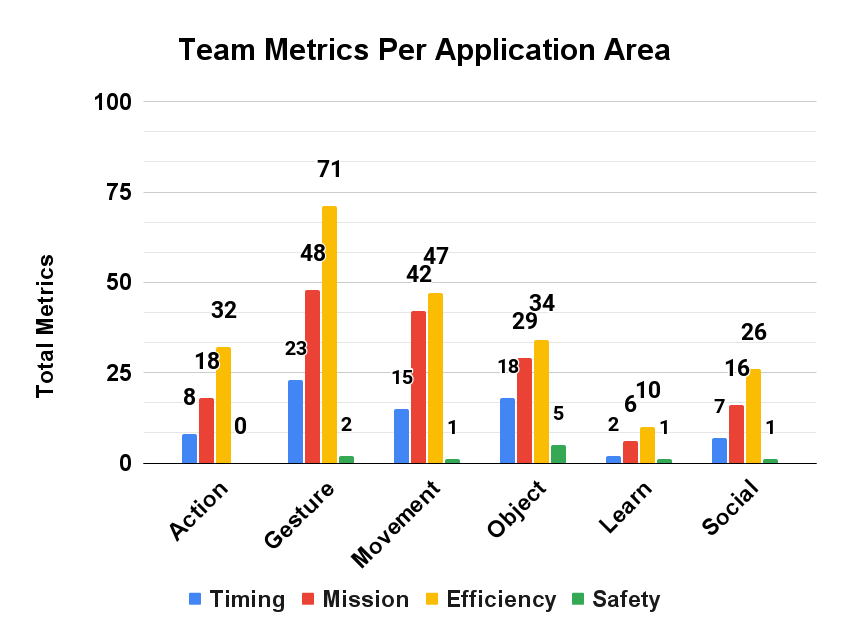}\label{team:app}}
\hfil
\subfloat[Domain]{\includegraphics[width=0.45\columnwidth]{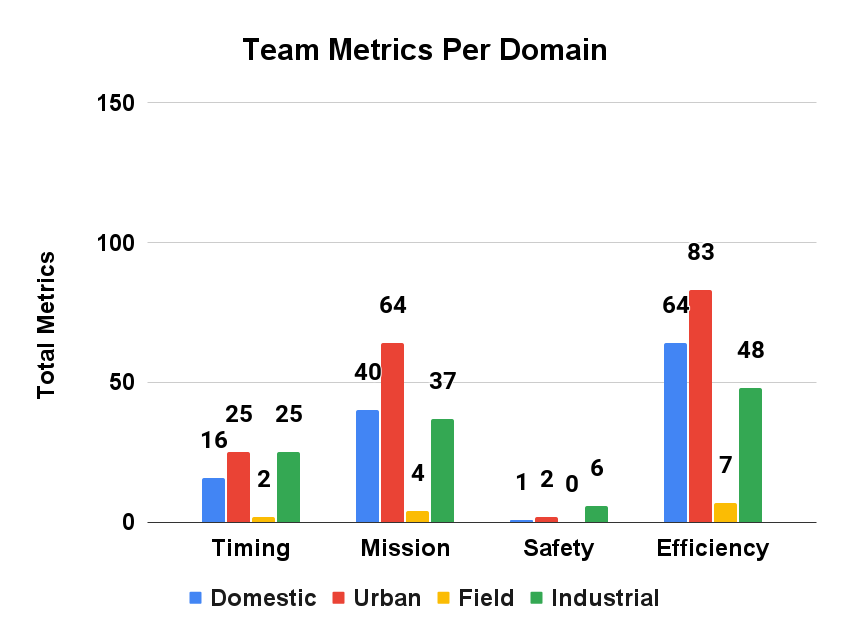}\label{team:domain}}
\hfil
\subfloat[Robot Type]{\includegraphics[width=0.45\columnwidth]{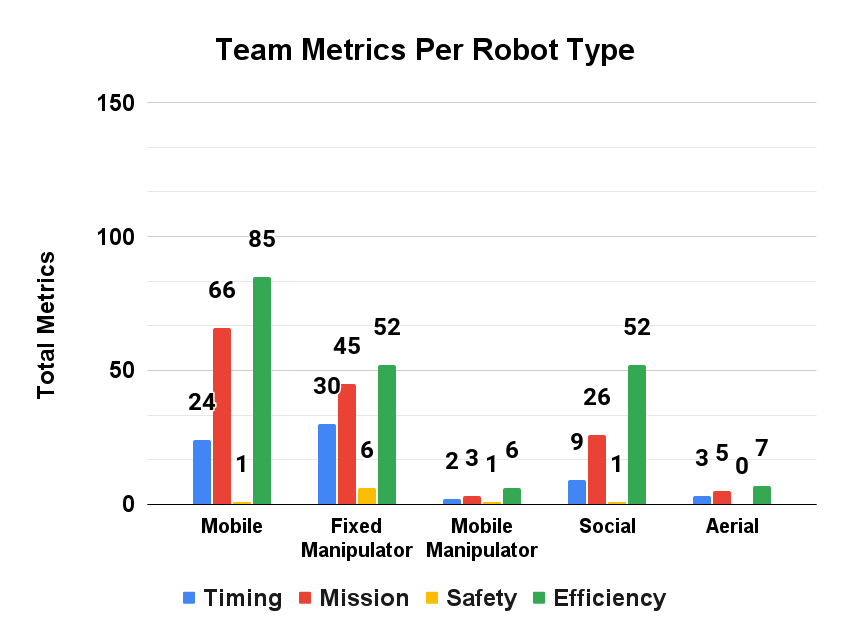}\label{team:robot}}
\hfil
\subfloat[Camera Type]{\includegraphics[width=0.45\columnwidth]{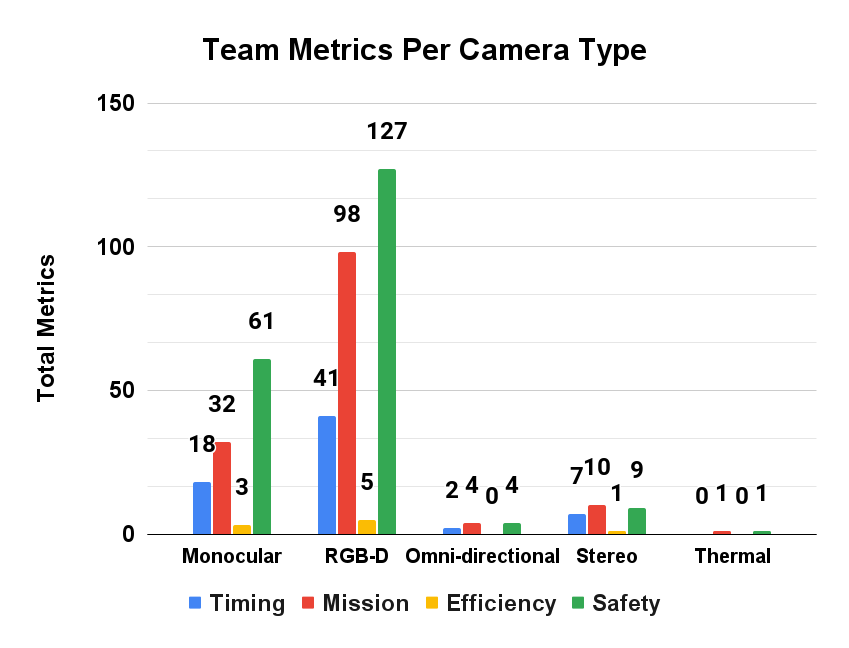}\label{team:camera}}
\hfil
\caption{Trends in Robotic Vision Evaluation in Human-Robot Interaction and Collaboration}
\label{fig:metrics}
\end{figure}

Many published works reported little human-relevant information. In the \DIFdelbegin \DIFdel{303 }\DIFdelend \DIFaddbegin \DIFadd{310 }\DIFaddend papers, only \DIFdelbegin \DIFdel{58 (19}\DIFdelend \DIFaddbegin \DIFadd{66 (21}\DIFaddend \%) reported details around a human experiment or testing with people. In the studies that did report participant numbers, there was a calculated total of \DIFdelbegin \DIFdel{1123 }\DIFdelend \DIFaddbegin \DIFadd{1228 }\DIFaddend participants across all papers ($M = 20$, Range = 1 - 150, $SD = 22$; \cite{christiernin_interacting_2016} did not report numbers). In studies that reported participant age (22\%), participants were on average 32 years old (Range = \DIFdelbegin \DIFdel{6 }\DIFdelend \DIFaddbegin \DIFadd{1 }\DIFaddend - 88, $SD = 24$). For papers that reported gender (\DIFdelbegin \DIFdel{38}\DIFdelend \DIFaddbegin \DIFadd{35}\DIFaddend \%), there was an average percent split of \DIFdelbegin \DIFdel{13 male and 8 }\DIFdelend \DIFaddbegin \DIFadd{70\% male and 30\% }\DIFaddend female participants. No studies reported participant country of origin. Gesture recognition had the highest number of participants at 302 (\DIFdelbegin \DIFdel{19.5}\DIFdelend \DIFaddbegin \DIFadd{20}\DIFaddend \%, 23 out of \DIFdelbegin \DIFdel{118}\DIFdelend \DIFaddbegin \DIFadd{116}\DIFaddend ) papers. Few experiments had direct evaluation for robotic vision performance. Instead, experiments often had a clear focus on robot evaluation as part of its overall intended task or role in the interaction. Evaluation metrics were more likely to involve objective metrics (\DIFdelbegin \DIFdel{$N=337$}\DIFdelend \DIFaddbegin \DIFadd{$N=384$}\DIFaddend ) compared to subjective metrics (\DIFdelbegin \DIFdel{$N=39$}\DIFdelend \DIFaddbegin \DIFadd{$N=40$}\DIFaddend ), with the nature of the metrics around robot components such as overall perception, robot task performance, or robot preference when compared to other modalities or system setups. Common quantitative questionnaires included the NASA-TLX, e.g.~\cite{valle_personalized_2019,kogkas_free-view_2019,jevtic_comparison_2015}, and GODSPEED, e.g. ~\cite{foster_two_2012, werner_evaluation_2013, zhang_automating_2018}. Others included the PARADISE framework~\cite{foster_two_2012}, Positive or Negative Affect Scale \cite{werner_evaluation_2013}, Robot Acceptance Scale \cite{werner_evaluation_2013}, or a custom-made scale, such as a 9-point evaluation~\cite{prediger_robot-supported_2014}, 7-point comfort rating~\cite{scheggi_human-robot_2014}, or 5-point robot performance rating~\cite{lee_learning_2017}. Figure~\ref{fig:metrics} depicts the total number of team metric categories, number of reported human-robot team metrics, team metrics for each application area, domain, robot type and camera type. A selection of exemplar use cases will be provided in the next section (Section~\ref{sec:rq7}) to describe state-of-the-art robotic vision in human-robot interaction and collaboration.

\subsection{Evaluation Metrics in Each Application Area}
\subsubsection{Gesture Recognition} 
Robot evaluation scores were often on preference ratings, how enjoyable people found the interaction, and how favourable people found the robot. For instance, the robot was found to be engaging during a social interaction~\cite{csapo_multimodal_2012}, the interaction was enjoyable in a navigation task~\cite{lalejini_evaluation_2015}, and the robot was the preferred choice when compared to other control methods such as joystick or gamepad~\cite{yoshida_evaluation_2011}, or when compared to other modalities such as a screen in a public service centre~\cite{kalidolda_towards_2018}. People reported high rating or preference for a robotic vision component involved in the interaction, such as for gestures were natural and easy to use (79\%, $n=24$, \cite{canal_gesture_2015}), for specific gesture styles (62\% preference for elbow to finger, and 38\% for eye to finger, \cite{abidi_human_2013}). This was not systematic, with other modalities not related to robotic vision also rated more favourably, such as physical interaction rated as the least demanding and most accurate method to guide a mobile robot to different waypoints ($n=24$, \cite{jevtic_comparison_2015}), or handheld devices being easier to use than gestures ($n=23$, \cite{lalejini_evaluation_2015}). Other robot evaluations included learning speed for use of the robot in a gesture controlled grasping task ($n=10$, \cite{lima_real-time_2019}), number of errors in terms of distance from the robot when signaling to pick up items ($n=16$, \cite{droeschel_towards_2011}) or accuracy as seen during a gesture game with a robot (15\% lost due to out of distribution gestures and 10\% due to classification error, ($n=30$, \cite{gori_all_2012}). 

\subsubsection{Action Recognition} Robot evaluations often involved preference scores, willingness to use, and satisfaction levels with the robot. People rated their impression of the robot's behaviour compared to several baselines on a 1-5 scale ($n=12$, ~\cite{lee_learning_2017}). Trust was assessed in a home service robot for when it could detect the persons' actions with 75\% of people ($n=16$) reporting that the feature was important~\cite{sorostinean_activity_2018}, and more than 40 out of 50 people (exact number not reported) reported a satisfaction level of at least 4 out of 5~\cite{lang_research_2020}. In preference scores, 77\% ($n=30$) preferred a robot for exercise compared to training videos~\cite{werner_evaluation_2013} and 92\% ($n=32$) reported willingness to continue the robot program~\cite{avioz-sarig_robotic_2020}. Other evaluations included 65\% (13 out of 20) who were unable to tell if the robot was autonomously operated or teleoperated from its behaviour~\cite{zhang_automating_2018}. Lastly, a robot that used activity recognition for physical activity had people report an increase in exercise success rates with 12 elderly users over a 3 week period~\cite{gorer_autonomous_2017}. 

\subsubsection{Robot Movement in Human Spaces}
Evaluations were often in performance, preference and acceptability. In one example, a mobile robot was successfully controlled in 77.4\% of the interactions with 8.9\% of unsuccessful interactions attributed to fast rotational robot movement or large distance between the camera and person~\cite{ehlers_human-robot_2016}. Considering preference, a human following task revealed that people in general found the robot's behaviour appropriate, but most reported being uneasy with it ($n=13$, \cite{prediger_robot-supported_2014}), and 46\% (6 out of 13) were willing to adapt their living environment to accommodate a mobile robot, but 10 did not want to adjust their walking speed~\cite{prediger_robot-supported_2014}. 

\subsubsection{Object Handover and Collaborative Actions}
Robot evaluations involved both performance and preference rubrics. For performance, twelve people used an UR10 for a user controlled pick and place task and after three trials, those with no experience achieved greater than 75\% success with a interaction time of 69 seconds on average~\cite{bolano_towards_2018}. In addition, higher mental load was found for 7 novice and experienced operators in a shared work space when the robot operated closer to the person with a higher velocity~\cite{tan_safety_2010}. For preference, four people (100\% of the sample) reported that a robot arm did not always match their expectations~\cite{christiernin_interacting_2016}, collaborative actions from robots in surgical assistance could help them perform their role more efficiently ($n=16, $\cite{kogkas_free-view_2019}) and reduced workload for a shoe fitting task when the robot was personalised with preferences, including shorter time and fewer commands~\cite{valle_personalized_2019}. Lastly, some found no differences in time or error rates between humans for passing  instruments to a person~\cite{kogkas_free-view_2019}.

\subsubsection{Social Communication}
There were performance and preference scores for social communication. For performance, 16 people had high detection accuracy for interest to engage (99\%) as well as if the person was not interested (92\%, ~\cite{saleh_nonverbal_2015}). A robot served drinks successfully with high accuracy (100\% for single person scenario) with a good response (658ms) and expected interaction time (49.4s, \cite{foster_two_2012}). Gesture recognition and speed in a human-robot gesture game had good recognition accuracy (92\%, $n = 5$, ~\cite{yuan_natural_2020}). Other performance results were for a PR2 robot that achieved a 100\% success rate to answer user requests (72.7\% first attempt, 18.2\% second, remainder on the third, ~\cite{mollaret_multi-modal_2016}). Some behaviours were also improved with robotic vision, such as a humanoid receptionist robot conversed with 26 people to compare engagement aware behaviour, which included small improvements in eye gaze toward the robot (78.8\% with and 73.7\% without, ~\cite{li_visual_2015}). people perceived the robot as more intelligent, and were more satisfied with the interaction, though no effect was noticed on task performance~\cite{li_visual_2015}. A robot bartender was rated as likable, intelligent, and safe by 31 people~\cite{foster_two_2012}. Others found larger increases with robotic vision, such as gaze and pause detection to determine when a person had finished speaking resulted in a two times increase in talking time compared with filled pause detection (\DIFdelbegin \DIFdel{$N = 28$}\DIFdelend \DIFaddbegin \DIFadd{$n = 28$, ~\cite{bilac_gaze_2017}}). Some also had reported engagement with the robot but no statistics~\cite{taheri_social_2014}. 

\section{RQ7. What is the state-of-the-art in vision algorithm performance for robotic vision in human-robot collaboration and interaction?}\label{sec:rq7}

Considering robot evaluation and vision algorithm performance, state-of-the-art performance is paramount to the functional benefit of the robot, including what tasks or services the robot could provide to the human. However, Section~\ref{sec:rq6} demonstrated that few papers reported standardized metrics that directly evaluated robotic vision performance. This makes it challenging to fairly compare the performance of the vision algorithms used in these works. It is nonetheless important to highlight works that make superior use of robotic vision in HRI/C systems. Therefore, we present selected works that well-represent the use of robotic vision in human-robot collaboration and interaction with respect to the criteria of novelty, impact, and/or robustness. Exemplar studies are identified that showcased creative and/or robust use cases of robotic vision, given that systematic differences could not be calculated across studies from metric and result reporting. These examples help to direct to future pathways in the field to increase experimental rigor with more experimentation and more systematically evaluate the feasibility for the capacity, speed and accuracy for robotic vision to be used with people. 

\subsection{Gesture Recognition}
Mazhar et al.~\cite{mazhar_real-time_2019} demonstrated control of a KUKA arm via hand gestures by fine-tuning an Inception V3 convolutional neural network on a custom dataset (OpenSign). This resulted in a system that was able to detect 5 gestures in a row at 40 Hz (250ms per detection). OpenPose was used to localize hands in the dataset images and the Kinect V2 depth map was used to segment the hands from the background, allowing background substitution for data augmentation. Inception V3 fine-tuning resulted in a validation accuracy of 99.1\% and a test accuracy of 98.9\%. The dataset had RGB-D images with 10 gestures performed by 10 people, including 8646 original images and 12304 synthetic images from background substitution. Waskito et al.~\cite{waskito_wheeled_2020} tested the robustness of their hand gesture classifier as the hand was rotated or when lighting conditions were varied to find an average total accuracy of 96.7\% with each gesture having a 0.141s average response time. Lastly, Pentiuc~\cite{pentiuc_drive_2018} used skeleton data from the Kinect and the dynamic time warping algorithm to detect 5 gestures with an accuracy of >86\%. While these methods cannot be compared directly, since different gestures and settings were being evaluated, the overall trend was to use higher capacity models trained with more data to increase the accuracy and robustness of gesture recognition.

\subsection{Human Detection and Tracking}
To detect and track a specific person, Hwang et al.~\cite{hwang_interactions_2020} integrated a Single-Shot Detector, FaceNet, and a Kernelized Correlation Filter. With this system, Hwang et al.~\cite{hwang_interactions_2020} were able to detect humans up to 8m away and recognize specific faces, achieving a maximum position error of 4cm and 5$^\circ$ orientation. For tracking a person from a mobile robot, Weber et al.~\cite{weber_follow_2018} achieved a 59.7\% mean Average Precision (mAP) with a Single-Shot Detector and tracking-as-repeated-detection strategy. Zhang et al.~\cite{zhang_vision-based_2019} compared their detection and tracking method using target contour bands to several others that used videos from the object tracking benchmark (OTB) dataset. This work showed the presented method was more accurate (94\%) with the fastest processing time (34fps). Once deployed on a mobile robot, the robot could follow an identified target for around 648m. Fang~\cite{fang_vehicle-mounted_2019} found that dynamic body poses could be recognized with high accuracy (>96\% classification accuracy on 300 tests) but limited details were provided on the method~\cite{fang_vehicle-mounted_2019}.
To detect and distinguish people based on walking aids, Vasquez et al.~\cite{vasquez_deep_2017} found that combining a Kalman filter, a hidden Markov model, and a Fast R-CNN region extractor improved system performance by a factor of 7 compared to a dense sliding window method. 

\subsection{Non-Gestural Action Recognition} 
In an action recognition task, Lee et al.~\cite{lee_real-time_2020} achieved an accuracy of 71\% from an RGB camera on the NTU RGB-D dataset (75\% from Kinect) at 15fps. For recognising actions from a child, Efthymiou~\cite{efthymiou_multi-_2018} used dense trajectories from multi-view fusion as the input to their action recognition system and evaluated on a test set of 25 children performing 12 actions with comparisons to a test set of 14 adults~\cite{efthymiou_multi-_2018}. Finally, to adapt the motion of a robotic assistant rollator to the patients, Chalvatzaki et al.~\cite{chalvatzaki_learn_2019} found that their model-based reinforcement learning method that uses predicted human actions obtained a smaller tracking error than several other control methods~\cite{chalvatzaki_learn_2019}. 

\section{RQ8. What are the upcoming challenges for robotic vision in human-robot collaboration and interaction?}\label{sec:rq8}
This section will discuss potential and known challenges for robotic vision in human-robot collaboration and interaction, as well as a brief discussion of general robotic vision challenges. Overall summaries on future human-related challenges and general robotic vision challenges demonstrate target areas for consideration in the design, deployment and future use of robotic vision in human spaces. Challenges specific to vision during human-robot collaboration include the ethical use of human data, human model selection and optimisation, experimental design and validation and appropriate trust. 

\subsection{General Challenges of Robotic Vision}
\DIFaddbegin \DIFadd{In addition to challenges specific to human-robot interaction, there are also more general challenges related to robotic and computer vision. While the performance of robotic vision systems are often bounded by what can be achieved by state-of-the-art computer vision algorithms, there are many reasons why state-of-the-art computer vision techniques have not been transferred to robotic platforms.  
}\DIFaddend 

\DIFaddbegin \DIFadd{An algorithm that uses visual data may not be sufficiently robust to perform in real-world conditions and edge cases relevant for HRI/C~\mbox{
\cite{sunderhauf2018limits, corke2011robotics}}\hspace{0pt}
. While visual data may contribute to better multi-modal understandings of human states, actions and involvement with the robot~\mbox{
\cite{10.1007/s11042-020-09004-3}}\hspace{0pt}
, vision still presents with challenges. For the robot to understand its given task or action, robots may require large training data for new tasks and/or need access to processing capabilities that are unavailable on the robotic platform (such as multiple GPUs). Hardware performance can be limited by robot on-board processing compared to cloud processing and the availability, performance and cost of hardware solutions, such as the RGB-D camera as an inexpensive source of depth information compared with expensive alternatives like laser range sensors \mbox{
\cite{susperregi2013rgb}}\hspace{0pt}
. Computationally intensive algorithms can lead to the requirement to have a graphics processing unit in the robot, which can be heavy, noisy and require a lot of power \mbox{
\cite{pena2017benchmarking}}\hspace{0pt}
.
}

\DIFadd{Sim-to-real transfer can be a particular challenge for many learning-based robotics applications, and progress is often relinquished to large companies who can implement large-scale data collection and testing \mbox{
\cite{doi:10.1177/0278364917710318}}\hspace{0pt}
. In addition, deep neural networks often fail to generalize, with a reduction in accuracy when tested outside of benchmark data sets \mbox{
\cite{recht2019imagenet}}\hspace{0pt}
; a method that achieves state-of-the-art performance on a benchmark dataset may not generalize immediately to a real-world setting on a robotic platform. Transferring cutting-edge computer vision techniques may also be too recent to have been adopted in physically embodied robots, let alone applied to scenarios related to HRI/C. In the scenarios when techniques have been transferred into HRI/C applications, robots can encounter failures due to computational delays or challenges around the complexity of the vision-based activities related to the task, such as to perceive and understand the diversity of hand-based gestures from multiple people across different countries~\mbox{
\cite{suma2019computer}}\hspace{0pt}
. Real world vision-based challenges can also occur with robots operating around people, such as important information being occluded when perceiving the person, and critical information not being identified or perceived during the interaction. However, robots can better overcome some of these challenges, such as by controlling camera positioning, adjusting to capture missing information, and orienting visual capture to help fill in the missing gaps~\mbox{
\cite{sunderhauf2018limits}}\hspace{0pt}
. Additional challenges include software challenges such as the availability of open source libraries, software development kits, and the lack of training data for specific use cases.
}

\DIFadd{Robots must also be able to act upon the visual information in a relevant and suitable way. For instance, there continue to be challenges translating signals from computer vision into actionable and useful robot functions for robots, such as movement and manipulation actions that can improve the robot's utility for a given task~\mbox{
\cite{macdonald2019active}}\hspace{0pt}
. This could be impacted by the limited use of participatory design to select suitable applications for robots to assist people~\mbox{
\cite{muller2007participatory}}\hspace{0pt}
, or the inability of current vision systems to address tasks and actions that people want robots to assist them on \mbox{
\cite{yan2014survey}}\hspace{0pt}
. Either of these could have contributed to slowing down the deployment of robotic vision use cases in human domains.} However, human detection and tracking is clearly a key capability for many HRI/C systems, and therefore likely that the current state-of-the-art in computer vision will be rapidly transferred to these systems in the near future.

\subsection{\DIFadd{Fair and Ethical Use of Human Data}} 
\DIFaddend Firstly, there are important challenges around the fair and ethical use of human data \DIFaddbegin \DIFadd{for the purpose of HRI/C when interacting and collaborating with robots }\DIFaddend ~\cite{clark2019advancing, mehrabi2021survey}. \DIFdelbegin \DIFdel{Challenges around the use of human data use include the capacity for people to give informed consent in the context of human-robot interaction that involves their data. }\DIFdelend \DIFaddbegin \DIFadd{For example, the General Data Protection Regulation (GDPR) describes regulations on the processing of personal data in Europe, including the consent of individuals for use of their personal data~\mbox{
\cite{EUdataregulations2018}}\hspace{0pt}
. Fair and ethical use will therefore need to use data processing and management methods that comply with national and/or international data protection regulation and laws. 
}\DIFaddend For the next decade, \DIFdelbegin \DIFdel{there is a high likelihood that while robots in human spacesare still considered an emergent technology, there will }\DIFdelend \DIFaddbegin \DIFadd{robots are likely to continue to enter human spaces, and there may }\DIFaddend be limited public knowledge and awareness on how \DIFdelbegin \DIFdel{robot interactions will use human-based information to }\DIFdelend \DIFaddbegin \DIFadd{interacting with a robot may use their personal information to help }\DIFaddend facilitate the interaction. \DIFdelbegin \DIFdel{Consent should therefore }\DIFdelend \DIFaddbegin \DIFadd{Examples include the robot using its camera system to perceive and classify the person's facial features, body pose and actions, as well as using visual information to make inferences on ways to engage the person, such as by classifying the person's age range, intent to interact with the robot and future actions. Similar to other data-intensive fields, there are important implications on the use of human data in HRI/C, such as the capacity for people to give informed consent and for the appropriate collection, management and storage of data in the context of human-robot interaction. Consent to use data should }\DIFaddend be obtained with clear explanation \DIFaddbegin \DIFadd{or capacity to access detailed information }\DIFaddend on what data is being collected, how it will be used, and how long it will be stored \DIFdelbegin \DIFdel{. This includes }\DIFdelend if data will be used for personal, private, or third-party use. For instance, previously recorded data (images and videos) \DIFdelbegin \DIFdel{can be }\DIFdelend \DIFaddbegin \DIFadd{could be captured and approved to be }\DIFaddend used to improve future robotic interactions as commonly used in computer vision by fine-tuning pre-trained neural networks, e.g.~\cite{bilen2016dynamic}. Other \DIFdelbegin \DIFdel{considerations include data processing and management processes that comply with national and}\DIFdelend \DIFaddbegin \DIFadd{future challenges around fair and ethical use will include data storage and}\DIFaddend /or \DIFdelbegin \DIFdel{international data protection regulation and laws}\DIFdelend \DIFaddbegin \DIFadd{ownership of any images and video collected from robots in human-spaces, including the right for people to access, edit or request deletion of any or all images or video streams collected from them}\DIFaddend . Robot interactions with a physical hardware system does not always include screens or terminals, and these interactions can be located in high-volume areas with frequent turnover of people, such as in a public space. These robot interactions also often do not facilitate similar user agreements or consent notices as other digital methods such as website or smartphone application use~\cite{doi:10.1177/2050157919843961}. Future challenges therefore should involve methods to address clear and transparent notices of intent to use human data when images or video captured for the purpose of core vision-based features, such as to follow the correct person or identification of a specific customer to complete an order transaction. This could include consent as provided by active or passive consent, and/or accessible information about the robot deployed in the public space with the potential for people to avoid or remove themselves from the robots field of view. This could also include detailed consideration for processes to obtain appropriate informed consent for certain groups which can have a second person involved in the consent process, such as guardians of young children or those who are unable to consent for themselves. 

Other challenges also involve the concept of privacy and helping to mediate negative effects around invasion of privacy from robot use~\cite{doi:10.1177/2050157919843961}. For instance, weighing up potential benefits with risks for each deployment to ensure that visual information is collected only when required for functionality and if so, it is handled and stored with proper care. 
\DIFdelbegin \DIFdel{Other future challenges around fair and ethical use will include data storage and/or ownership of any images and video collected from robots in human-spaces, including the right for people to access, edit or request deletion of any or all images or video streams collected from them}\DIFdelend \DIFaddbegin \DIFadd{Furthermore, there have been advances made in the domain of privacy-preserving computer vision, such as to anonymize faces during action detection~\mbox{
\cite{Ren_2018_ECCV}}\hspace{0pt}
, as well as privacy-preserving visual SLAM~\mbox{
\cite{10.1007/978-3-030-58542-6_7}}\hspace{0pt}
, given that point clouds can retain a sufficient level of information to re-create the surrounding environment, potentially compromising privacy if people were intentionally or unintentionally involved in the scene~\mbox{
\cite{pittaluga2019revealing}}\hspace{0pt}
. Continued deployment of robots in public spaces or in private or sensitive contexts, such as at home, may require the consistent use of privacy-preserving vision techniques to ensure that human data is handled appropriately, and humans do not have unresolved concerns about how robots with robotic vision capacity will operate safely in their own space}\DIFaddend . 

\subsection{Human Models}
\DIFdelbegin 

\DIFdelend Secondly, there are important challenges on model selection and optimisation for human behaviours, including the reliability and validity of behavioural phenomena to be captured and responded to through robotic vision. \DIFaddbegin \DIFadd{This systematic review presents several important use cases, including gesture and action recognition, human walking trajectories, object handover, social communication and learning from human demonstration. These range from both simple and complex behaviours that require the robot to understand the person. However, some of these behaviours are not as simple to interpret through the use of model selection and optimisation. }\DIFaddend One key example of this \DIFdelbegin \DIFdel{involves }\DIFdelend \DIFaddbegin \DIFadd{is }\DIFaddend the use of emotion recognition. Robotic vision that is dependent on state classification of human intent or emotion into static categories (i.e. happy or sad) can result in inaccurate identification and/or irrelevant responses provided from the robot without consideration of a more complex human emotional spectrum~\cite{doi:10.1177/1529100619832930}. For instance, behavioural science research has drawn attention to the unsuitable use of current computer vision methods to detect a person’s emotional state from their facial movements, instead calling for research that explores how people actually move their face to express emotion or other signals in different contexts~\cite{doi:10.1177/1529100619832930}. This research demonstrates that care should be taken around selection for what robotic vision should and should not be used during human interaction with robots. For example, inferring other human characteristics from visual data that could cause more harm than benefit to the interaction, such as classifying sexuality, race, or serious underlying medical conditions not known to the person. Incorrect model selection and optimization could also cause notable long-term problems in future robot deployments, if development and testing continues to optimise for behaviour that is not accurate or representative of the person, further contributing to bias~\cite{10.1145/3457607}. This also raises the question of whether simulated humans need to be involved in the simulation process, and the level of realism needed for this to be meaningful to the learning process. In addition, current methods that require large data sets may use data sets originally collected for other purposes, and therefore may not easily translate to a new context, such as a data-set of a busy crowd re-purposed to help robots learn about social norms in small groups. Lastly, people may eventually develop long-term hesitancy and rejection to use robotic systems due to perceptions that their capacities are nonfunctional after repeated errors, given the importance of robot performance on trust in the system~\cite{doi:10.1177/0018720811417254}.

\subsection{Experimental Design and Evaluation}
\DIFdelbegin 

\DIFdelend Thirdly, there has been limited human experimentation and evaluation of robotic vision with humans. As reported in the sections above, few studies report direct testing with humans, and for those who did report a form of testing with people, there were limited participant numbers across all studies. This is a challenge for future deployments, because exploration of participant characteristics found that many did not involve a large range of people who were representative of the general population, and instead involved a narrow sample with limited diversity~\cite{10.1145/3457607}. Therefore, robotic vision for human-robot interaction and collaboration could lead towards design and optimisation for a very restricted sample, further contributing to bias~\cite{10.1145/3457607}. For instance, people who provide feedback in experimental testing become the leading designers in future iterations of robot behaviour and function. This can create barriers for wider scale adoption when robots are deployed in the general community and inevitably encounter different kinds of people who have not been taken into consideration during design and refinement stages. Greater inclusion of different people has been the recent focus of co-design methodology for human-robot interaction and collaboration to ensure that robots demonstrate a more inclusive behaviour for a wider range of people who are likely to use them, e.g.~\cite{doi:10.1177/00187208211037465}. In addition, reported experiments often used single or simple evaluation metrics to measure robot perceptions and human-robot team performance, which may skew robotic vision evaluation in collaborative scenarios, without taking into consideration the human, the robot, and the team dynamic \cite{hoffman2019evaluating,damacharla2018common}. Such a simplified approach to testing and evaluation could further contribute to skewed development around how robotic vision should work to help people, if testing on human participants continues to remain low and only on restricted samples. 

\subsection{Appropriate Trust}
\DIFdelbegin 

\DIFdelend Lastly, there is the high potential that humans perceive robots that can interpret visual information to have a high sense of intelligence and general capacity to function with the person and within the environment~\cite{10.1145/3371382.3378347,7451807}. For instance, it may not be clear to the person to what extent the robot can identify only a limited visual field or target areas of interest, instead assuming the robot can view all of its surroundings and the activities that occur within it. The use of visual information to interact with the person may also contribute to an increased sense of anthropomorphic interpretation of the robot, leading people to perceive the robot as having more emotional expression or intelligence~\cite{duffy2003anthropomorphism}. This can also lead people to overly increase their confidence, trust or perception of the system, leading to people relinquishing greater autonomy or responsibility to the robot beyond what the robot is capable to perform on its own~\cite{Robinette2017}. For instance, people may falsely assume that the robot has human-like perception and cognitive abilities~\cite{10.1145/3371382.3378347,7451807,duffy2003anthropomorphism}, which can lead people to assume that the robot can perform better than it can. Misunderstanding around the capability of the robot could have notable consequences for safe and effective human-robot interaction and collaboration. For instance, the human assuming that the robot will detect visual hazards for the person, or recognise its own errors or mistakes in collaborative work. Therefore, visual information in human-robot collaboration and interaction should be used for the functional purpose of the robot, as well as explained to the people who use the robot, which includes both its potential strengths and limitations within the intended context to help regulate expectations and define the intended role of the robot~\cite{10.1145/3371382.3378347,7451807}. 

\section{Promising Areas of Future Research}\label{sec:future}

\DIFdelbegin \DIFdel{In addition to challenges specific to human-robot interaction, there are also more general challenges related to robotic and computer vision . While the performance of robotic vision systems are often bounded by what can be achieved by }\DIFdelend
\DIFaddbegin 
\DIFadd{There were several relevant computer vision methods that were not represented in the corpus of selected HRI/C papers. Four prominent examples are video convolutional networks, 3D human pose estimators, human--object interaction classifiers, and sign language recognition. Each of these could potentially have a significant impact on HRI/C research in the future.
}

To begin, recent state-of-the-art methods for action classification process video data using 3D convolutional networks~\cite{carreira2017quo, feichtenhofer2019slowfast},
unlike the predominantly frame-based classification approaches used in the HRI/C literature (see Section~\ref{sec:rq4}). New techniques include the inflated 3D convolutional networks
\cite{carreira2017quo} 
and the two-stream slow--fast network~\cite{feichtenhofer2019slowfast}. Action classification from video is critical to many HRI/C systems, but 3D convolutional networks tend to require significant \DIFaddend training data for \DIFdelbegin \DIFdel{new tasks, processing capabilities that are unavailable on the robotic platform (such as multiple GPUs), or it may have been invented too recently to have been adopted yet. For example, deep neural networks often fail to generalize, with a reduction in accuracy when tested outside of benchmark data sets \mbox{
\cite{recht2019imagenet}}\hspace{0pt}
; a method that achieves state-of-the-art performance on a benchmark dataset may not generalize immediately to a real-world setting on a robotic platform.
Sim-to-real transfer can be a particular challenge for many learning-based robotics applications, and progress is often relinquished to large companies who can implement large-scale data collection and testing~\cite{doi:10.1177/0278364917710318}. Additional challenges include software challenges such as the availability of open source libraries, software development kits, and components }\DIFdelend \DIFaddbegin \DIFadd{fine-tuning and significant GPU resources for inference, making it challenging to transfer to many HRI/C systems. However, the expansion of models pre-trained on diverse datasets, coupled with developments in transfer learning, help mitigate this difficulty. Therefore, it is expected that pre-trained video action recognition networks could become a commonly used tool in HRI/C research. While 2D human pose estimation was well-represented, 3D human pose estimation was mostly absent, despite the fact that providing spatial and shape information could be very useful for HRI/C to enable the robot to perform more accurate and functional actions with the person. Monocular 3D human pose estimation from a single image or video is a very popular topic in computer vision. Model-free approaches \mbox{
\cite{pavlakos2018ordinal, pavllo20193d} }\hspace{0pt}
include VideoPose3D \mbox{
\cite{pavllo20193d}}\hspace{0pt}
, which estimates 3D joint locations using temporal convolutions over a sequence of 2D joint detections. Model-based approaches \mbox{
\cite{kanazawa2018end, kanazawa2019learning, kolotouros2019convolutional, kocabas2020vibe} }\hspace{0pt}
predict the parameters (e.g., joint angles, shape, and transformation) of a body model, }\DIFaddend such as the \DIFdelbegin \DIFdel{lack of training data for specific use cases and platforms. Hardware performance can also be limited by robot on-board processing compared to cloud processing and }\DIFdelend \DIFaddbegin \DIFadd{SMPL mesh model \mbox{
\cite{loper2015smpl}}\hspace{0pt}
. Adversarial learning can be used \mbox{
\cite{kanazawa2018end, kanazawa2019learning, kocabas2020vibe} }\hspace{0pt}
to generate realistic body poses and motions, which tends to generalize better to unseen datasets, and therefore, may be more appropriate for HRI/C tasks. Modelling humans in 3D allows physics to be taken into account, allowing the robot to plan and respond more appropriately and preventing it from making non-physical predictions.
}

\DIFadd{Another set of techniques of relevance to HRI/C are those developed for human--object interaction classification \mbox{
\cite{chao2018learning, qi2018learning, gupta2019nofrills}}\hspace{0pt}
. This task aims to extend action recognition to interaction recognition: localising and describing pairs of interacting humans and objects in the scene. A robot that collaborates with people to perform a task would strongly benefit from knowing which object the person is interacting with at that point in time, and what type of interaction is taking place. For example, an airport assistance robot may need to detect instances of ``person carrying suitcase'' to determine where best to provide support to the person. Methods for this task almost always detect human and object bounding boxes first, before combining information from different modalities (appearance, relative geometry, human pose) using multi-stream networks \mbox{
\cite{chao2018learning, gupta2019nofrills} }\hspace{0pt}
or graph neural networks \mbox{
\cite{qi2018learning,zhang2021scg}}\hspace{0pt}
. There is a clear case for widespread use of these techniques in HRI/C, to facilitate higher-order reasoning about what the people proximal to }\DIFaddend the \DIFdelbegin \DIFdel{availability, performance and cost of hardware solutions, such as the RGB-D camera as an inexpensive source of depth information compared with expensive alternatives like laser range sensors \mbox{
\cite{susperregi2013rgb}}\hspace{0pt}
. Computationally intensive algorithms can lead to the requirement to have a graphics processing unit in the robot , which can be heavy, noisy and require a lot of power \mbox{
\cite{pena2017benchmarking}}\hspace{0pt}
. Other challenges include the translation of signals from computer vision into actionable and useful robot functions, such as movement and manipulation actions that can improve the robot's utility for a given task \mbox{
\cite{macdonald2019active}}\hspace{0pt}
. This could slow down the deployment of robotic vision in human domains}\DIFdelend \DIFaddbegin \DIFadd{robot are doing, and with what objects.
}

\DIFadd{In general, substantial progress has been made in computer vision and machine learning since 2020. While beyond the scope of this manuscript, there are significant opportunities for HRI/C arising from these developments. In particular, the Transformer architecture~\mbox{
\cite{vaswani2017attention}}\hspace{0pt}
, originally proposed for natural language processing, has begun to supplant or supplement convolutional neural networks for vision tasks, with large performance increases across many tasks. These include image recognition~\mbox{
\cite{dosovitskiy2020image}}\hspace{0pt}
, object detection~\mbox{
\cite{carion2020end}}\hspace{0pt}
, video understanding~\mbox{
\cite{bertasius2021space,arnab2021vivit,patrick2021keeping}}\hspace{0pt}
, and human--object interaction~\mbox{
\cite{zhang2021upt}}\hspace{0pt}
. This represents a significant opportunity for the HRI/C community, because it is a general-purpose architecture that facilitates multi-modal sensor processing \mbox{
\cite{jaegle2021perceiver}}\hspace{0pt}
, allowing a robot to reason about its video, audio, and other inputs jointly. This is likely to expand and robustify robot capabilities while interacting or collaborating with people}.

\DIFadd{There are also additional areas in which computer vision research has potential impact to adapt or improve HRI/C across different settings. One notable area is sign language recognition \mbox{
\cite{pfister2013largescale, li2020transferring, albanie2020bsl}}\hspace{0pt}
, which was not present in the corpus of papers. This represents an opportunity for further developing methods for non-verbal communication in HRI/C. The techniques developed, involving fine-grained gesture recognition and multi-modal learning, are relevant for HRI/C, since these techniques can provide benefit to human-robot communication, as well as general situational awareness. Other area is autonomous and assisted driving, in which robotic vision for HRI/C could have a notable impact to increase the uptake, efficiency and safety of autonomous vehicles~\cite{sadigh2016planning,driggs2017integrating,halin2021survey}. For example, autonomous driving requires that the car can detect and predict the trajectories of others around and on the road, such as drivers, cyclists, and pedestrians. This process can involve close monitoring and coordination to ensure that humans can safely move around autonomous vehicles while vehicles can also get to their intended destination. There is currently a growing body of work around the human component in autonomous driving, but most of these works have so far been tested in simulated environments and without vision, creating notable opportunities to explore new areas of robotic vision for HRI/C style tasks in the near future~\cite{sadigh2016planning,driggs2017integrating,halin2021survey, 8876650, 9743954}. Other areas also include to further explore the capacity to anticipate human actions in advance, resulting in robot behaviour that can be more reactive than passive to respond to dynamic interaction patterns over time~\cite{7451737,ognibene2013contextual,8594452}. 
}

\DIFaddbegin 

\DIFaddend 
\section{Conclusion}\label{conclusion}
\DIFdelbegin \DIFdel{Robotic vision has }\DIFdelend \DIFaddbegin \DIFadd{This survey and systematic review provided a comprehensive overview on robot vision for human-robot interaction and collaboration with a detailed review of papers published in the last 10 years for robots that can perceive and take action to facilitate a high-level task. Robotic vision had }\DIFaddend the capacity to improve human robot interaction and collaboration (HRI/C), including to create new ways for humans and robots to work together. This survey and systematic review provided an extensive analysis on the use of robotic vision in human-robot collaboration and interaction into common domains, areas and performance metrics for robotic vision. This includes exploring \DIFdelbegin \DIFdel{important }\DIFdelend how computer vision has been adapted and translated through robotic vision to improve aspects of human-robot interaction and collaboration. This survey and systematic review also contributed to identifying application areas that had not yet been attempted, and how techniques from the computer vision research could help to inform human-focused vision research in robotics. It was found that robotic vision for improving the capacity of robots to collaborate with people is still an emerging domain. Most works involved a one-on-one interaction, and focused on using robotic vision to enhance a specific feature or function related to the interaction. It was also found that only some high-impact and novel techniques from the computer vision field had been translated for human-robot interaction and collaboration, highlighting an important opportunity to improve the capacity of robots to engage and assist people. More novel and emerging areas in the HRI/C field such as multi-human, multi-robot teams were less represented in the corpus of papers~\cite{iqbal2017coordination, liu2016multirobot, 1200160}. Furthermore, robotic vision was often tested in a simple or single application field for each specific use case, showing limited depth in its current form. 

\DIFaddend Future pathways for human-robot interaction and collaboration involve the creation and development of robotic platforms using vision-related information to create more competent robots that can operate in dynamic environments with people. For instance, improving robots \DIFaddbegin \DIFadd{to better handle multiple visual inputs at once }\DIFaddend to open up new domains or collaborative tasks, such as multi-human multi-robot teams. Robotic vision could therefore help to break down some barriers present in long-term human-robot teamwork, such as better adaptation to dynamic environments and different kinds of people over a long period of time. 

\section{Acknowledgments}
This research was supported by the Australian Research Council project number CE140100016. D.C received funding from Continental AG.

\bibliographystyle{acm}
\bibliography{references, includes}
\end{document}